\documentclass[journal]{IEEEtran}
\usepackage{color}
\usepackage{enumitem,xcolor}
\usepackage{soul}
\newlist{coloritemize}{itemize}{1}
\setlist[coloritemize]{label=\textcolor{blue}{\textbullet}}

\usepackage{cite}

\ifCLASSINFOpdf
   \usepackage[pdftex]{graphicx}
\else
\fi
\usepackage[cmex10]{amsmath}
\usepackage{amssymb,amsfonts}
\usepackage{siunitx} 
\ifCLASSOPTIONcompsoc
 \usepackage[caption=false,font=normalsize,labelfont=sf,textfont=sf]{subfig}
\else
 \usepackage[caption=false,font=footnotesize]{subfig}
 \usepackage{dblfloatfix}

\usepackage{url}

\usepackage{fix-cm}
\usepackage[colorinlistoftodos, textwidth=3.7cm]{todonotes}
\usepackage{bm}
\usepackage{nomencl}
\makenomenclature

\def\x{{\boldsymbol{x}}}
\def\y{{\boldsymbol{y}}}
\def\z{{\boldsymbol{z}}}
\def\zhatx{{\widehat{z}_{x}}}

\def \ztildeyb{{\boldsymbol{\widetilde{z}_y}}}
\def\zhatyx{{\widehat{z}_{y|x}}}
\def\zhatyxb{\boldsymbol{{\widehat{z}_{y|x}}}}
\def\zhaty{{\widehat{z}_{y}}}
\def\zhatyb{{\boldsymbol{\widehat{z}_{y}}}}
\def\xcal{{\mathcal X}}

\def\zcal{{\mathcal Z}}
\def\D{{D}}
\def\G{{G}}

\newcommand{\figr}{Fig.~\ref}				
\newcommand{\tabref}{Tab.~\ref}				
\newcommand{\eq}{Eq.~\eqref}					
\newcommand{\Sec}{Sec.~\ref}					

\newcommand{\sen}{Sentinel-1}

\begin{document}
%
\title{On the potential of sequential and non-sequential regression models for Sentinel-1-based biomass prediction in Tanzanian miombo forests}
%
%

\author{Sara~Björk, \textit{Associate Member}, \textit{IEEE},
        Stian~Normann Anfinsen, \textit{Member}, \textit{IEEE}, 
        Erik~N\ae sset, 
        Terje~Gobakken
        and~Eliakimu~Zahabu
\thanks{Manuscript received February 6, 2022; revised 10 May, 2022; accepted 23 May; 2022.}
\thanks{S.\ Björk and S.\ N.\ Anfinsen are with the Machine Learning Group, Department
of Physics and Technology, UiT The Arctic University of Norway, 9037 Troms\o,  Norway (e-mail: sara.bjork@uit.no).}
\thanks{S.\ Björk is with the Applied Deep Learning DevOps team, KSAT Kongsberg Satellite Services, 9011 Troms\o, Norway.}
\thanks{S.\ N.\ Anfinsen is with the Earth Observation Group, Energy \& Technology Department, NORCE Norwegian Research Centre, 9019 Troms\o, Norway.}
\thanks{E.\ N\ae sset and T.\ Gobakken are with Faculty of Environmental Sciences and Natural Resource Management, Norwegian University of Life Sciences, 1432 \AA s, Norway.}
\thanks{E.\ Zahabu is with Department of Forest Resources Assessment and Management, Sokoine University of Agriculture, P.O. Box 3013, Chuo Kikuu, Morogoro, United Republic of Tanzania.}
}

%
%

\markboth{IEEE JOURNAL OF SELECTED TOPICS IN APPLIED EARTH OBSERVATIONS AND REMOTE SENSING}%
{XXXX}
%



\maketitle

\begin{abstract}
This study derives regression models for above-ground biomass (AGB) estimation in miombo woodlands of Tanzania that utilise the high availability and low cost of Sentinel-1 data. The limited forest canopy penetration of C-band SAR sensors along with the sparseness of available ground truth restrict their usefulness in traditional AGB regression models. Therefore, we propose to use AGB predictions based on airborne laser scanning (ALS) data as a surrogate response variable for SAR data. This dramatically increases the available training data and opens for flexible regression models that capture fine-scale AGB dynamics. This becomes a sequential modelling approach, where the first regression stage has linked in situ data to ALS data and produced the AGB prediction map; We perform the subsequent stage, where this map is related to Sentinel-1 data. We develop a traditional, parametric regression model and alternative non-parametric models for this stage. The latter uses a conditional generative adversarial network (cGAN) to translate Sentinel-1 images into ALS-based AGB prediction maps. The convolution filters in the neural networks make them contextual. We compare the sequential models to traditional, non-sequential regression models, all trained on limited AGB ground reference data. Results show that our newly proposed non-sequential Sentinel-1-based regression model performs better quantitatively than the sequential models, but achieves less sensitivity to fine-scale AGB dynamics. The contextual cGAN-based sequential models best reproduce the distribution of ALS-based AGB predictions. They also reach a lower RMSE against in situ AGB data than the parametric sequential model, indicating a potential for further development.
\end{abstract}

\begin{IEEEkeywords}
Aboveground biomass (AGB), airborne laser scanning (ALS), conditional adversarial generative network (cGAN), \sen, sensor fusion, synthetic aperture radar (SAR).
\end{IEEEkeywords}

%
\IEEEpeerreviewmaketitle


\section{Introduction}
\label{sec:intro}

\IEEEPARstart{A}{s} a consequence of climate change, there is an increasing need for accurate carbon accounting systems for measuring, reporting, and verification (MRV) on a national level. Through the REDD+ programme (officially named "Reducing emissions from deforestation and forest degradation and the role of conservation, sustainable management of forests and enhancement of forest carbon stocks in developing countries"), developing countries are motivated to implement such an MRV system to monitor the potential reduction of carbon emissions from tropical forests \cite{UNFCC2011}. The documentation of reduced deforestation on a national level could potentially result in a financial reward being released through the program for the countries associated with the REDD+ programme \cite{ene2017}. 

Forests are well known for being one of the major carbon sinks and need to be properly and accurately monitored by the MRV system. This can be achieved by accurately estimating the amount of forest aboveground biomass (AGB), as AGB is a primary variable related to the carbon cycle \cite{Kaasalainen2015,bombelli2009biomass}. To calibrate the MRV system, AGB data over the area of interest (AOI) is needed. It can be collected either through destructive or non-destructive \textit{in situ} sampling. The former implies harvesting, drying, and weighing the plants to estimate the biomass. The latter does not involve harvesting trees but measuring parameters such as tree height, stem diameter, etc. Measured parameters from the non-destructive sampling can be used to predict AGB by allometric models developed for the AOI \cite{bombelli2009biomass}. Unfortunately, AGB \textit{in situ} measurements of both above categories are costly and time-demanding to collect manually. As a consequence, most research instead focuses on establishing a relationship between a small amount of AGB field data and remote sensing (RS) data using different sensors \cite{le2002relationships, landsat_biomass, ene2016, ene2017, SANTI2017ThePotential,GHOSH2018AGBestimation, StelmaszczukGorska2018EstimationofAGB, Sinha2019MultiSensorApproach, Debastiani2019EvaluatingSAR, Narine2019SynergyofICESat2, Zhang2019DLBasedRetreival, Chen2019OptimalCombination, santi2020MLapplications, Li2020ForestAGBestimation, Zhang2020AnEvaluationof, Nuthammachot2020Combineuse}. 

Among different platforms and sensor types, airborne laser scanning (ALS) systems are shown to provide AGB models that are significantly more accurate than models developed using radar or passive optical data \cite{zolkos, Galidaki}. The reason is probably that ALS can provide accurate data describing canopy cover density and canopy height, which is highly correlated with forest AGB \cite{Galidaki, Kaasalainen2015}. This result was also confirmed in \cite{naessetMappingEstimatingForest2016}, where the ALS-based regression model achieved the highest accuracy of AGB estimates in the miombo woodlands of Tanzania. However, airborne data are associated with high acquisition cost, which limits the use of ALS data in national MVR systems that require regular acquisitions to keep forest inventories up to date \cite{Kaasalainen2015, Galidaki}.

One of the advantages of employing spaceborne SAR sensors to AGB estimation is that it provides data with extensive spatial coverage that can be acquired with high temporal frequency. SAR data can thus yield frequently updated AGB predictions over large areas. Another advantage is the SAR sensor's ability to penetrate clouds, which makes it effective to monitor regions with a significant amount of cloud coverage. Unfortunately, the use of SAR data for AGB estimation is limited by the saturation level, the property that SAR intensity does not increase with AGB beyond a certain AGB level. This property is dependent on the specific wavelength used by the SAR sensor and implies, in general, that AGB at middle-to-high level cannot be distinguished in the SAR intensity data \cite{Kaasalainen2015,Urbazaev2018EstiamtionofAGB, Santoro2010, Santoro2011}. Additionally, SAR data are strongly dependent on the environmental conditions on the ground, where a change in moisture conditions impacts the measured backscatter \cite{Urbazaev2018EstiamtionofAGB}. The former is a well known limitation of SAR data that may restrict its use in MRV systems of high precision, the latter might be circumvented by the use of SAR data acquired at e.g.\ dry seasons \cite{Santoro2010}. The different challenges of SAR and ALS have fostered studies on their combined use for forest AGB estimation. Several of these studies were reviewed in \cite{Kaasalainen2015} and \cite{ Galidaki}, which conclude that the combination of SAR and ALS may improve AGB estimation, especially when SAR data are used to upscale and extend accurate ALS measurements of forest height to obtain accurate AGB predictions over large areas \cite{Kaasalainen2015}.

Well-known regression models from statistics have traditionally been used to directly relate a small set of ground reference data of AGB to RS data from a single sensor. A popular choice among the conventional regression models is a variation of traditional linear regression: multiple linear regression and stepwise multiple regression, see e.g \cite{Phua2017SynergicUse, Sinha2019MultiSensorApproach, Zhang2019DLBasedRetreival, Chen2019OptimalCombination, Li2020ForestAGBestimation, Nuthammachot2020Combineuse, sinha2021assessment}. The evolution of machine learning (ML) methods has introduced many alternative methods for AGB estimation, with random forests, artificial neural networks (ANNs) and support vector machines for regression as some of the most prominent, see e.g.\ \cite{ GHOSH2018AGBestimation, StelmaszczukGorska2018EstimationofAGB, Debastiani2019EvaluatingSAR, Zhang2019DLBasedRetreival, Chen2019OptimalCombination, santi2020MLapplications, Li2020ForestAGBestimation, Zhang2020AnEvaluationof, Chen2018EstimationofAGB, Vafaei2018ImprovingAccuracy, CHEN2019Assessmentofmultiwave, Yang2020ANewMethod}. Like the traditional statistical regression models, these ML-based models also directly relate ground reference data of AGB to RS data from a single sensor. Due to the limited amount of ground reference AGB data, both traditional statistical regression models and ML-based models are restricted to relate single observations of the ground reference AGB data to single pixels from the RS data source. Thus, the spatial contextual information from neighbouring pixels in the RS data source are generally not incorporated in the learning of the regression model. This is likely to inhibit the learning of the AGB dynamics and fine scale variability. The emerging field of deep learning (DL) methods has further opened many new possibilities in the analysis of RS images. Deep neural networks (DNNs) have, among other things, increased the ability to perform accurate regression between different image modalities acquired from different sensors at possibly different times. 
The combination of multimodal RS images, such as e.g.\ SAR and ALS, has been shown to improve AGB estimation results through regression models of increased complexity. Although the different RS images cover the same scene, their pixel measurements represent different domains, like for example ALS-derived measurements of heights or SAR-based backscatter intensity data. Transfer learning (TL), domain adaptation (DA) \cite{panyang, weiss_TL, Zhang_TL} and image translation \cite{isola} are some theoretical frameworks of recent popularity that can be used to handle such challenging and complex problem settings. Also, a challenging regression problem arises when data from different multimodal RS sensors are combined to upscale the extent of an accurate sensor-based AGB prediction map. In the context of such a data fusion task, sequential approaches with two subsequent regression models become relevant as an alternative to the simpler strategy with a single-stage regression model.

In this work we refer to \emph{sequential modelling} as the process where two regression models are used in a chain to achieve more training data for AGB prediction. Sequential modelling can also be used to upscale the spatial extent of an initial AGB prediction map. In the first stage, one regression model relates ground reference AGB data to a single RS data source with high information content about the target variable, but with limited geographical coverage. The outcome of the first model is an accurate sensor-based AGB prediction map, which is used in the second regression model as a surrogate for ground reference data to regress on data from an additional RS sensor with larger spatial extent. Both traditional regression models, such as simple and multiple linear regression (see e.g.\ \cite{NEIGH2013TakingStock, solbergEstimatingSprucePine2010, englhartAbovegroundBiomassRetrieval2011}), and ML-based models, such as random forest and support vector regression (e.g.\ \cite{WANG202010EstimatingAGB, Hudak2020ACarbonMonitoringSystem, Wang2019mappingHeight, Cartus2012MappingCanopy}), have previously been applied in a sequential modelling fashion for AGB estimation. In this work we differentiate between sequential modelling and the traditional approach with a single-stage regression model by referring to the latter as a \emph{non-sequential modelling} approach.

Both sequential and non-sequential regression models for AGB estimation have traditionally operated on an individual pixel level. That is, the prediction at a pixel location is based on regressors exclusively from the same location, without any use of spatial context of neighbouring pixels. However, a key feature of DNNs, that partly explains their success in many prediction and regression problems, is their use of convolutional filters. This implies that the prediction of any single pixel is based on regressors from a spatial neighbourhood that surrounds it. It also means that the prediction is done by processing blocks of pixels, with image layers of regressor variables in input and a corresponding layer for the response variable in output. This mapping of predictor images to a response variable image is equivalent to the operation known as \emph{image translation} in DL.
Isola \emph{et al.}\ \cite{isola} define image translation as follows: \textit{Given sufficient training data, image-to-image translation is defined as the problem of translating one possible representation of a scene into another}. Within DL, the family of generative models is known to enable cross-modal image translation by translating data from one known distribution to another target distribution. Amongst the generative models are the generative adversarial networks (GANs) \cite{gan2014} particularly popular, see e.g.\ \cite{StarGAN, isola, Karras2020a, karras_style-based_2019, Radford2015DCGAN, sarsim, Rev1Suggestion, Review2_suggestion}. GANs are trained to capture the data distribution of a target domain in a minimax optimisation procedure. After training, the generator network, $\G$, can be used to map a random noise vector to a target output image. This idea was later extended to the conditional generative adversarial network (cGAN) architecture \cite{mirza2014conditional}. In the cGAN setting, the learnt mapping to the target output image distribution is conditioned on the distribution of an input image \cite{isola}.
Considering the enormous potential of GANs, we wish to address AGB prediction from a DL perspective. However, as a DNN, the cGAN model requires a substantial amount of training data for cross-modal image translation. Therefore, it cannot learn to directly translate between a small set of AGB ground reference data and spatially continuously RS data. Thus, we propose to tackle the regression problem through sequential modelling by applying the cGAN architecture in the second regression model in the sequence. This approach is only possible as we propose to use an AGB prediction map as a surrogate for ground reference data, which makes a large amount of spatially continuous target data available to the regression model. The cGAN's convolutional filters open for the use of spatial contextual information in the predictions. Based on the discussion above, the definition of the research problem in this article is described as follows.

\begin{figure}
    \centering
    \includegraphics[height=9cm,width=1\columnwidth]{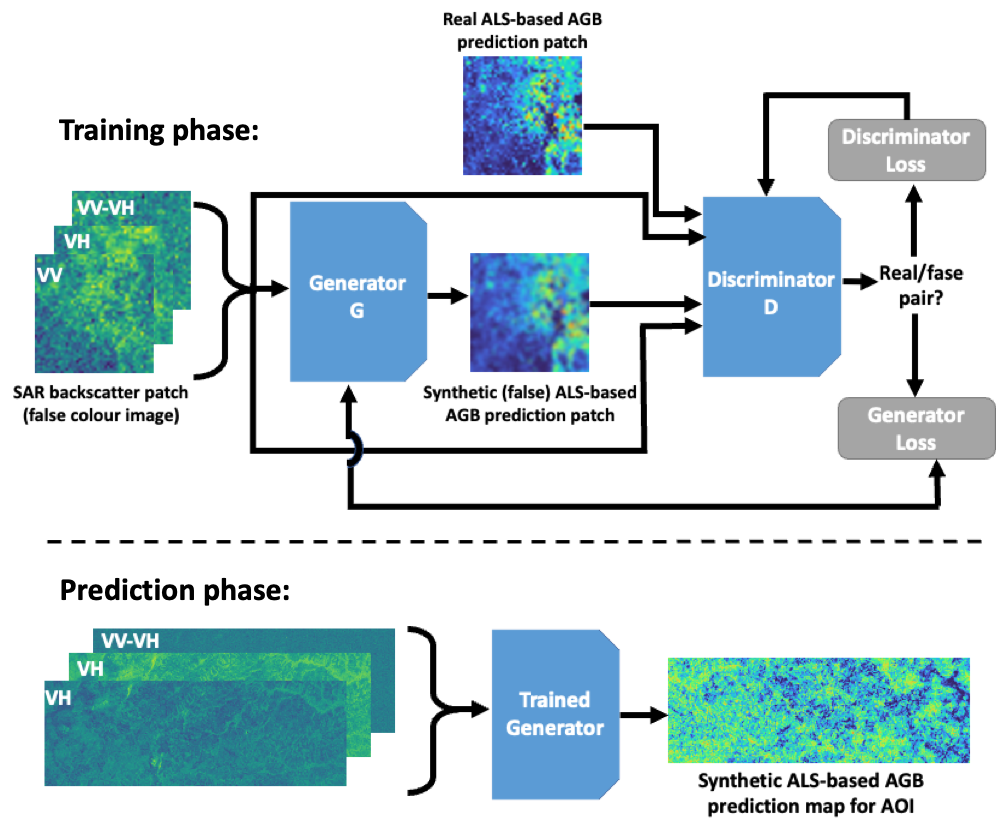}
    \caption{Flowchart over the proposed cGAN-based sequential modelling approach. The generator network is trained to translate false colour Sentinel-1 backscatter patches (consisting of the VV and VH band and their difference, i.e.\ VV-VH) into realistic looking synthetic ALS-based AGB prediction patches. The discriminator network is trained to distinguish between a "real" combination of the input patch from \sen\ and the actual AGB prediction patch to a "fake" combination of the input patch from \sen\ and the synthetic AGB prediction patch. The cGAN components, the $\G$ network and the $\D$ network, are trained in a minimax optimisation procedure. After training, the $\G$ network can generate realistic looking synthetic ALS-based AGB prediction patches in an AOI from corresponding false colour \sen\ data in the AOI (see prediction phase). Both the individual bands of the false colour SAR patch and the ALS-based AGB patches only consist of one channel but are here represented in colours to ease the interpretation.}
    \label{fig:flowchart}
\end{figure}

\subsection{Problem Definition}\label{sec:probdef}

As a developing country and associated with the REDD+ programme, Tanzania has the potential to achieve a financial benefit by implementing an MRV system to monitor their forests. Therefore, the primary aim of this work is to develop forest AGB prediction models that could be implemented in an MRV system for Tanzania. For an AGB prediction model to be of practical use in the MRV system of Tanzania, the model should be able to provide frequently updated AGB predictions with extensive spatial coverage, of a high accuracy, and at a low cost. This puts some constraints on the data used: 
\begin{enumerate}
    \item We need to rely on remote sensing data, as large scale \textit{in situ} sampling will be infeasible, 
    \item We cannot afford performing frequent ALS campaigns to frequently update a low-cost MRV system, 
    \item Due to its location, Tanzania experiences rain periods, which constrains the use of passive sensors, as they are not able to penetrate clouds. 
\end{enumerate}
The second constraint further limits the use of RS data from sensors that are neither freely available, nor easily accessible. Based on the constraints of this project, we have decided to utilise the \sen\ sensor, as it provides us with freely available and frequently updated data with extensive spatial coverage. However, a simple SAR-based AGB prediction model may limit the precision of the MRV system and consequently the advantage of implementing the system for operational forest monitoring.

Both \cite{Kaasalainen2015} and \cite{luSurveyRemoteSensingbased2016} advocate the potentials of combining ALS and SAR for large-scale AGB mapping with improved accuracy. Encouraged by this, we restrict the focus of this work to an AOI in the Liwale district in southeast Tanzania. Here, we have access to a small amount of ground reference vector data and continuous raster of ALS data, which has previously been used in combination with four other RS datasets: optical  RapidEye  and  Landsat  imagery, interferometric TanDEM-X radar imagery (X-band SAR), and ALOS-PALSAR (hereby PALSAR) radar imagery (L-band SAR), to develop five different traditional non-sequential regression models, see \cite{naessetMappingEstimatingForest2016}. The ALS-based prediction model of Næsset \emph{et al.}\ was further used in \cite{naessetMappingEstimatingForest2016} to create a wall-to-wall map of ALS-based forest AGB predictions. Their ground reference dataset and the wall-to-wall map of ALS-based forest AGB predictions were provided to us for this work, and will be used together with \sen\ data to develop low cost AGB prediction models for the AOI. However, since we aim to contribute with AGB prediction models that can be applied not only in the AOI, but also in extended areas, we put further restrictions on the focus of this work:
\begin{enumerate}[label=(\roman*)]
    \item To develop AGB prediction models of high accuracy and with potentially extensive spatial coverage, we wish to investigate if a sequential modelling approach is better than a traditional non-sequential regression model.
    \item By utilising the wall-to-wall ALS-based AGB prediction map as a surrogate for AGB ground reference data, we are able to implement the second part of the sequential model with a deep neural network. Thus, in the case of sequential modelling, we additionally investigate the possible benefits of applying a DL-based model instead of a traditional regression model.
\end{enumerate}

Our approach to sequential modelling is to co-register and resample the SAR intensity image data to the same spatial resolution as the available wall-to-wall map of ALS-based AGB predictions, produced with the classical non-sequential regression model presented in \cite{naessetMappingEstimatingForest2016}. Motivated by the achievements of image-to-image translation, we propose to utilise a cGAN model for the second model in the sequence. We train the cGAN model to synthesise ALS-based AGB maps from false colour SAR intensity images. As far as we know, this is the first time contextual DNNs, in the form of cGAN models, have been utilised in a sequential modelling strategy to upscale a limited amount of ground reference data and simulate AGB predictions. We see any modification of the ALS-based regression model as outside the scope of this work. \figr{fig:flowchart} shows the overall view of the proposed cGAN-based sequential approach used to generate synthetic ALS-based AGB predictions from false colour \sen\ image patches. We validate the proposed cGAN-based sequential model against two non-contextual \sen-based regression models, also proposed for this work; a non-sequential model and a traditional sequential model. The non-sequential regression model relates single pixels of \sen\ data to the small set of AGB ground reference data. For the non-contextual sequential regression model, we trained the second model in the sequence to relate ALS-based AGB predictions to single pixels of \sen\ data. For both non-contextual models, we use the state-of-the-art regression model in the AOI, i.e. a multiple linear regression model with square root transformation of the response variable. This is the same regression model as used by Næsset \emph{et al.}\ in their work \cite{naessetMappingEstimatingForest2016}. 

\subsection{Contribution}\label{sec:contribution}

To summarise, the contributions of this paper are: \begin{enumerate}
    \item We extend the work in \cite{naessetMappingEstimatingForest2016} by developing a similar type of regression model based on \sen\ data. 
        \item We propose to model forest AGB by a novel sequential modelling approach, in which the second model relates SAR data to ALS-based AGB predictions. We propose two different regression models for the second stage of regression:
    \begin{enumerate}[label=(\roman*)]
        \item one traditional regression model, similar to 1).
        \item one DL-based regression model based on image-to-image translation with a cGAN \cite{isola}.
    \end{enumerate}
     \item Since the application of cGANs as AGB regression models is uncommon, we provide a comprehensive study on different hyperparameters, objective functions, and $\G$ and $\D$ networks.
    \item We empirically evaluate the three proposed  AGB prediction models against previous results presented in  \cite{naessetMappingEstimatingForest2016} and against each other.  
    \item We demonstrate the potential of using \sen\ data for AGB predictions and show that our C-band-based models performs better than some of the previously developed models for the AOI.
\end{enumerate}
While we argue for the benefit of using \sen-based models to extend the spatial coverage of the AGB predictions, the scope for this study is to develop models for the AOI. We therefore see the construction of AGB prediction maps over an extended area as outside the scope of this work.

The remainder of this paper is organised as follows: In \Sec{sec:back} we introduce our proposed sequential modelling approach for forest AGB prediction. \Sec{sec:related} presents published research in related areas within non-sequential and sequential regression models for AGB prediction through sensor fusion, and related research on image translation through GANs. \Sec{sec:method} presents the datasets, and formally define the proposed non-sequential and sequential regression models. Results are presented and analysed in \Sec{sec:res}, while we discuss our work in \Sec{sec:dis}. Finally, we conclude our work in \Sec{sec:conclusion} by summarising the most important findings. Additional experiments and  methodological contributions are collected in the Appendix. 
%
%
%
%

\begin{figure*}
\centering
\includegraphics[scale=.5]{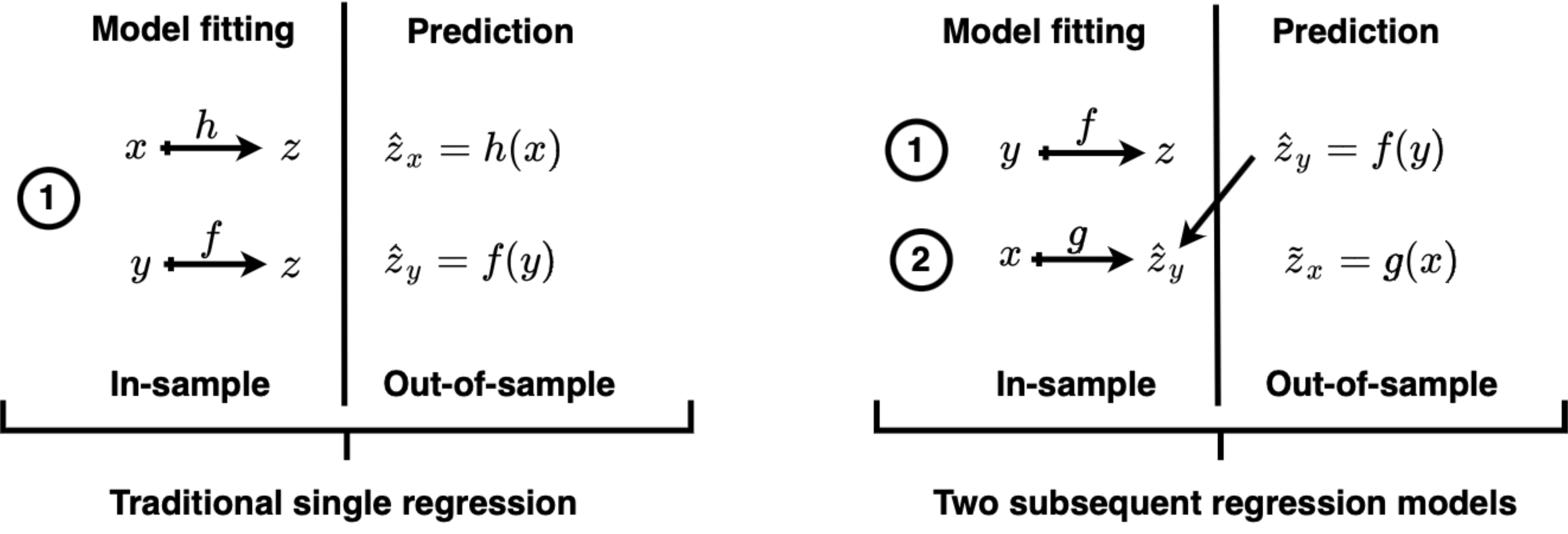}
\caption{Illustration of the difference between a traditional non-sequential regression model and the proposed sequential regression models. We let $x$ denote data from a SAR sensor,  $y$ denote ALS data and $z$ denote AGB ground reference data. Regression models are represented by \textit{f}, \textit{g} and \textit{h}, where \textit{f} is a regression model between $y$ data and $z$, \textit{h} is a regression model between $x$ data and $z$, while \textit{g} is a regression model between $x$ data and ALS-based AGB predictions denoted $\widehat{z}_y$. Additionally, $\widehat{z}_x$ denote SAR-based AGB predictions from a traditional non-sequential regression model. In the sequential setting, $\zhatyx$ denote the outcome from the second part of the two subsequent regression models, i.e.\ a generated synthetic ALS-based AGB predictions retrieved from $x$ data.}
\label{fig:workflow}
\end{figure*}

\section{Background}\label{sec:back}
In this section, we introduce the proposed sequential modelling approach for forest AGB prediction in both general terms and with a particular emphasise on employing a cGAN for the second part of the sequential model. We continue with a general introduction to the concepts of the cGAN model and how it can be utilised for image-to-image translation in our sequential modelling approach.

\subsection{Non-sequential modelling}
As previously introduced, co-located ALS data ($y$) and AGB ground reference data ($z$) consisting of 88 field plots were in  N\ae sset~\emph{et~al.} \cite{naessetMappingEstimatingForest2016} used to fit a traditional non-sequential regression model $f:y\mapsto z$. The specific regression model from \cite{naessetMappingEstimatingForest2016}, denoted \textit{f}, uses a square root transformation of the response variable and was trained using ordinary least squares (OLS) regression with stepwise forward selection of the variables. It was used to map spatially continuous ALS measurements into what we refer to as a ALS-based AGB prediction map by:
\begin{equation*}
\widehat{z}_y=f(y)\,,
\end{equation*} 
where $\zhaty$ denotes each individual ALS-based AGB prediction. The regression coefficients are published in \cite{naessetMappingEstimatingForest2016} and the resulting prediction map has been made available to us by the authors.
The traditional non-sequential approach is illustrated on the left-hand side of \figr{fig:workflow}, where a single regression model is trained to relate some remotely sensed predictor, such as SAR backscatter intensity (denoted $x$) or ALS data ($y$), to a co-located set of sparse AGB ground reference data ($z$). Here, $\widehat{z}_x$ refers to SAR-based AGB predictions obtained with the traditional non-sequential regression model. The ALS-based biomass prediction map, $\widehat{z}_y$, is of relatively high accuracy compared to maps made from other remote sensing data sources in the same work \cite{naessetMappingEstimatingForest2016}. 

\subsection{Sequential modelling}

In the modelling strategy with two sequential regression models, we keep the regression model from \cite{naessetMappingEstimatingForest2016}, i.e.\ $f$, as the first model in the sequence. We then propose the second regression model in the sequence to relate SAR backscatter intensity data, $x$, to wall-to-wall maps of ALS-based forest AGB predictions, $\widehat{z}_y$. We thereby utilise $\widehat{z}_y$ as a dense surrogate for $z$. This gives rise to the second regression model, $g:x\mapsto\zhaty$, which in the prediction phase can be used to map SAR images, unseen by the model, to generate synthetic ALS-based AGB maps by:
\begin{equation*}
\zhatyx=g(x)\,,
\end{equation*} 
where $\zhatyx$ denotes each individual generated synthetic ALS-based AGB prediction. Thus, the two regression models $f$ and $g$ link SAR intensity data to AGB ground reference data in a sequential process. The main benefit of the sequential modelling approach is that the model $g$ can be trained with a large amount of spatially continuous data instead of the few ground reference field plots. Consequently, our sequential modelling approach additionally facilitates for the full exploitation of convolutional DL models for AGB regression as they require access to spatially continuous data.  Our proposed sequential modelling approach is shown on the right-hand side of \figr{fig:workflow}. It should be noted that the described sequential approach is lacking in one respect: The SAR-based prediction, $\zhatyx$, is regressed against a surrogate regression target $\zhaty$, which despite its relatively high accuracy must necessarily contain some uncertainty. Therefore, the sequential modelling could be followed by a calibration step step where the mean of $g$ is calibrated against the original ground reference data, $z$. This is discussed in footnote \ref{footnote:calibration} on page \pageref{footnote:calibration}. 

We propose two different versions for model $g$: a traditional sequential model and a DL-based sequential model. In the traditional sequential regression setting, we let \textit{g} take the same form as \textit{f}, i.e.\ a multiple linear regression model with square root transformation of the response variable. In the DL-based sequential regression setting we instead use a cGAN model as the second regression model. The latter is only possible due to the sequential modelling approach, which allows \textit{g} to be trained on the wall-to-wall map of ALS-based AGB predictions. As the cGAN model utilises convolutional filtering to exploit the contextual information between neighbouring pixels, it carries the potential to capture more information and possibly make better predictions of forest AGB compared to a non-contextual sequential regression model. We let $\zhatyx$ denote generated synthetic ALS-based AGB predictions from the non-contextual sequential model, while $\zhatyxb$ denote generated synthetic ALS-based AGB predictions from the contextual sequential model. The bold font therefore specifies that both the input and the output of $g$ is an image patch (i.e., a subimage from the AOI) and not a single pixel value. For the remaining of this work, we use plain font for variables representing single pixels while a notation in bold font represents a set of pixels. 

\begin{figure}
\hspace{.5cm}
\includegraphics[scale=0.3]{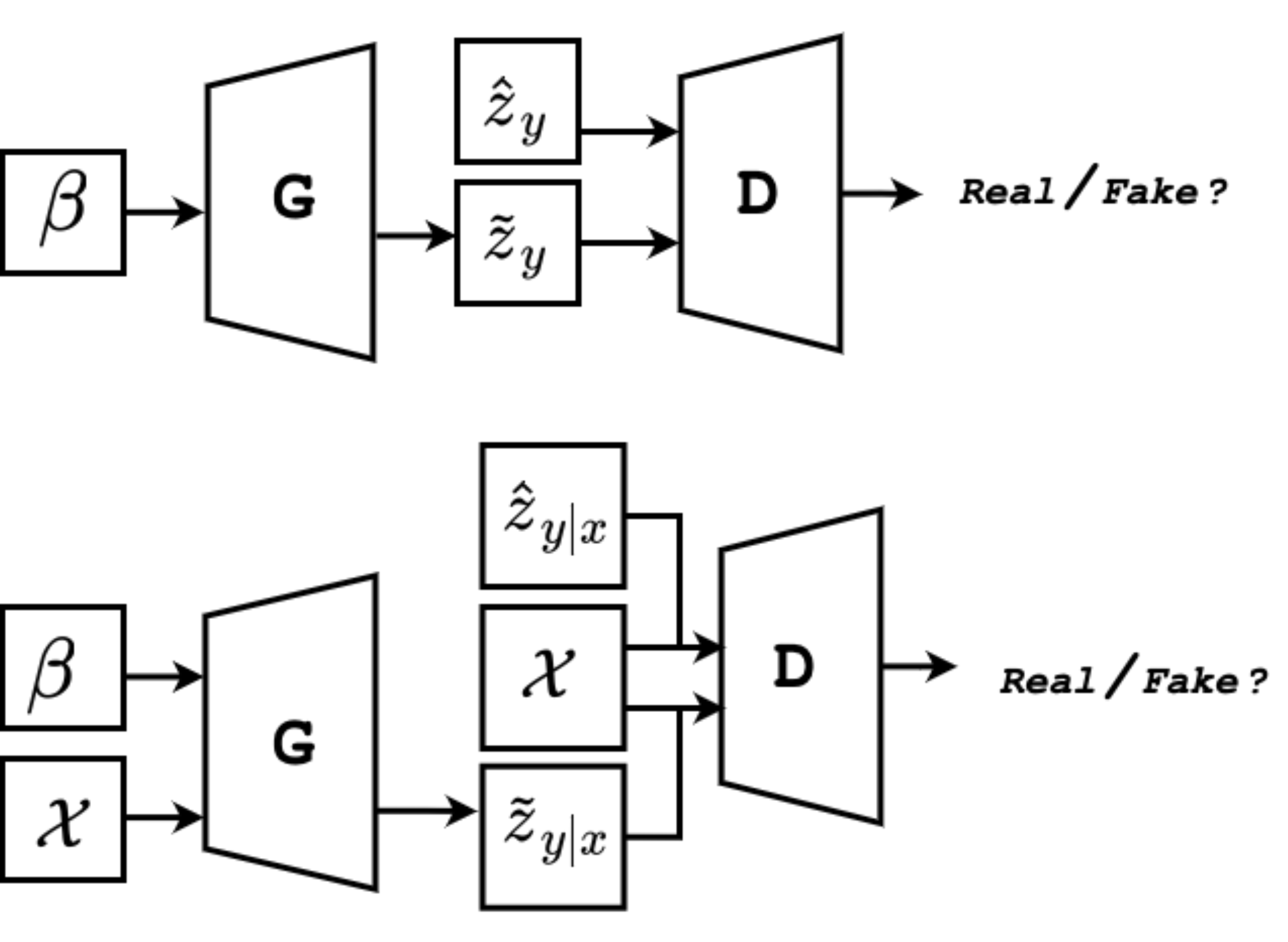}
\caption{Illustration of a GAN (upper) and cGAN (lower) model $\G$ and $\D$ denote the generator and discriminator networks. $\boldsymbol{x}$ represents images of SAR backscatter intensity from the input domain $\mathcal{X}$. ALS-based AGB predictions are in both models denoted as $\boldsymbol{\widehat{z}_y}$. The subscript $y$ indicates that the AGB predictions are retrieved from a model trained on ALS data. Generated synthetic ALS-based AGB predictions retrieved from $\mathcal{X}$ domain using a cGAN are denoted $\zhatyxb$. $\boldsymbol{\widetilde{z}_y}$ represents generated synthetic ALS-based AGB predictions retrieved from random noise, $\bm{\beta}$ in a GAN.}
\label{fig:cGAN}
\end{figure} 

\subsection{Conditional generative adversarial networks}\label{sec:cgan}
Cross-modal image translation based on generative adversarial networks (GANs) has drawn considerable attention since the architecture was proposed in 2014 \cite{gan2014}. Image translation is achieved through a generative model, referred to as the generator $\G$, that is trained to capture the data distribution of the target domain. Simultaneously, a discriminative model, referred to as the discriminator $\D$, is trained to distinguish between image samples generated by $\G$ and images from the actual target domain. The GAN components $\G$ and $\D$ are trained in a minimax optimisation procedure, where they are adapted alternatingly while seeking to optimise conflicting performance criteria. The convergence of both benefits from the battle with the adversary as long as the alternating adaption is appropriately balanced. After training, $\G$ can be utilised separately to generate data from the specific distribution. 

In the standard GAN setting, the generative model $\G$ learns a mapping from a random noise vector to a target output image, while the discriminative model $\D$ is trained to distinguish between the generated output image and the corresponding target output image. The whole process, with respect to AGB estimation, is illustrated in \figr{fig:flowchart} and the upper part of \figr{fig:cGAN}. Here, $\bm{\beta}$ denotes a random noise vector, the target output image, i.e.\ ALS-based AGB predictions, is represented by $\zhatyb$, while the generated synthetic output image is represented by $\ztildeyb$. Thus, $\ztildeyb$ represent an approximation to $\zhatyb$, generated from random noise. 

In the cGAN setting, the learned mapping to the target output image is conditioned on the distribution of an input image. Consequently, the discriminative model, $\D$, instead learns to distinguish between a real pair or false pair of images. The training process of a cGAN, with respect to AGB estimation, is shown in the lower part of \figr{fig:cGAN}. When we let the second part of the sequential model, i.e.\ \textit{g}, be represented by a cGAN model, we condition the regression model on a patch of SAR backscatter intensity data, $\x$. By the condition on SAR data, the generated synthetic output image of ALS-based AGB predictions is now  denoted $\zhatyxb$. In the cGAN setting, the aim of $\D$ is to distinguish between $\{\x, \zhatyb\}$ and $\{\x, \zhatyxb\}$.

\section{Related work}\label{sec:related}
This section frames our work within related research literature on sensor fusion with a particular emphasis on fusion between ALS and radar, traditional non-sequential regression modelling, sequential regression modelling, and image translation through GANs. 

\subsection{Traditional non-sequential regression by sensor fusion} \label{sec:lr_fusion}
In this context, we refer to traditional non-sequential regression as the conventional process of relating ground reference data of AGB directly to RS data through a single regression model. This process is illustrated on the left-hand side of \figr{fig:workflow}. Research on traditional regression models that map SAR backscatter to forest AGB has gained considerable research attention over the years. Two seminal and much-cited works from the year 1992 are the publications of Dobson \emph{et al.} \cite{dobson1992dependence} and Le Toan \emph{et al.} \cite{le1992relating}, which both investigate the dependence between forest AGB and SAR intensity data acquired with different frequencies. Since then, a natural research progression has been to investigate traditional non-sequential regression models by utilising sensor fusion, i.e.\ fusion of different RS data sources. Some popular models within traditional regression methods are linear regression, multiple linear regression and stepwise multiple regression \cite{Phua2017SynergicUse, Sinha2019MultiSensorApproach, Zhang2019DLBasedRetreival, Chen2019OptimalCombination, Li2020ForestAGBestimation, Nuthammachot2020Combineuse, sinha2021assessment} for fusion of different radar data sources \cite{sinha2021assessment}, fusion of radar and optical data \cite{Sinha2019MultiSensorApproach, CHEN2019Assessmentofmultiwave, Nuthammachot2020Combineuse, Li2020ForestAGBestimation} or fusion of ALS and optical data \cite{Phua2017SynergicUse, Zhang2019DLBasedRetreival}. 

Since \cite{dobson1992dependence} and \cite{le1992relating} published their classical statistical approaches, the possibilities of using ML and DL models for forest AGB retrieval through sensor fusion have also been investigated widely. Within these fields have fusion of radar and optical data attracted considerable attention \cite{ Chen2018EstimationofAGB, Vafaei2018ImprovingAccuracy, GHOSH2018AGBestimation, Chen2019OptimalCombination, Debastiani2019EvaluatingSAR, CHEN2019Assessmentofmultiwave, Li2020ForestAGBestimation}, but also fusion of ALS with a multitude of data sources \cite{Zhang2019DLBasedRetreival, Yang2020ANewMethod, Zhang2020AnEvaluationof, santi2020MLapplications} and fusion of different radar data sources \cite{StelmaszczukGorska2018EstimationofAGB}. Among the different ML and DL algorithms, random forest-based algorithms are some of the most popular for AGB estimation, see for example \cite{Chen2018EstimationofAGB, Vafaei2018ImprovingAccuracy, StelmaszczukGorska2018EstimationofAGB, GHOSH2018AGBestimation, Chen2019OptimalCombination, Debastiani2019EvaluatingSAR, Zhang2019DLBasedRetreival, CHEN2019Assessmentofmultiwave, Zhang2020AnEvaluationof, Li2020ForestAGBestimation}, in addition to ANNs (in particular multilayer perceptrons) \cite{Chen2018EstimationofAGB, Vafaei2018ImprovingAccuracy, Debastiani2019EvaluatingSAR, CHEN2019Assessmentofmultiwave,Yang2020ANewMethod, Zhang2020AnEvaluationof, santi2020MLapplications} and support vector machines for regression \cite{Chen2018EstimationofAGB, Vafaei2018ImprovingAccuracy, Zhang2019DLBasedRetreival, CHEN2019Assessmentofmultiwave, Zhang2020AnEvaluationof, santi2020MLapplications}. Research on pure DL methods applied to sensor fusion within traditional non-sequential regression for AGB estimation is still limited. This can probably be explained by the sparsity of ground reference data, which makes it challenging to train deep learning models. However, one example is found in the work by Zhang \emph{et al.} \cite{Zhang2019DLBasedRetreival}, where ALS data and optical Landsat 8 imagery are integrated to achieve both structural and spectral information predictors for forest AGB estimation. The DL-based model they consider is a stacked sparse autoencoder (SSAE) network, which consist of several sparse autoencoder networks (SAE), each consisting of an encoder and a decoder network. After training each individual SAE, they remove all decoder networks to establish an SSAE by stacking the remaining encoder networks layer-wise. The final SSAE regression network is obtained by adding an unspecified regression model to the end of the SSAE model. While not explicitly mentioned in \cite{Zhang2019DLBasedRetreival}, their SSAE model is a non-contextual model that operates on a single pixel level as it learns to relate RS predictor variables to single AGB measurements, retrieved from a total of 236 field plots. The SSAE network obtains the best performance in comparison with four other traditional regression models and ML models evaluated in \cite{Zhang2019DLBasedRetreival}. 

\subsection{Data fusion with sequential regression models}
In this section, we review related research that, like us, applies a modelling strategy with sequential regression models. Characteristic for this review is that it does not focus on the choice of estimation technique. We instead emphasise research on forest AGB estimation through data fusion of different types of RS data sources, which all employs a chain of two models. Common for the research we identified is that the second model exploits predictions from the first model as a dependent variable in the second modelling stage, see right-hand side of \figr{fig:workflow}. We found that research on AGB estimation applying this particular modelling strategy has been a topic in several studies from year 2008 \cite{BOUDREAU2008RegionalAGB} until today, see for example \cite{Nelson2009EstimatingQP, solbergEstimatingSprucePine2010, WANG202010EstimatingAGB, englhartAbovegroundBiomassRetrieval2011, SUN2011ForestBiomassMapping, Cartus2012MappingCanopy, TSUI2013IntegratingAirborneLiDAR, NEIGH2013TakingStock, Margolis2015CombiningSatellite, saarela2016hierarchical, HOLM2017Hybridthreephase, Kauranne2017Lidarassisted, Shao2017StackedSparseAE,  saarela2018generalized,  Urbazaev2018EstiamtionofAGB, Wang2019mappingHeight,  Qi2019ForestBiomassEstimation, Hudak2020ACarbonMonitoringSystem}. While reviewing earlier research that applies two sequential regression models in their modelling strategy, we noted a variety of terms describing the same concept in the literature. While we choose to refer to this as a sequential regression approach, we additionally found the following use of terminology for similar, but not necessary identical approaches:
\textit{two-step modelling strategy} \cite{Hudak2020ACarbonMonitoringSystem,Qi2019ForestBiomassEstimation, SUN2011ForestBiomassMapping}, \textit{two-stage regression} \cite{Kauranne2017Lidarassisted, Wang2019mappingHeight},  \textit{two-stage up-scaling method} \cite{Cartus2012MappingCanopy, Urbazaev2018EstiamtionofAGB},  \textit{two-phase estimator} \cite{Margolis2015CombiningSatellite}, \textit{two-phase (or three-phase) sampling design} \cite{Nelson2009EstimatingQP, HOLM2017Hybridthreephase},  \textit{hybrid and hierarchical model-based inference} \cite{saarela2016hierarchical, saarela2018generalized}, \textit{three-phase design} \cite{NEIGH2013TakingStock}. Additionally, \cite{WANG202010EstimatingAGB, Shao2017StackedSparseAE, TSUI2013IntegratingAirborneLiDAR, englhartAbovegroundBiomassRetrieval2011, solbergEstimatingSprucePine2010,  BOUDREAU2008RegionalAGB} also apply a modelling approach with two sequential regression models without labelling it by any particular term.
Most of the previous research that we identified focuses on relating ground reference data to ALS, and then relates ALS-derived AGB estimates to spaceborne LiDAR data \cite{HOLM2017Hybridthreephase, Margolis2015CombiningSatellite, TSUI2013IntegratingAirborneLiDAR, NEIGH2013TakingStock, Nelson2009EstimatingQP, BOUDREAU2008RegionalAGB} or a combination of different sensors \cite{Qi2019ForestBiomassEstimation, Urbazaev2018EstiamtionofAGB, Shao2017StackedSparseAE, Cartus2012MappingCanopy, englhartAbovegroundBiomassRetrieval2011, saarela2018generalized, saarela2016hierarchical}. Some others relate the ALS-derived AGB estimates to a single sensor, such as Sentinel-2 \cite{WANG202010EstimatingAGB, Wang2019mappingHeight}, Landsat \cite{Hudak2020ACarbonMonitoringSystem, Kauranne2017Lidarassisted}, GEDI Lidar \cite{Qi2019ForestBiomassEstimation}, PALSAR, \cite{SUN2011ForestBiomassMapping} or SRTM X-band radar \cite{solbergEstimatingSprucePine2010}. 

In previous research that adopts a modelling strategy with two sequential regression models, we found traditional regression models to be most common  \cite{Qi2019ForestBiomassEstimation, saarela2018generalized, saarela2016hierarchical,  HOLM2017Hybridthreephase, Margolis2015CombiningSatellite, NEIGH2013TakingStock, SUN2011ForestBiomassMapping, englhartAbovegroundBiomassRetrieval2011, solbergEstimatingSprucePine2010, Nelson2009EstimatingQP, BOUDREAU2008RegionalAGB}, such as e.g.\ \cite{englhartAbovegroundBiomassRetrieval2011}, which focuses on multiple linear regression for upscaling biomass estimates to large areas in the tropical forest of Indonesia.  Although Englhart \emph{et al.} \cite{englhartAbovegroundBiomassRetrieval2011} included neither ML nor DL, their overall idea has similarities with our modelling strategy. Their work starts by relating collected AGB sample plots to co-located ALS measurements, resulting in a regression model used to predict AGB on the whole ALS dataset. In the final stage, their second regression model relates X- and/or L-band SAR data to ALS-based AGB estimates to extend the AGB estimates to the spatial coverage of the SAR data. 

Different ML models have also been applied for AGB estimation that involves data fusion and sequential modelling. As for traditional regression, we find that random forest is one of the most commonly used ML methods, see e.g.\ \cite{Shao2017StackedSparseAE, WANG202010EstimatingAGB, Hudak2020ACarbonMonitoringSystem, Wang2019mappingHeight, Cartus2012MappingCanopy}, while e.g.\ \cite{Shao2017StackedSparseAE, Urbazaev2018EstiamtionofAGB} can be consulted for some additional examples of ML-based methods. In the intersection between traditional regression models and ML models, we also find \cite{TSUI2013IntegratingAirborneLiDAR}, which applies three different kriging methods \cite{krige1966two}:  co-kriging, regression kriging, and regression co-kriging, to extend ALS-derived biomass transects to wall-to-wall AGB maps by including L- and C-band data. 

Among research that applies a modelling strategy with two sequential regression models, we notice an absence of research using DL models for the regression task. Only one study was identified \cite{Shao2017StackedSparseAE}, which similarly to \cite{Zhang2019DLBasedRetreival} employs an SSAE for the regression task\footnote{See \Sec{sec:lr_fusion} for a discussion on the SSAE and reference \cite{Zhang2019DLBasedRetreival}.}. While \cite{Shao2017StackedSparseAE}, like us, use a sequential modelling approach to establish a relationship between ALS-derived forest biomass predictions and satellite predictors from e.g.\ \sen\ data, there are some distinct differences. Although Shao \emph{et al.} consider some contextual predictor variables, their SSAE model is a non-contextual model that only considers single pixels in the training and prediction phase. A novelty of this work is that the cGAN model lets us exploit the contextual information between neighbouring pixels through its convolutional filters. Secondly, \cite{Shao2017StackedSparseAE} adds a non-specified regression model to the end of the trained SSAE network to perform AGB predictions, as does \cite{Zhang2019DLBasedRetreival}. In our case, the cGAN model is in itself the regression model and there is no need for additional models to accomplish AGB predictions. Thus, by letting one of our proposed sequential models employ a cGAN model we contribute with new insight on how DL and RS data can be combined for AGB prediction. 

\subsection{Image translation with generative adversarial networks}
Image to image translation is the task of translating a representation of the imaged scene into another. Examples of this process can, for example, be to translate from a greyscale representation into an RGB image or translating an aerial photo into a map view of the same area \cite{isola}. In such a translation process, the $\G$ network is commonly conditioned on the first representation, i.e.\ the input signal or distribution, to achieve better translation. This makes the cGAN and the \textit{Pix2pix} architecture \cite{isola}, as one specific example, better suited for this task than a generator network conditioned on noise, as the traditional GAN \cite{digan}. In this work, we choose to condition the $\G$  network on SAR measurements of the backscatter coefficient in the same area, from which we wish to generate ALS-based biomass prediction maps. 
  
Research on RS data simulation through image translation can be found in e.g.\ \cite{Review2_suggestion, sarsim, digan}. Li \emph{et al.} \cite{Review2_suggestion} focus on change detection (CD) and propose a GAN-based deep translation network for translation between SAR and optical images. By translating images from both sensors into a common feature domain, image characteristics from both images become comparable and can aid the network in the CD task. Ao \emph{et al.} \cite{digan} proposed a framework for translation between different SAR sensors. By condition their dialectical GAN on urban input images from the low-resolution (LR) Sentinel-1 sensor they enable generation of corresponding high-resolution TerraSAR-X images. The dialectical GAN uses a modification of the \textit{Pix2pix} cGAN proposed in \cite{isola} and combines concepts of both the cGAN and traditional neural networks. In \cite{sarsim} are Bao \emph{et al.} considering three non-conditional GAN networks to simulate SAR data of vehicles from random noise. While \cite{Review2_suggestion} focuses on translating between instruments with different physical measurement principles, does neither of  \cite{Review2_suggestion, sarsim, Review2_suggestion} focus on using image translation through GANs for regression purposes as we intend to. 

In general, most of today's research on semi-supervised learning through GANs focuses on solving a classification task, see e.g\ \cite{Rev1Suggestion} which propose the DLR-GAN architecture to perform low resolution (LR) image classification. To improve classification on this challenging task they propose to let the $\G$ network learn to recover the LR components and the high-frequency components of the LR image. Only a very very few studies were identified that apply their architecture to regression tasks \cite{reggan}. Within the GAN literature, Rezagholizadeh and Haidar presented one of the first models aimed at regression, named the Reg-GAN \cite{reggan}. They use two different networks, where one learns data generation while the other predicts continuous labels. It is applied in a computer vision task for self-driving vehicles, where the GAN generates images of a road segment and a regression network predicts the matching steering angle. Olmschenk  \emph{et al.}  \cite{olmschenk} later proposed the feature contrasting loss function and outperformed \cite{reggan} on the same semi-supervised GAN regression task. Additional examples were also shown in \cite{olmschenk} on the combined task of face generation and age prediction as well as on crowd counting. The proposed work in our paper differs from earlier related research \cite{reggan, olmschenk}, as we do not perform any additional regression on the image content of the generated synthetic patches. This is possible due to the nature of our proposed modelling strategy with two sequential regression models, which results in a cGAN-based model that is able to make predictions in new unseen areas through the image translation. 

\section{Material and Methodology}
\label{sec:method} 
The related work presented in \Sec{sec:related} positions our work with respect to published research in related areas. Based on this literature survey, and previous published research on AGB estimation in the AOI, we  make the following methodological contributions: 
\begin{enumerate}
\item By proposing our \sen-based non-sequential AGB regression model, we extend the work of Næsset \emph{et al.}\  \cite{naessetMappingEstimatingForest2016}.
\item The two proposed sequential models extends previous work on sensor fusion in the AOI. Furthermore, by introducing the DL-based sequential model, this work also contribute with novel insight on the possibilities for AGB prediction by using DL models for sensor fusion.  These deep neural networks have convolutional layers that extract contextual spatial information, which has been exploited both in other types of regression problems \cite{ren2015faster} and also for AGB prediction \cite{Luofan206ApplicationOf}, but not in  a sequential regression approach to upscaling and information enhancement.
\item The proposed method applies image translation to truly heterogeneous images and domains in a regression context. Similar image translation has previously been done for general purposes \cite{Fuentes2019SARtoOptical} and within image analysis tasks like change detection \cite{Luppino2021DeepImage}, but is new in the biomass estimation and regression setting.
\end{enumerate}

We accomplish the mentioned novelties in 2) and 3) for the DL-based sequential model by using a modification of the Pix2Pix image translation architecture \cite{isola} to generate synthetic yet realistic ALS-based AGB predicted maps with SAR intensity data as input. We refer to the Appendix, i.e.\ \Sec{sec:mod_pix2pix}, for a list over these modifications and their motivation.

We will in the following describe the datasets used in this paper, the preprocessing steps applied to the data, and give an overview of the different models we consider. 

\subsection{Study area and dataset description}
\label{sec:dataset}
\begin{figure}
    \centering
    \includegraphics[scale=0.7, height=6cm]{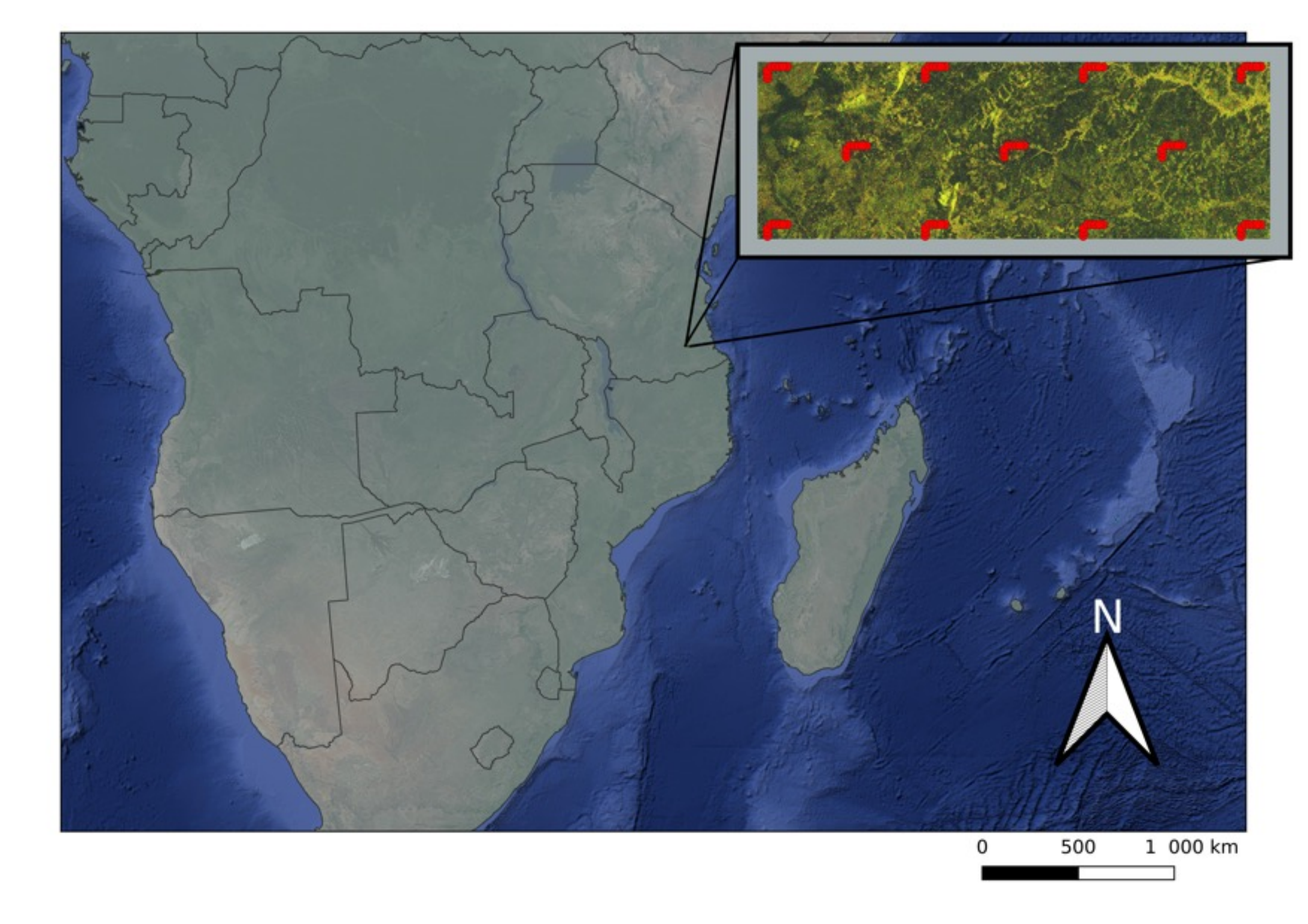}
    \caption{The location of a subset of the \sen\ scene, as  well as the location of the ground reference plots (in red) in the country of Tanzania.} 
    \label{fig:map}
\end{figure}

\subsubsection{Study area}
The AOI is a rectangular area with size $11.25 \times 32.50$ km (WGS 84/UTM zone 36S), located in the Liwale district in southeast Tanzania (\ang{9}52'-\ang{9}58'S, \ang{38}19'-\ang{38}36'E). \figr{fig:map} shows the relative location of the AOI in Tanzania. The Liwale district experiences two rain periods each year: a shorter period from late November to January and a longer period from March to May. Liwale's main dry season occurs between July to October. The miombo woodlands of the Liwale districts is characterised by a large diversity of tree species, with \textit{Brachystegia} sp.\ \textit{Julbernadia} sp.\ and \textit{Pterocarpus angolensis} being the most dominant ones \cite{naessetMappingEstimatingForest2016, ene2016, ene2017}.

\subsubsection{Field data}
The field data used in this work, from now on referred to as AGB ground reference data or $z$, were collected within 88 field plots during January-February 2014 \cite{naessetMappingEstimatingForest2016}. These field plots were distributed in groups of eight in each of the 11 L-shaped clusters, shown with red dots in the \sen\ scene in Fig.~\ref{fig:map}. The sample plots are circular, each of size 707m$^2$, i.e.\ they have a radius of 15 m. We refer to \cite{Tomppo2014ASampling} for a thorough work on the national level sampling design for Tanzania, and to e.g.\ \cite{ene2016, ene2017, naessetMappingEstimatingForest2016} for reference work on e.g.\ the use of field data in the AOI for large-scale AGB estimation. Measured AGB in the AOI ranged from 0 to 213.4 $\mathrm{Mg\,ha}^{-1}$ \cite{naessetMappingEstimatingForest2016}.

\subsubsection{ALS-based AGB data} \label{sec:als}
The ALS data were acquired in 2014, see \cite{naessetMappingEstimatingForest2016, ene2016} for details of this process.
N\ae sset \emph{et al.} trained in \cite{naessetMappingEstimatingForest2016} a regression model on the ALS data to make ALS-based AGB predictions on a grid with square pixels of size 707m$^2$. Their model, referred to as $f$, is the first regression model in our proposed modelling strategy with two sequential regression models.  
The output from the ALS-based regression model in \cite{naessetMappingEstimatingForest2016}, i.e.\ ALS-based AGB predictions, $\widehat{z}_y$, was made available for this work. These ALS-based AGB predictions will serve as a surrogate for the AGB ground reference data in the second regression model $g$, when SAR data is used with either a traditional regression model or a cGAN model for image translation to upscale the ALS-based AGB predictions. See right-hand side of \figr{fig:workflow} for an illustration of the sequential modelling strategy with notation. 

\subsubsection{SAR data}\label{sec:sar}
Our SAR data consists of a C-band SAR scene obtained from the \sen sensor, which provides data in two bands, i.e.\ the VV and VH polarisation. This sensor was chosen since an AGB model trained on data from this sensor meet most of the needs listed in \Sec{sec:probdef}; the data is frequently updated, it has extensive spatial coverage and is freely available. For this paper, we choose a \sen\ scene acquired on 15 September 2015, as it fulfils three additional criteria: 1) it covers our area of interest, 2) it is closest in time to acquisition of the ALS data, and 3) it was acquired during one of the area's two yearly dry seasons. The latter implies that the scene achieves optimal sensitivity to dynamic AGB levels. We initially aimed to create a multitemporal stack of \sen\ scenes, but as only one scene meets all the three additional criteria, we had to settle for this single scene. The SAR data are obtained in a high-resolution Level-1 ground range detected (GRD) format, with a pixel size of 10 m. It was downloaded from Copernicus Sentinel Scientific Data Hub\footnote{See \url{https://scihub.copernicus.eu/dhus/#/home}}. Fig.~\ref{fig:map} visualises the scene and indicates its relative location in Tanzania. 

\subsection{SAR data processing and preparation of datasets}\label{sec:preprosar}
To process the \sen GRD product, we used the ESA SNAP toolbox \cite{snap} and followed the workflow suggested in \cite{Filipponi2019Sentinel1GRDPreprocessing} with some modifications. The final processing workflow is summarised as:
\begin{enumerate}
    \item apply orbit file,
    \item thermal noise correction,
    \item border noise removal,
    \item calibration, 
    \item range Doppler terrain correction (bilinear interpolation),
    \item (conversion to dB). 
\end{enumerate}
We also experimented with speckle filtering, using a refined Lee filter \cite{refinedlee} with the SNAP default window size of $7 \times 7$ as an optional additional processing step between step 4) and step 5). However, since models trained on speckle filtered \sen\ data experience higher variations in AGB predictions than models trained without speckle filtered \sen\ data, we decided to omit speckle filtering in the processing workflow. See \Sec{exp:data} in the Appendix for details. Step 6) was only applied to the cGAN-based sequential regression model. We provide an investigation of the impact that \sen\ data on dB scale or linear scale have on AGB predictions for cGAN-based models in the Appendix, see \Sec{sec:lindb}.
During step 6) for the data used in the cGAN-based regression model or after step 5) for the two other models, we also applied the same map projection as in \cite{naessetMappingEstimatingForest2016}, i.e.\ WGS 84/UTM zone 36S, to make sure that the \sen\ dataset and the ALS-based AGB prediction dataset are aligned. 

After performing the above processing steps, our \sen\ dataset was further processed in QGIS \cite{qgis}. In QGIS, we first reprojected the \sen\ dataset to the same projection that the ALS-based AGB grid pixel dataset used in \cite{naessetMappingEstimatingForest2016}. Then, cubic convolution resampling was applied to resample the pixel size of the \sen\ dataset from its original pixel spacing of 10 m$\times$10 m to the same pixel size as the grid pixels of the ALS-based AGB predictions, i.e.\ 26.6 m $\times$ 26.6 m. As a final step, a subset of the entire \sen scene corresponding to the extension of the ALS-based AGB data was extracted.

\begin{figure}[htb]
  \includegraphics[width=\columnwidth,height=2.5cm]{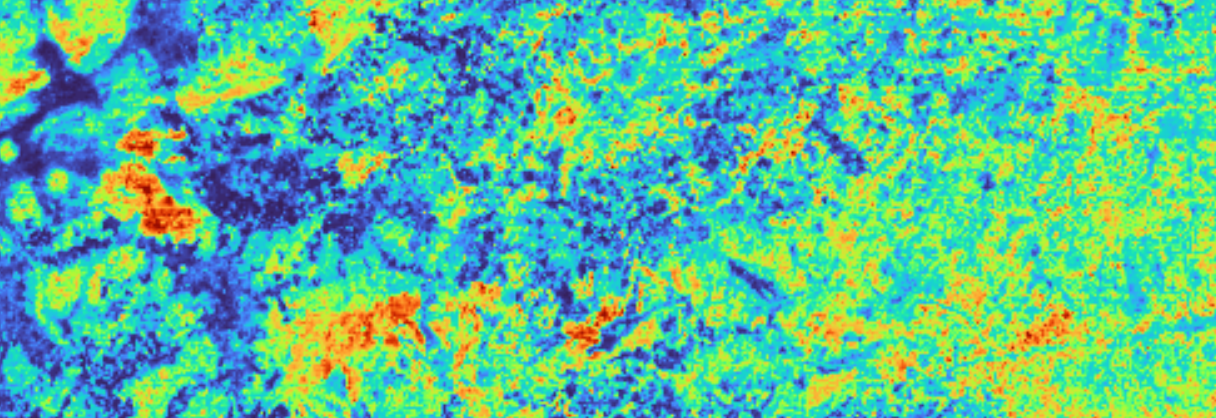}
\vspace{0.2cm}
  \includegraphics[width=\columnwidth, height=2.5cm]{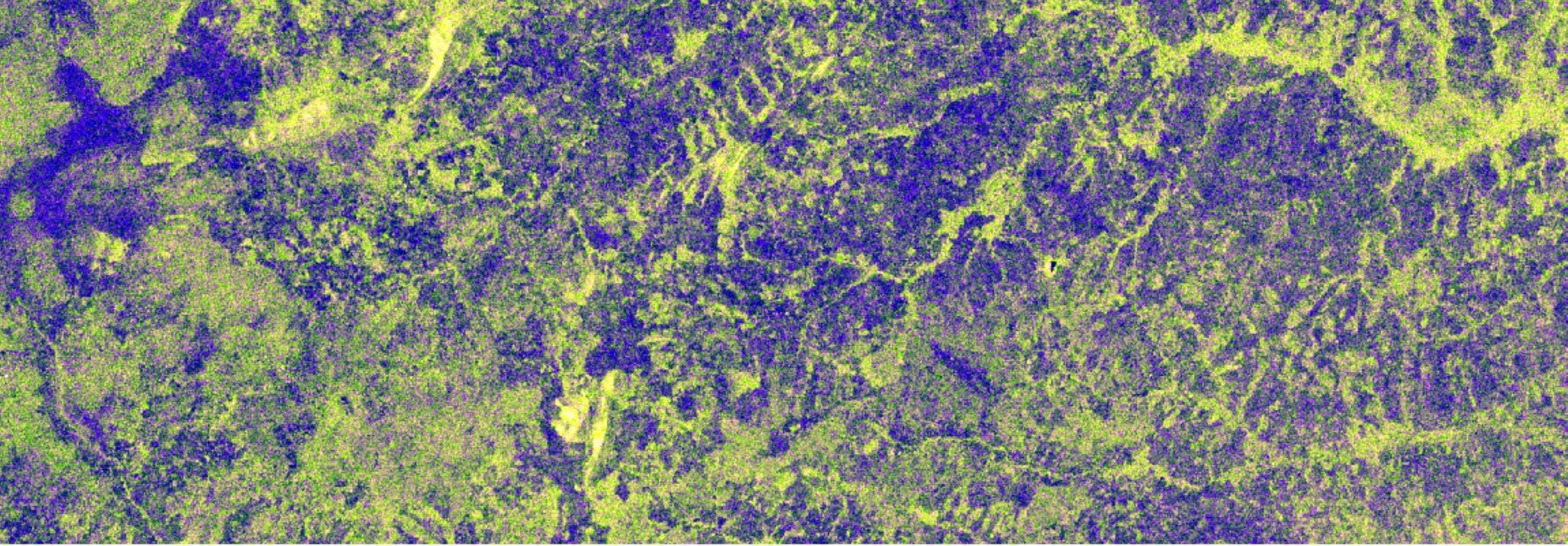}
\caption{\textbf{Top row:} ALS-based AGB predictions from \cite{naessetMappingEstimatingForest2016}. \textbf{Bottom row:} False-colour image of the \sen dataset.} 
\label{fig:lidarsen}
\end{figure}

For the image-to-image translation task, i.e. the cGAN-based model $g$, a false-colour image was created from the processed \sen dataset. This was done since the chosen cGAN architecture, Pix2Pix, requires three-channel RGB images or greyscale images as input. The false-colour image was created as follows: red = VV, green = VH, and blue = VV-VH. The ALS-based AGB prediction dataset was kept as a greyscale image as each grid pixel in the dataset only consist of one feature, i.e.\ an AGB prediction. Fig.~\ref{fig:lidarsen} shows the ALS-based AGB prediction dataset and the corresponding false-colour \sen\ scene after performing all processing steps with the ESA SNAP toolbox and QGIS. For illustrative purposes, we choose to show the ALS-based AGB prediction dataset of \figr{fig:lidarsen} in pseudo-colours, where dark blue pixels indicate biomass closer to 0 $\mathrm{Mg\,ha}^{-1}$ while green through yellow to red pixels indicates increasing biomass content ($\mathrm{Mg\,ha}^{-1}$).

\subsection{ Traditional Sentinel-1A-based AGB regression models}

In \cite{naessetMappingEstimatingForest2016} several different models were explored to construct traditional non-sequential regression models for AGB relating different remotely sensed datasets and the 88 field plots. They settled for a model with square root transformation of the response variable for ALS, RapidEye, Landsat and PALSAR, since this model performed equally well as more complex models and since it always predicts values $>0$. Inspired by their findings, we develop a similar baseline non-sequential regression model for AGB between \sen\  and the same 88 field plots accordingly to
\begin{equation}
    E\big[\sqrt{AGB}\big] = \alpha_0 + \Sigma_{j=1}^J\alpha_jx_j\,,
    \label{eq:sqrtagb}
\end{equation}
where $\alpha_0$ is the intercept, i.e.\ a constant, $\alpha_j$ are regression coefficients and $x_j$ are explanatory variables. We followed the procedure in \cite{naessetMappingEstimatingForest2016} and performed ordinary least-squares regression with step-wise forward selection of the variables. Our inclusion criteria focus on variables being significant at 5\% level using an F-test. For the \sen\ product, VH and VV backscatter coefficients on a linear scale plus square and square root transformations of these variables were subject to the step-wise selection. We follow the procedure from \cite{naessetMappingEstimatingForest2016} and correct for bias when transforming our model to arithmetic scale in accordance with \cite{gregoire2008regression}:
\begin{equation}
    \widehat{AGB} = \big(\hat{\alpha}_0 + \Sigma_{j=1}^J\hat{\alpha}_jx_j\big)^2 +MSE,
    \label{eq:agb}
\end{equation}
where MSE is the mean square error computed from the fitted model on square root form, i.e.\ from \eq{eq:sqrtagb}.
\newline

\subsection{cGAN-based AGB regression models}

\begin{table}[]
    \caption{Symbols and notation introduced in \Sec{sec:probdef} and used throughout the paper for the different datasets, in non-sequential modelling, sequential modelling, the GAN and the cGAN model. Notation in plain font indicate variables represented by single pixels, while notation in \textbf{bold} font indicate variables represented by image patches consisting of a pixel neighbourhood.}
    \centering
    \begin{tabular}{c l}
    \hline
        $\boldsymbol{\beta}$  &  Noise vector, input to the $\G$ of a GAN/cGAN \\
        $\D$ & Discriminator network of a GAN/cGAN \\
        $\G$ & Generator network of a GAN/cGAN\\
        $\mathcal{X}$ & Represent the input domain, SAR data \\
        $\mathcal{Y}$ & Represent the domain of ALS data \\
        $\mathcal{Z}$ & Represent the domain of AGB data \\
        $\widehat{z}_{x}$ & SAR-based AGB predictions, $\widehat{z}_{x}\in \mathcal{Z}$ \\
        $\ztildeyb$ & A patch of generated synthetic ALS-based AGB predictions\\
        & from a GAN. Retrieved from $\boldsymbol{\beta}$ data, $\ztildeyb \in \mathcal{Z}$\\
        $\zhaty, \zhatyb$ & ALS-based AGB predictions,
        $\zhaty, \zhatyb \in \mathcal{Z}$\\
        $\zhatyx$ & Generated synthetic ALS-based AGB predictions from the\\
                            &baseline sequential regression model, trained with $x$ data\\
                            &(SAR data) as the regressor. $\widehat{z}_{y|x}\in \mathcal{Z}$\\
        $\zhatyxb$ & A patch of generated synthetic ALS-based AGB predictions\\
                            &from a cGAN. Retrieved from  $\x$ data (SAR data), $\zhatyxb\in \mathcal{Z}$\\
        $f$ & A regression function between $y$ data and $z$ \\
        $g$ & A regression function between $x$ data and $\zhaty$\\
        $h$ & A regression function between $x$ data and $z$ \\
        $x, \x$ & Data from the SAR sensor, i.e. $x \in \mathcal{X}$ \\
        $y, \y$ & Data from the ALS sensor, $y\in \mathcal{Y}$ \\
        $z$ & Ground reference AGB data, $z\in \mathcal{Z}$\\
        \hline
    \end{tabular}
    \label{tab:notation}
\end{table}

This section formally introduces some popular choices of objective functions, the generator network, and the discriminator network of a cGAN, with a special focus on the Pix2Pix architecture \cite{isola}. We also relate the cGAN framework to model $g$ in our sequential modelling strategy by using the same notation that was introduced in \Sec{sec:probdef}. See \tabref{tab:notation} for a summary of the notation, and \figr{fig:workflow} and \figr{fig:cGAN} for illustrations of how the different entities of \tabref{tab:notation} are used in the sequential modelling approach or in the cGAN network.

In our application the input domain consists of image patches from the \sen\ scene, and the output domain of corresponding image patches from the ALS-based AGB wall-to-wall map. Thus, conditioned on images from the input domain, $\x \in \xcal$, the generator network $\G$ of the cGAN aims to capture the data distribution of the output domain $\zhatyb \in \zcal$, by generating corresponding synthetic image samples $\zhatyxb \in \zcal$. Image pairs are then presented to the discriminator network $\D$ of the cGAN, which aims to distinguish if it is presented with a real pair of images, $\{\x, \zhatyb\}$, or a fake pair, $\{\x, \zhatyxb\}$. The whole training process of a cGAN is illustrated in the lower part of \figr{fig:cGAN}.
As $\G$ aims to fool $\D$, its ultimate goal is to obtain $\zhatyx \approx \zhaty \approx z$, where $\zhatyx, \zhaty, z \in \zcal$. In words: At the position of each single AGB ground reference measurement, the generated synthetic ALS-based AGB predictions should resemble both $z$ and the ALS-based AGB predictions well on a pixel basis. During adaption of the cGAN, both $\G$ and $\D$ are trained simultaneously to outperform each other, resulting in the following minimax objective function \cite{gan2014}:
\begin{equation}
\begin{split}
 \min_{\G} \max_{\D} V(\D,\G)  =  & \mathbb{E}_{\x,\zhatyb}[\text{log} \D(\x,\zhatyb)]  + \\
& \mathbb{E}_{\x}[\text{log}(1 - \D(\x,\G(\x))].
\end{split} 
\label{eq:vanilla}
\end{equation}
A cGAN network trained with the objective function in \eq{eq:vanilla} is referred to as a Vanilla GAN. The least squares generative adversarial network (LSGAN) was proposed to overcome issues with stability during training of the Vanilla GAN \cite{lsgan}. Its objective functions in a conditional setting are  
\begin{equation}
\begin{split}
 \min_{\D}V_{LSGAN}(\D)  =  & \frac{1}{2}\mathbb{E}_{\x,\zhatyb}[(\D(\x,\zhatyb)-b)^2]  + \\
& \frac{1}{2}\mathbb{E}_{\x}[( \D(\x,\G(\x))-a)^2]\\
 \min_{\G}V_{LSGAN}(\G)  =  & \frac{1}{2}\mathbb{E}_{\x}[( \D(\x,\G(\x))-c)^2],
\end{split} 
\label{eq:lsgan}
\end{equation}
where $a$ and $b$ are labels for fake and real data, while $c$ denotes a value that $\G$ tricks $\D$ to believe for fake data \cite{lsgan}. 
Introduced in \cite{wgangp} for further stabilisation of training and high quality image generation, we also consider the Wasserstein GAN with gradient penalty (WGAN-GP). It considers real data, simulated data and a combination of these in its objective function, which in the conditional setting has the following form \cite{wgangp}:
 \begin{equation}
  \begin{split}
 \min_{\G} \max_{\D} V(\D,\G)  = & \mathbb{E}_{\x}[ \D(\x,\G(\x))] - \\
\mathbb{E}_{\x,\zhatyb}[\D(\x,\zhatyb)] & + \lambda \mathbb{E}_{\hat{\z}}[(||\nabla_{\hat{\z}}\D(\hat{\z})||_2-1)^2] \,
\end{split}   
\label{eq:wgangp}
 \end{equation}
with
\begin{equation}
    \hat{\z} = \epsilon\zhatyb + (1-\epsilon)\G(\x)\,.
\end{equation}
$\zhatyb$ in \eq{eq:vanilla}, \eq{eq:lsgan} and \eq{eq:wgangp} denotes a real ALS-based AGB image patch from the $\zcal$ domain while $\G(\x) =\zhatyxb$ represents a generated synthetic image patch.  

\subsubsection{Generator network} Three different $\G$ networks were tested, all based on the ResNet model \cite{resnet}: ResNet-4, ResNet-5, and ResNet-6. ResNet-6 is a part of the original Pix2Pix architecture \cite{isola} and consists of 2 encoder blocks followed by 6 residual blocks and 2 decoder blocks.
ResNet-4 and ResNet-5 consist of the same number of encoder-decoder blocks as ResNet-6,  but only 4 and 5 residual blocks, respectively. The two smaller networks were proposed as we work with small image patches of $64\times64$ pixels, see \Sec{sec:cganseq}. 

\subsubsection{Discriminator network}\label{sec:D}
In \cite{isola} Isola \emph{et al.} evaluate different variations of the neural network discriminator architecture by varying the patch size \textit{N} of the discriminator receptive fields from a $1\times 1$ \textit{PixelGAN} to an $N \times N$ \textit{PatchGAN}. 
Since we work with fairly small image patches in number of pixels we decide to settle for the following three discriminator networks: 
\begin{itemize}
    \item a $1\times 1$ \textit{PixelGAN},
    \item a $16 \times 16$ \textit{PatchGAN},
    \item a $34 \times 34$ \textit{PatchGAN}.
\end{itemize}
The two \textit{PatchGAN} networks were designed by adjusting the depth of the GAN discriminator to obtain a receptive field of $16 \times 16$ or $34 \times 34$, respectively. In a \textit{PixelGAN} the discriminator tries to classify each $1\times 1$ pixel in the image patch as real or fake, while for the two \textit{PatchGAN} networks the discriminator tries to classify each $N\times N$ patch of pixels in the image patch as real or fake. The discriminator network is applied across an image patch in a convolutional matter during the discriminator phase to produce several classification responses. Eventually, all responses are averaged to provide the discriminator output with a real or false decision. Thus, for each image patch pair,$\{\x, \zhatyb\}$ or $\{\x, \zhatyxb\}$, $\D$ outputs a binary prediction, based on $\D$'s belief of the input pair. Optimally, we wish $\D$ to predict a fake pair when the image par consists of an image patch from $\x$ and another from $\G(\x)$, i.e. $\{\x, \zhatyxb\}$. 

\begin{table*}[htp]
    \caption{Pearson correlation coefficient, R, RMSE, leave-one-out cross-validation RMSE (LOOCV RMSE) and mean absolute error (MAE) computed between ground reference plots of AGB, $z$, and area-weighted means of predicted AGB from either the five non-sequential regression models \cite{naessetMappingEstimatingForest2016} or our \sen-based non-sequential regression model. All units are in $\mathrm{Mg\,ha}^{-1}$.} 
    \centering
    \begin{tabular}{l|l|c|l|l|c|l}
    \hline
    Auxiliary data source & Modelling approach & Model & R & RMSE & LOOCV RMSE & MAE\\
    \hline
    ALS$^{a}$ & Non-sequential (traditional) & $^c$ & 0.68 & 33.39 & $^c$ & 24.61\\
    InSAR$^a$ & Non-sequential (traditional) & $^c$ & 0.49 & 40.40 & $^c$ & 29.44\\
   RapidEye$^a$ & Non-sequential (traditional) & $^c$ & 0.61 & 36.21 & $^c$ & 26.76\\
    Landsat$^a$ & Non-sequential (traditional) & $^c$ & 0.33 & 43.03 & $^c$& 33.10\\
    PALSAR$^a$ & Non-sequential (traditional) & $^c$&0.27  & 43.96 & $^c$ & 33.18\\
    \sen$^{b}$ & Non-sequential (traditional) &  $\widehat{AGB} = \big(2.96 +41.60VV\big)^2 +10.51$ & 0.54 & 38.52 & 39.6 & 30.04\\
    \hline
    \multicolumn{6}{l}{}\\
    \multicolumn{6}{l}{${}^\text{a}$ Indication of which remote sensed data source that were used in \cite{naessetMappingEstimatingForest2016} to train traditional their non-sequential regression models.}\\
    \multicolumn{6}{l}{${}^\text{b}$ The traditional non-sequential regression model developed between \sen\ and AGB reference data.} \\
    \multicolumn{6}{l}{${}^\text{c}$ See \cite{naessetMappingEstimatingForest2016} for reference to specific models and computed LOOCV RMSE.}\\
    \end{tabular}
    \label{tab:res_trad}
\end{table*}

\section{Experiments and results}\label{sec:res}
In this section, the proposed \sen-based regression models for AGB prediction are presented: the non-sequential regression model, the baseline sequential regression model and the cGAN-based sequential regression model. The performance of the proposed models is evaluated by comparing predicted AGB to AGB ground reference data and the constructed AGB prediction maps to each other, and the AGB prediction maps of \cite{naessetMappingEstimatingForest2016}. Qualitative and quantitative results are provided. We keep the notation introduced in \tabref{tab:notation} and let $z$ denote ground reference AGB data, $\zhatx$ denotes AGB predictions from the \sen-based non-sequential regression model, $\zhaty$ denotes AGB predictions from the non-sequential ALS-based model \cite{naessetMappingEstimatingForest2016}, $\zhatyx$ denotes either generated synthetic ALS-based AGB predictions from the baseline sequential  regression model or single predictions from the cGAN-based sequential regression model. In contrast, $\zhatyxb$ denotes a patch of predictions from the cGAN-based sequential regression model. We refer to the \sen-based non-sequential regression model as $h$, the ALS-based non-sequential regression model from \cite{naessetMappingEstimatingForest2016} as $f$ and either of the sequential models, i.e.\ the baseline traditional sequential regression model or the cGAN-based sequential regression model, as $g$.

\begin{figure}[htb]
 \hspace*{0.8cm}
 \subfloat[\label{fig:als_scatter}]{\includegraphics[width=0.44\columnwidth]{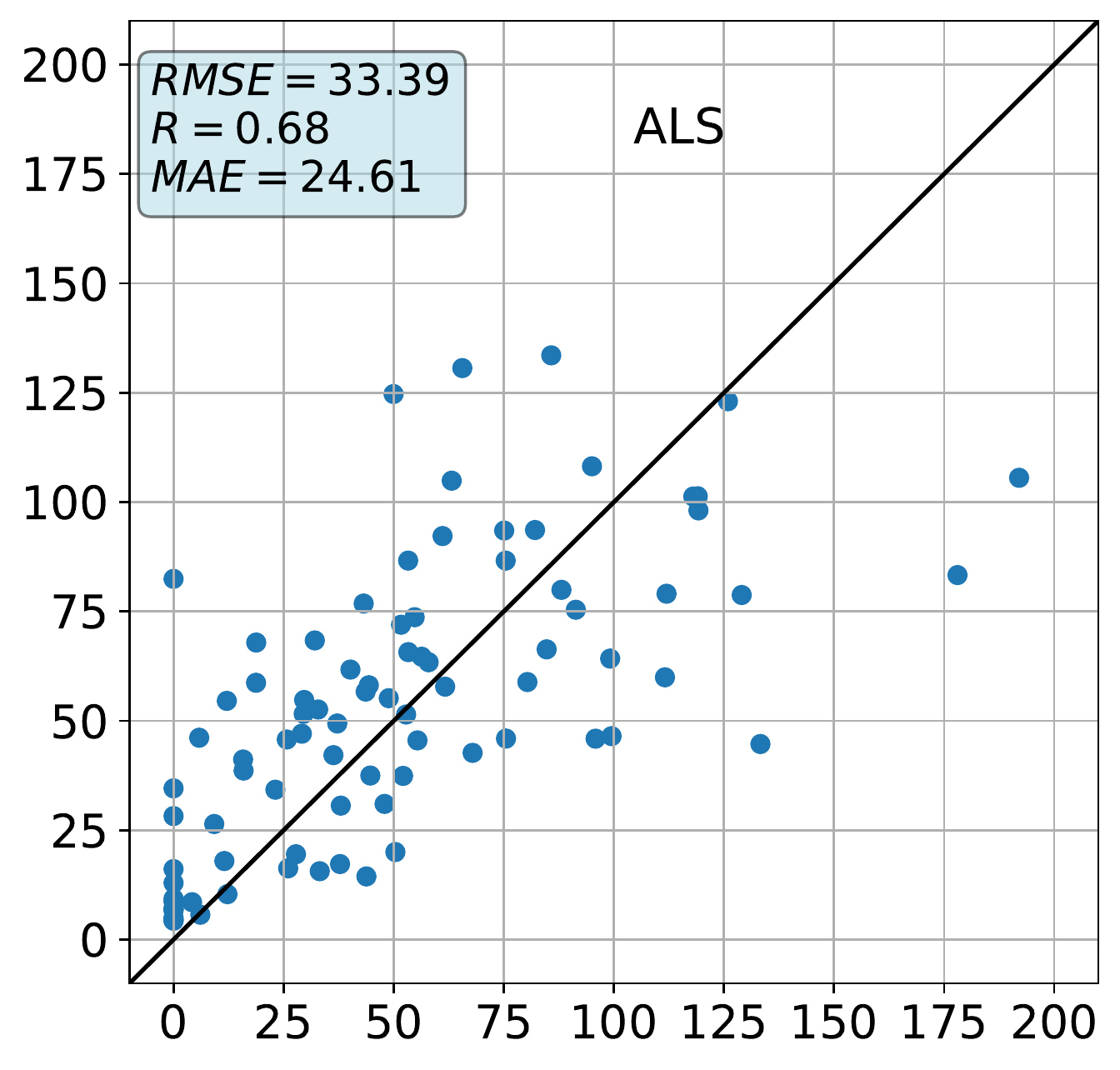}}
\subfloat[\label{fig:insar_scatter}]{\includegraphics[width=0.44\columnwidth]{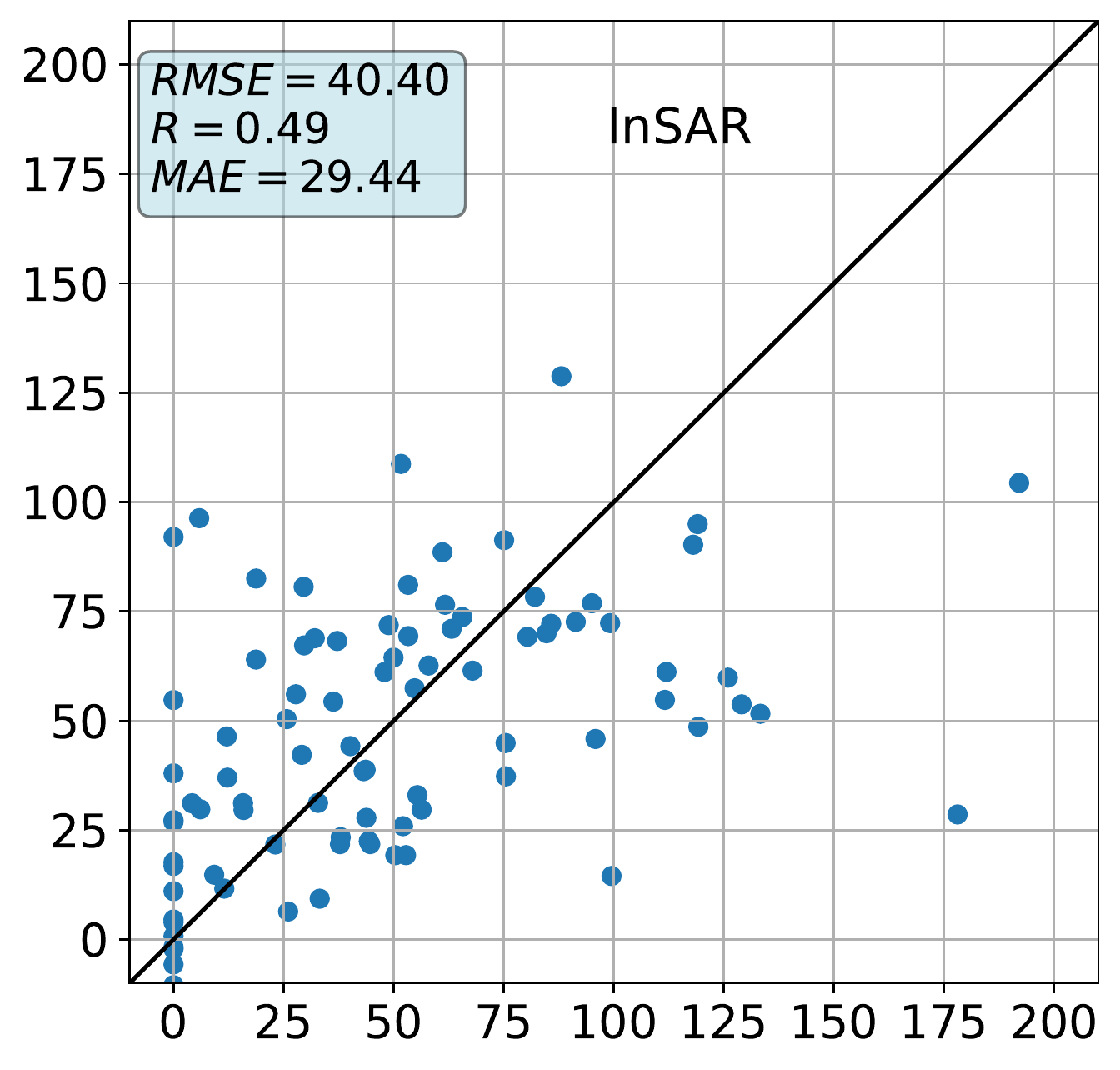}}\\
 \hspace*{0.8cm}
\subfloat[\label{fig:re_scatter}]{\includegraphics[width=0.44\columnwidth]{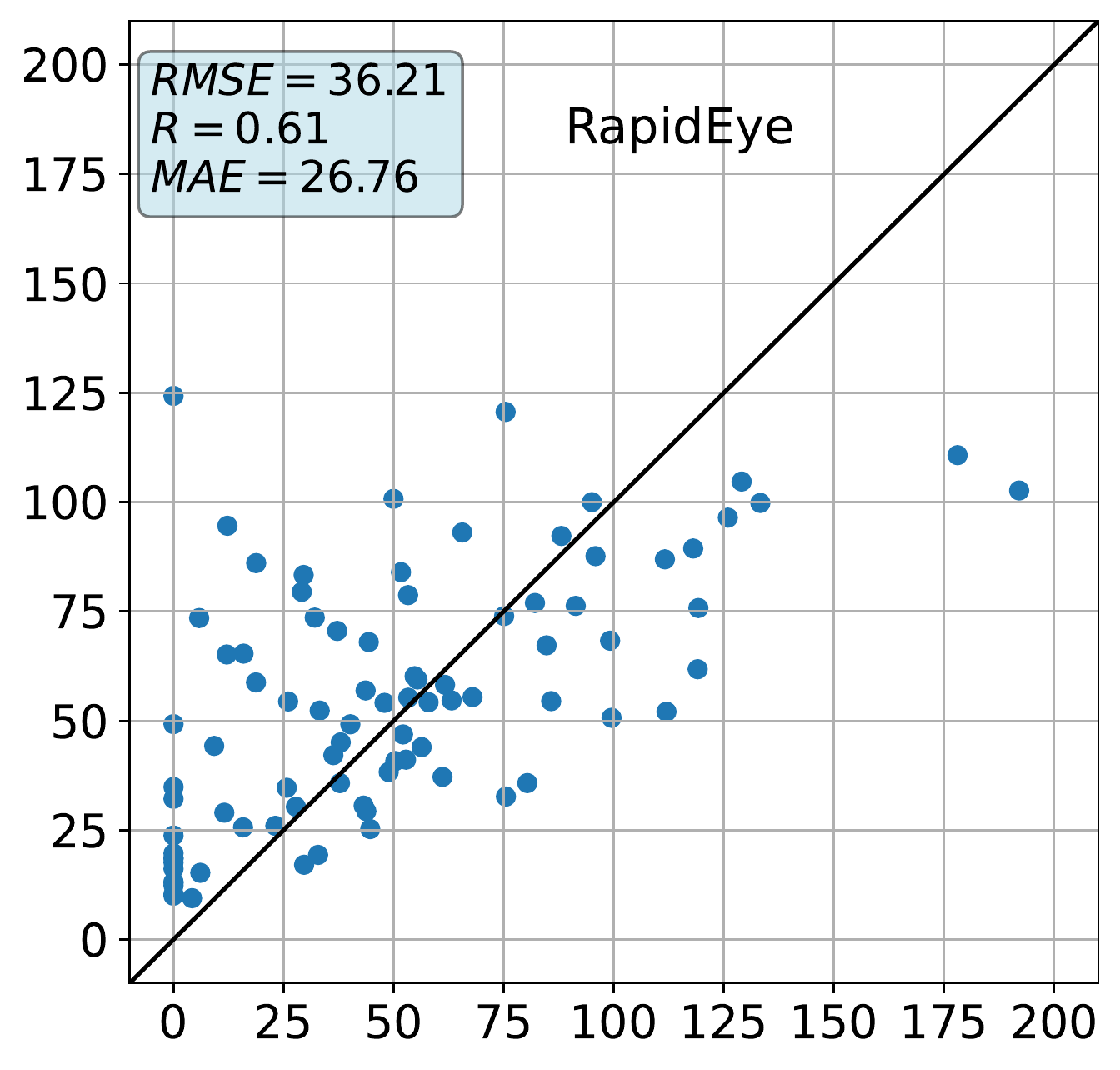}}
\vspace*{-.2cm}
 \subfloat[\label{fig:landsat_scatter}]{\includegraphics[width=0.44\columnwidth]{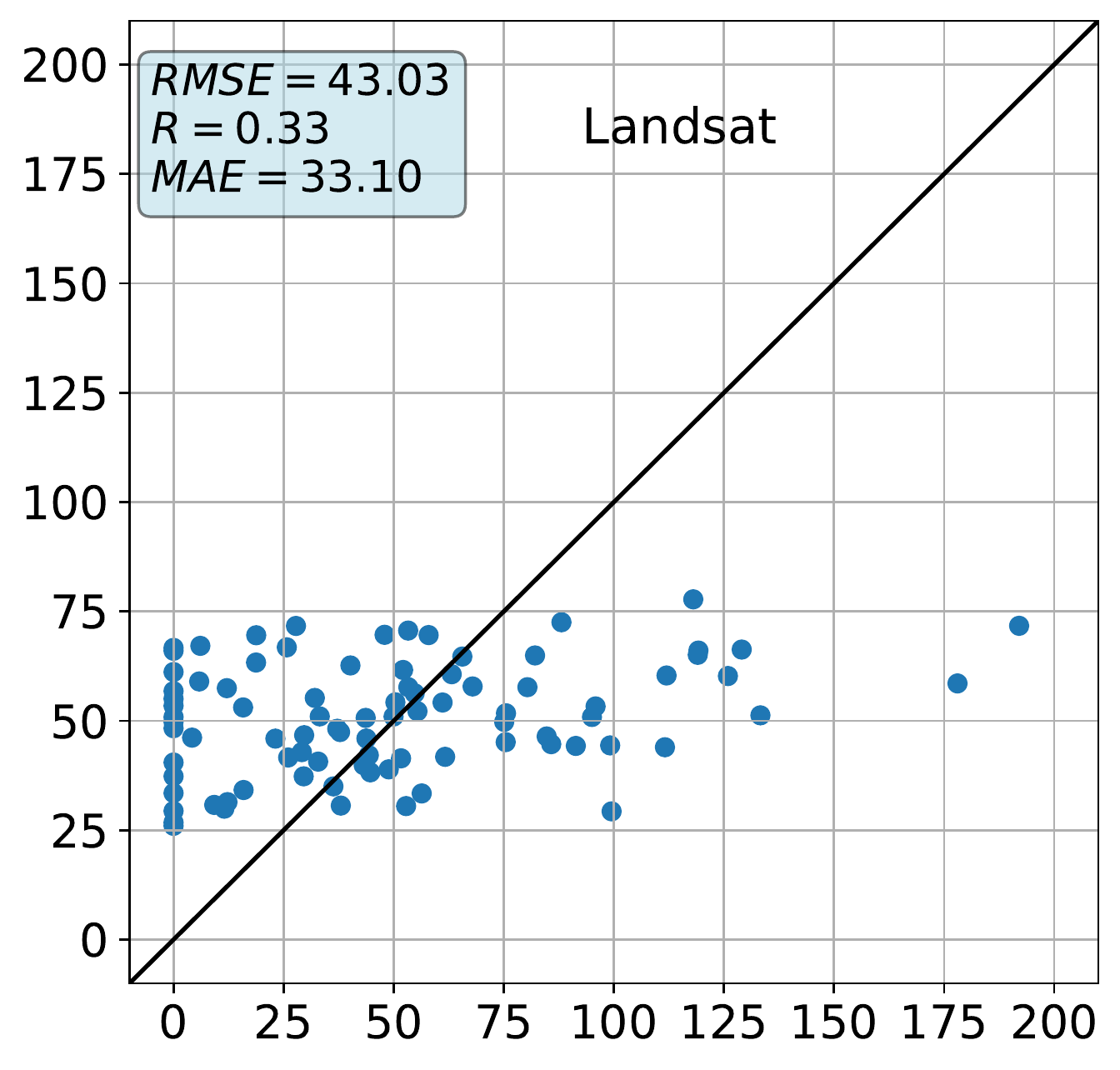}}\\
 \hspace*{0.8cm}
\subfloat[\label{fig:alos_scatter}]{\includegraphics[width=0.44\columnwidth]{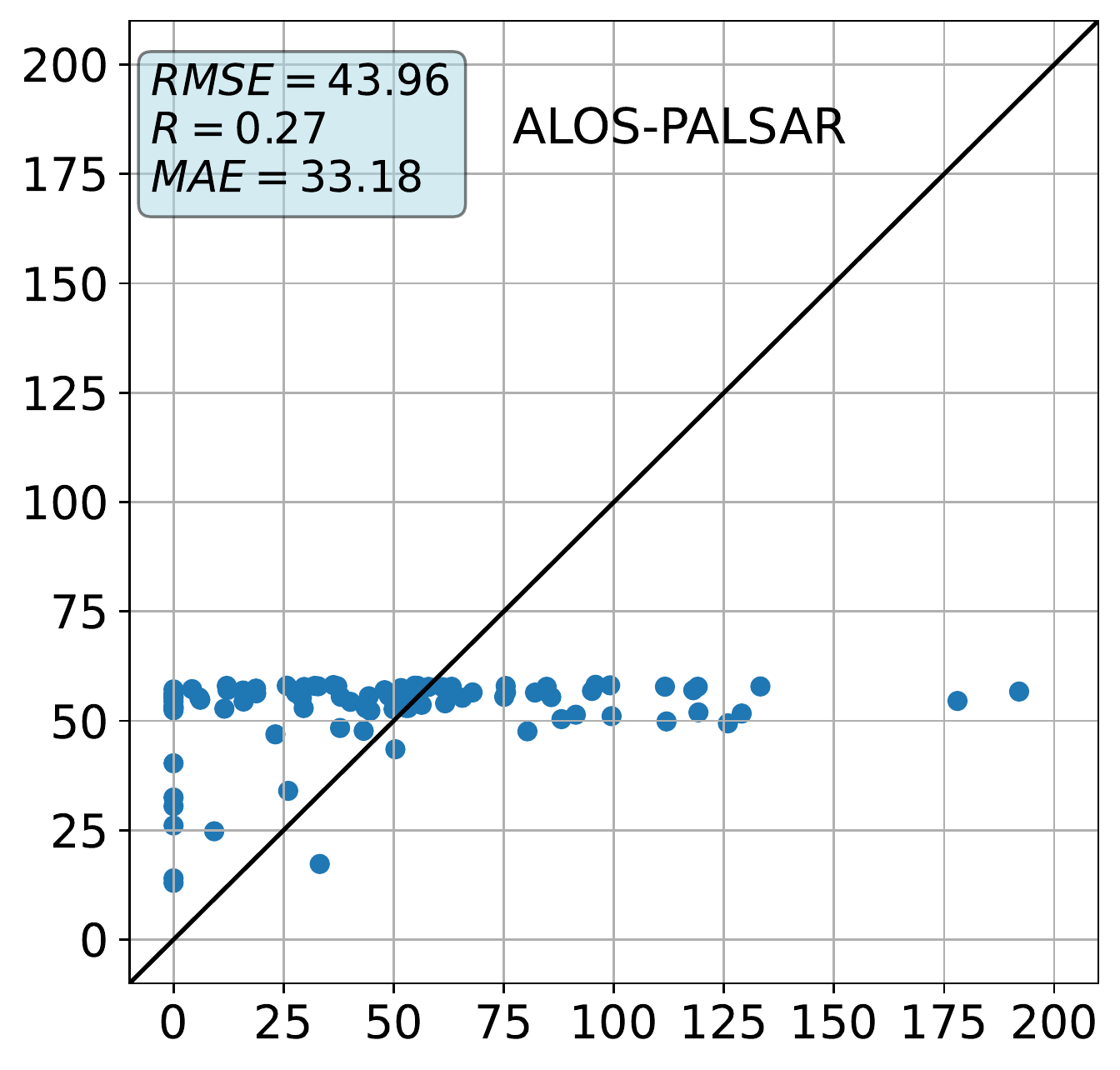}}
\subfloat[\label{fig:s1a_scatter}]{\includegraphics[width=0.44\columnwidth]{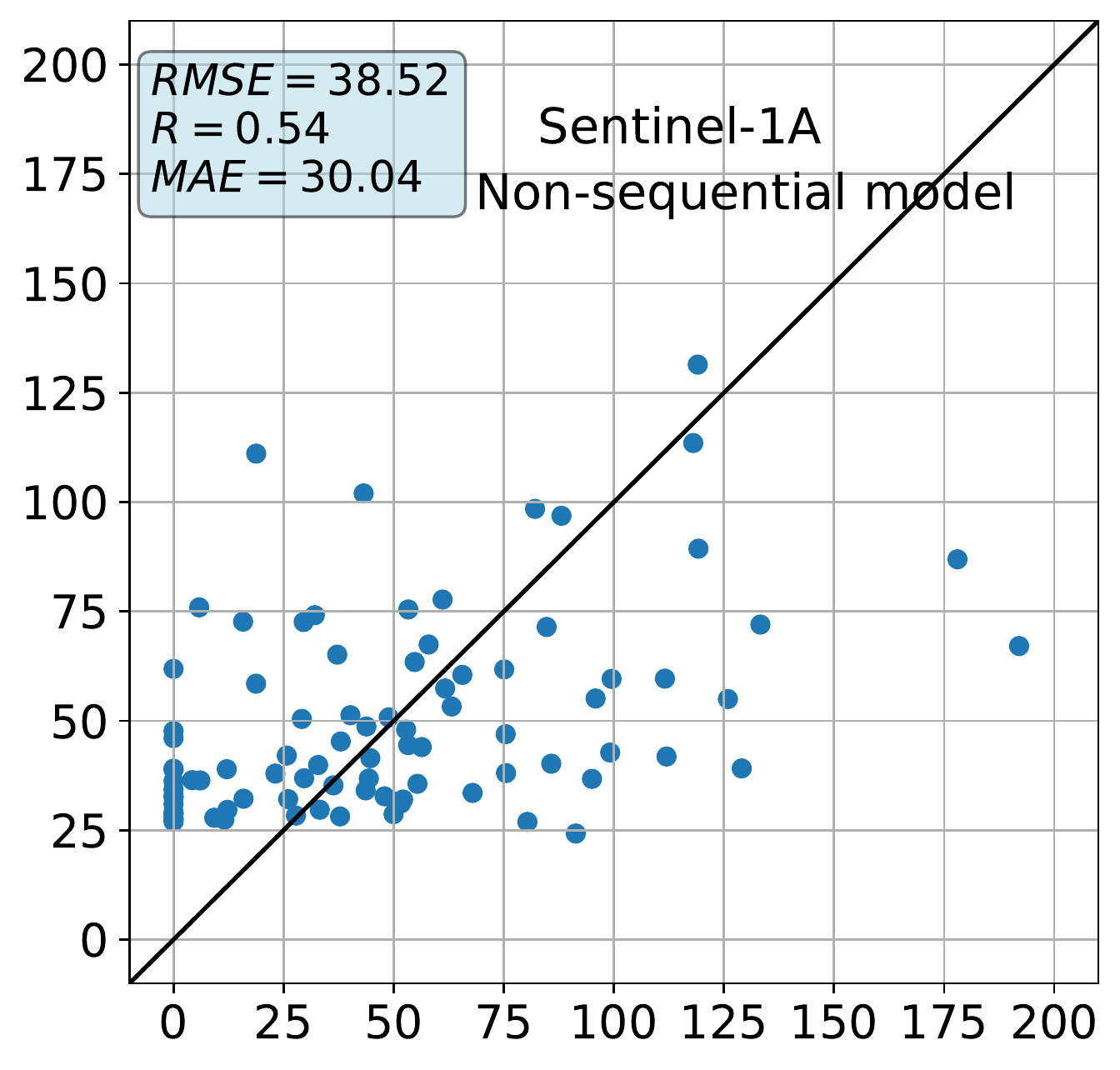}}
\vspace*{.5cm}
\\
\vspace*{.1cm}\hspace*{2cm}{Ground Reference biomass (Mg ha$^{-1}$)}\\
\leavevmode\smash{\makebox[0pt]{\hspace{3em}           
  \rotatebox[origin=l]{90}{\hspace{16em}
    Predicted AGB (Mg ha$^{-1}$)}
}} 
\vspace{-.3cm}
\caption{Scatter plots between ground reference AGB, $z$, and model-predicted AGB. Model-predicted AGB is retrieved from either the ALS \textbf{(a)}, InSAR \textbf{(b)}, RapidEye \textbf{(c)}, Landsat \textbf{(d)}, PALSAR \textbf{(e)} or our proposed \sen-based non-sequential regression model \textbf{(f)}. The black lines are reference lines indicating 100\% correlation between $z$ and predictions. Units are in Mg ha$^{-1}$}.
    \label{fig:scatter_trad}
\end{figure}

\begin{figure*}[htb] 
    \centering
\subfloat[\label{als_agb}ALS]{\includegraphics[width=0.5\textwidth]{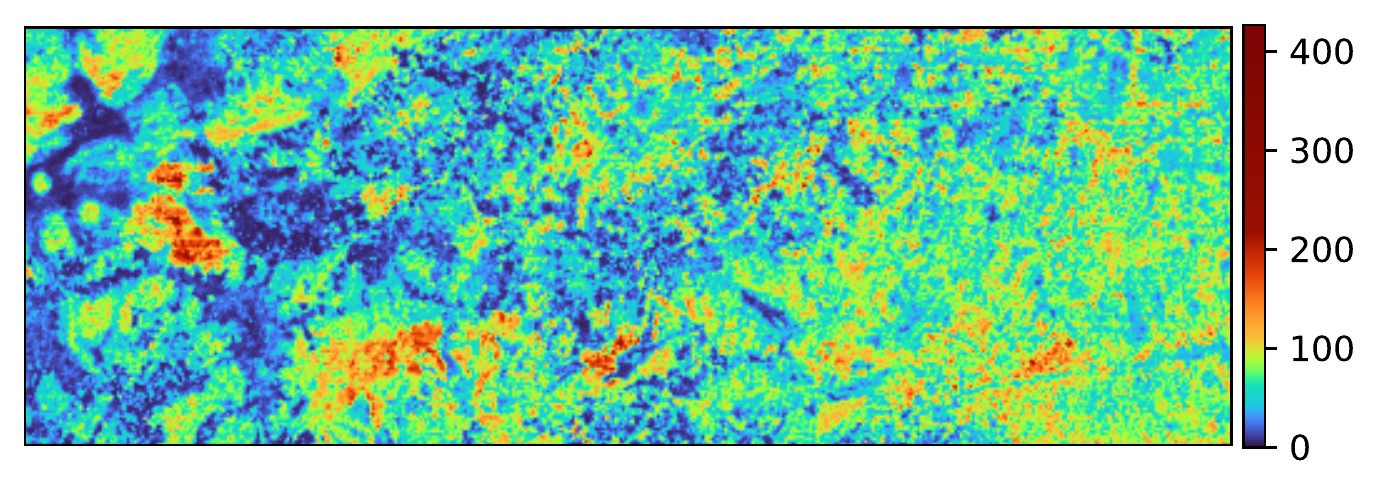}} 
\subfloat[\label{insar_agb}InSAR]{\includegraphics[width=0.5\textwidth]{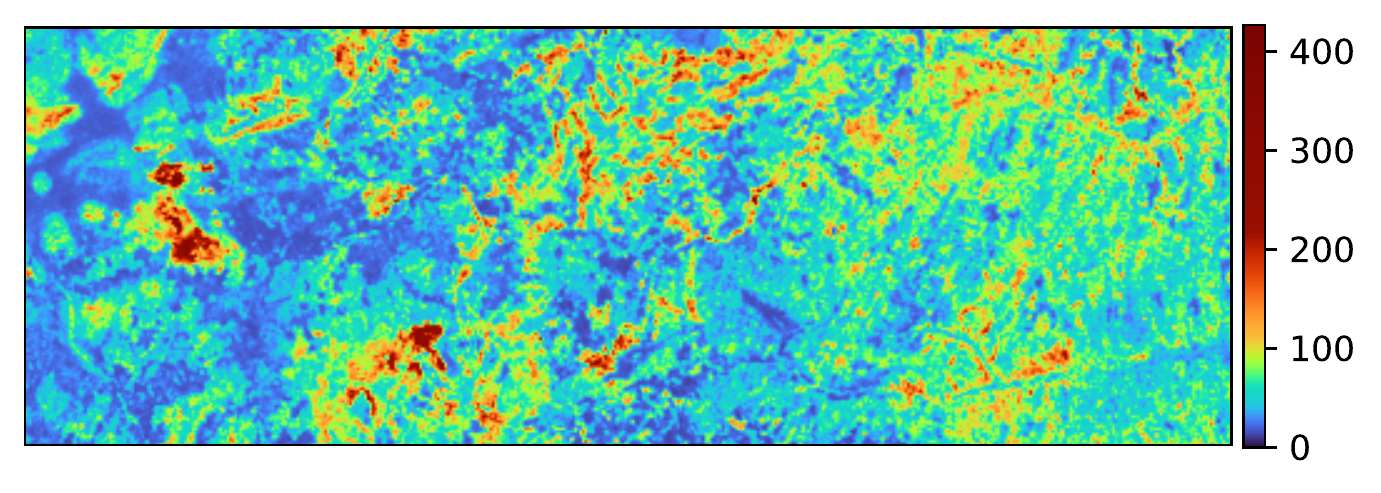}}\\
\vspace*{0.1cm}
\subfloat[\label{re_agb}RapidEye]{\includegraphics[width=0.5\textwidth]{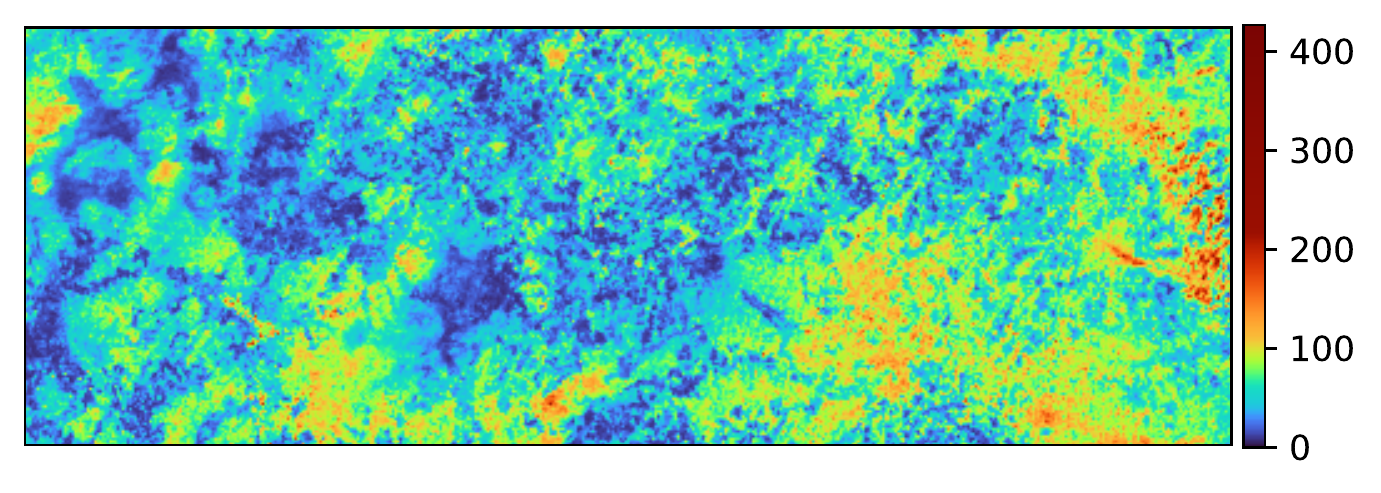}}
\subfloat[\label{landsat_agb}Landsat]{\includegraphics[width=0.5\textwidth]{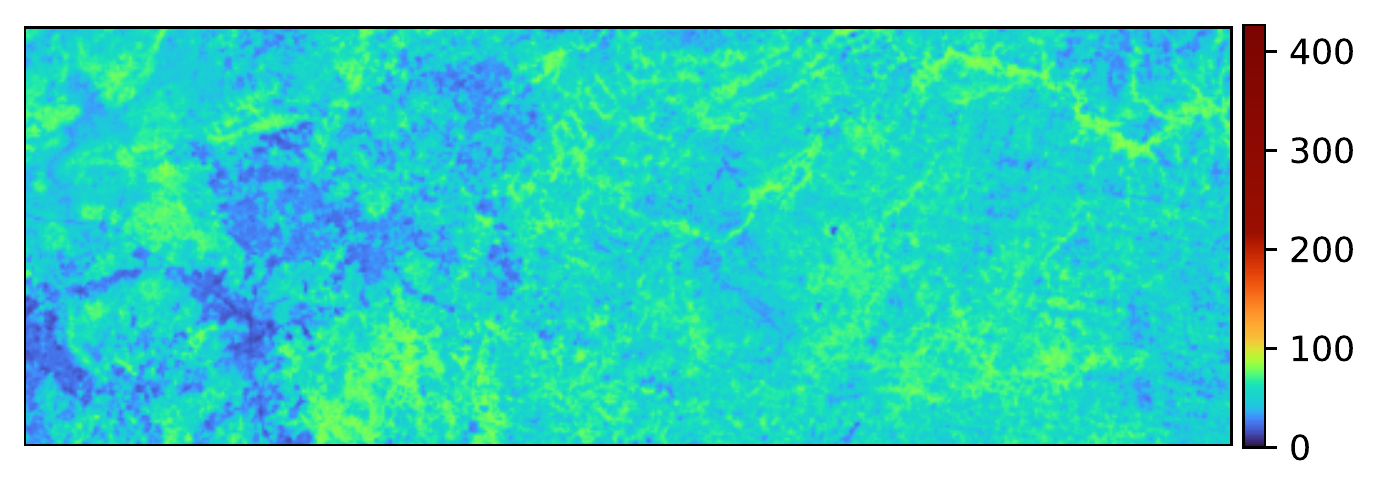}}\\
\subfloat[\label{alos_agb}PALSAR]{
       \includegraphics[width=0.5\textwidth]{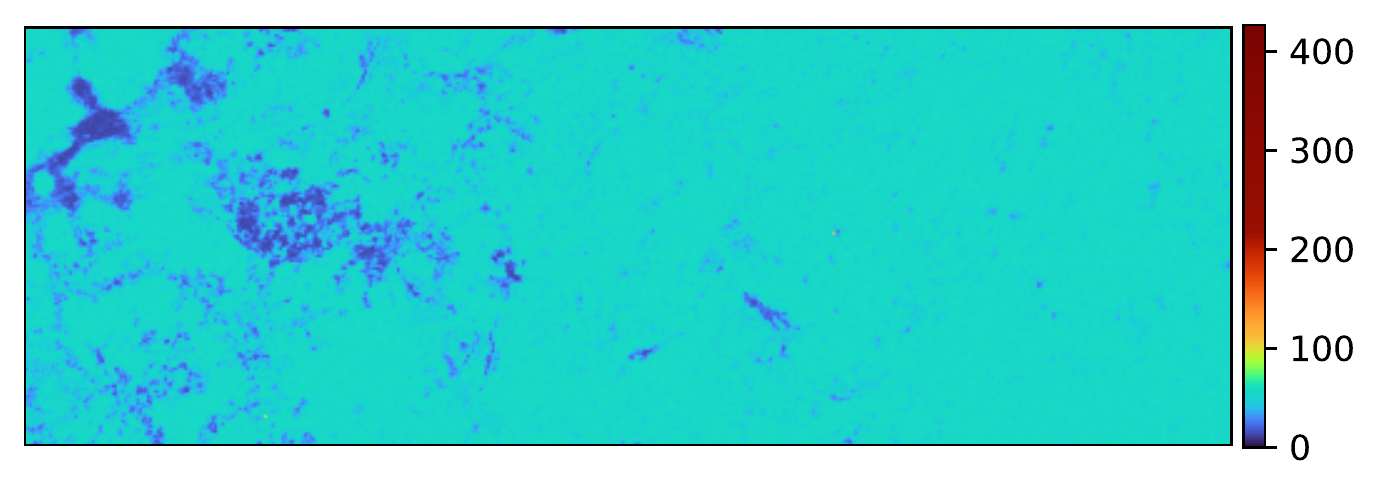}}
  \subfloat[\label{fig:s1a_non_seq_agb}Non-sequental \sen]{
        \includegraphics[width=0.5\textwidth]{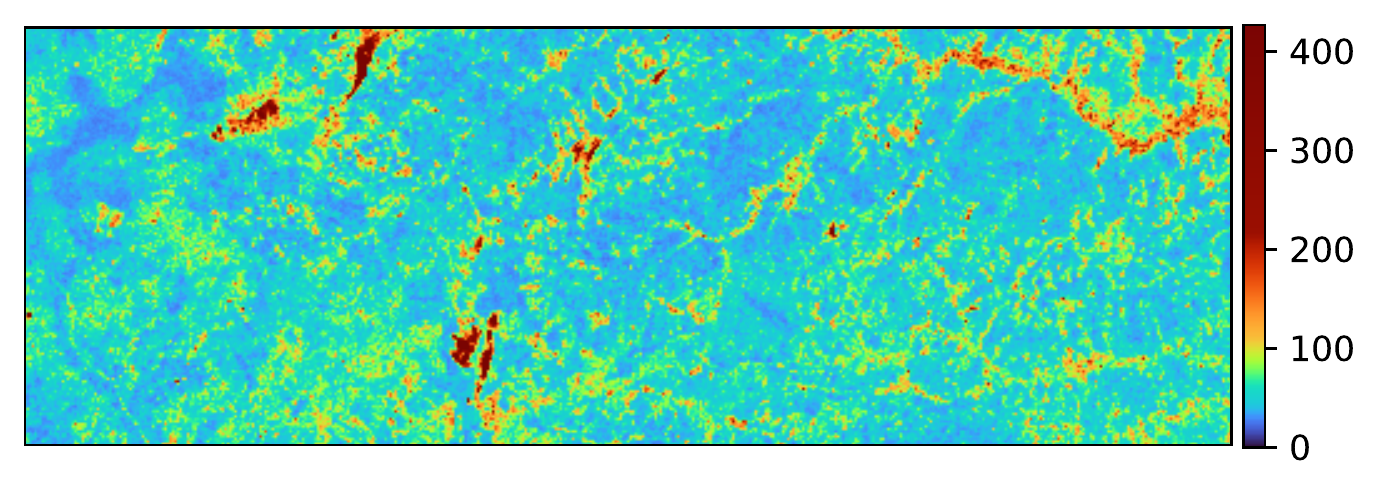}}\\
\caption{Aboveground biomass prediction maps (in $\mathrm{Mg\,ha}^{-1}$). Fig.\ \textbf{(a)}-\textbf{(e)} show results of the traditional non-sequential regression models presented in \cite{naessetMappingEstimatingForest2016}. The AGB biomass map in \textbf{(f)} were constructed from the proposed non-sequential \sen-based AGB model in \tabref{tab:res_trad}.}    
  \label{fig:AGBmaps_trad}
\end{figure*} 

\subsection{A traditional non-sequential regression model for AGB} \label{sec:sen1trad} 
We extend the work of \cite{naessetMappingEstimatingForest2016} by developing a traditional non-sequential regression model, $h$, for the 88 field plots of AGB ground reference data ($z$) according to \eq{eq:agb}. To do so, we laid the circular field plots of $z$ on top of the \sen\ pixel grid. VH and VV backscatter values corresponding to $z$ were found by computing the area-weighted mean of \sen\ pixels intersecting the field plots. Only one explanatory variable, i.e.\ VV, was selected in the step-wise forward selection procedure. 
The achieved model $h$, for AGB per hectare, is listed in \tabref{tab:res_trad}. Since the model was fitted on the whole ground reference dataset $z$, we follow \cite{naessetMappingEstimatingForest2016} and perform additional quantitative model assessment analysis through leave-one-out cross-validation (LOOCV) to compare the consistency of predicted AGB. We also compute the Pearson correlation coefficient (R), root mean squared error (RMSE) and mean absolute error (MAE) between model predicted AGB and $z$. These metrics are collected in \tabref{tab:res_trad} together with computed R and RMSE from the non-sequential regression model developed in \cite{naessetMappingEstimatingForest2016}.
Additionally, we qualitatively assessed our model against those developed in \cite{naessetMappingEstimatingForest2016} by plotting model-predicted AGB against $z$ in \figr{fig:scatter_trad} and by illustrating model-derived AGB wall-to-wall maps in \figr{fig:AGBmaps_trad}.
Minor differences between the scatter plots in \figr{fig:als_scatter}-e and data reported in \tabref{tab:res_trad}, compared to the corresponding figures and table in \cite{naessetMappingEstimatingForest2016} can be explained by differing pixel grids used in the area-weighting of remote sensing pixel values. In \cite{naessetMappingEstimatingForest2016}, N\ae sset~\emph{et~al.} developed their traditional non-sequential regression models for InSAR, RapidEye, Landsat and PALSAR by using the original pixel grid of the satellite data. When reporting metrics, they further used each sensor's original pixel grid to compute the area-weighted average of pixel values within the coverage of each field plot. After preprocessing the \sen\ scene, both the \sen\ dataset and the ALS-based AGB predictions are on the same grid with pixel size 707m$^2$, representing an area of 26.6 m $\times$ 26.6 m on the ground. In this work, we did not have access to the original pixel grids of the ALS, InSAR, RapidEye, Landsat and PALSAR data. Therefore, we chose to use the grid with pixel size 707m$^2$ for also these models whenever area-weighted metrics were computed. The resulting differences to \cite{naessetMappingEstimatingForest2016} must therefore be endured.

We observe from \tabref{tab:res_trad} that only two of the previously developed models in \cite{naessetMappingEstimatingForest2016}, i.e. the ALS-based ($f$) and the RapidEye-based models, experience lower RMSE and a higher Pearson correlation coefficient with respect to $z$ than our model $h$. Surprisingly, the respective InSAR and PALSAR-based models perform worse than the proposed model $h$ in terms of R and RMSE. The InSAR-based AGB model, used in \cite{naessetMappingEstimatingForest2016} and developed by \cite{Solberg2015Monitoring}, uses mean InSAR heights as the only explanatory variable. As canopy heights are highly correlated with AGB \cite{Kaasalainen2015, Galidaki}, this model was expected to correlate better with $z$ than our model $h$. However, Næsset \emph{et al.}\ \cite{naessetMappingEstimatingForest2016} highlight the temporal mismatch between the acquisition of the InSAR data (2012) and the acquisition of the field work (2014) as a probable explanation for the model's low performance. In one case, for example, they identified that a field plot recently had been harvested in 2014, while the InSAR data from 2012 identified biomass in the same area \cite{naessetMappingEstimatingForest2016}.
In theory, we expect a model based on the L-band ALOS PALSAR data to perform better than our C-band based \sen\ model, as C-band data is known to saturate at a lower biomass level than L-band data \cite{dobson1992dependence, le2002relationships, le1992relating}. However \tabref{tab:res_trad} show that this is not the case. As the PALSAR data used in \cite{naessetMappingEstimatingForest2016} consist of a mosaic of yearly scenes, the mosaic does not achieve optimal sensitivity to dynamic AGB levels as scenes from wet and dry seasons are mixed. The low dynamic range of the PALSAR-based and the Landsat-based models are also shown in \figr{fig:scatter_trad} and the wall-to-wall maps in \figr{fig:AGBmaps_trad}. Although most \sen\ predictions on the ground reference AGB dataset are bounded between 25-75 $\mathrm{Mg\,ha}^{-1}$, see \figr{fig:s1a_scatter}, the model as a whole is able to predict AGB up to around 200 $\mathrm{Mg\,ha}^{-1}$, see \figr{fig:s1a_non_seq_agb}. The upper limit of the $\zhatx$-based predictions \figr{fig:s1a_scatter} can probably be explained by the low saturation limit of C-band data. Nevertheless, our upper limit of C-band-based AGB predictions are still remarkable, compared to previous studies on biomass retrieval from C-band data, e.g.\ Imhoff \cite{Imhoff1995cband} who showed that C-band data saturates around 20 $\mathrm{Mg\,ha}^{-1}$ in the tropical forests of Hawaii. 
We wish to highlight the fact the proposed model $h$ is not able to predict biomass close to 0 $\mathrm{Mg\,ha}^{-1}$, see \figr{fig:s1a_scatter} and \figr{fig:s1a_non_seq_agb}. This is probably due to the square root transform in \eq{eq:sqrtagb} and the correction of bias in \eq{eq:agb}, the latter applied to achieve correct AGB predictions on arithmetic form, i.e.\ back-transformation from the $\sqrt{\mathrm{AGB}}$ domain. The InSAR-based model, on the other hand is able to predict AGB levels close to 0 $\mathrm{Mg\,ha}^{-1}$, see \tabref{tab:res_trad} and \figr{fig:insar_scatter} and also achieves lower MAE than the proposed model $h$.
  
\subsection{Sequential regression models for AGB}\label{sec:seqmod}
This section presents the two alternatives for $g$, the second model in the sequential modelling approach, i.e.\ the traditional baseline sequential model and the cGAN-based model. Since the regression model $f$ achieves the highest correlation to $z$, see \cite{naessetMappingEstimatingForest2016}, we train our two versions of $g$ to use the ALS-based AGB predictions (on pixel-wise form: $\zhaty$, or patch-wise form: $\zhatyb$) as a surrogate for $z$. Each AGB prediction, i.e.\ $\zhaty$, represents a square pixel of size 26.6 m $\times$ 26.6 m on the ground. Qualitative and quantitative results from both models are presented and discussed in \Sec{sec:seq_eval}.

\subsubsection{Baseline sequential regression model}\label{sec:baseseq}
The proposed baseline sequential regression modelling strategy utilises the traditional regression model in \eq{eq:agb} for both stages in the sequence. In \Sec{sec:sen1trad}, the small size of the $z$ dataset constrained us to use all available data during both model fitting and evaluation. Reusing all available data for both model fitting and evaluation is not optimal, which also \tabref{tab:res_trad} shows, i.e.\ the RMSE computed for model $h$ is lower than the corresponding LOOCV RMSE. In contrast to the situation in \Sec{sec:sen1trad}, the sequential model setting provides access to 516,906 AGB predictions to be used as surrogate response variables. Thus, the dataset size enables us to fit and evaluate model $g$ on different parts of the dataset.

We adopt a dataset split of 20\% validation data and 80\% test data. We use the validation data to select the models's explanatory variables through stepwise forward selection. Contrary to the non-sequential model $h$, which only selects VV as a regressor, all six explanatory variables are included in the baseline model $g$ by this method. The final baseline sequential regression model is shown in \tabref{tab:res_seqmod}. The test dataset was divided  into $k=5$ subsets for k-fold cross-validation (CV). The chosen test metric is CV RMSE (CV-RMSE), which is reported in addition to the Pearson correlation coefficient and the RMSE in \tabref{tab:res_seqmod}. The latter two metrics are  computed on the entire dataset. All reported metrics are computed between the surrogate, i.e.\ $\zhaty$, and AGB predictions achieved from the baseline sequential  model, i.e.\ $\zhatyx$. 

\subsubsection{cGAN-based sequential regression models}\label{sec:cganseq}
\begin{table}[]
\caption{The three optimal cGAN-based models applied for the second part of the sequential modelling approach. They were identified from experiments reported in the Appendix, see \Sec{exp:data} and \Sec{exp:exp2}. Vanilla GAN, LSGAN and WGAN-GP refer to specific objective functions. \textit{BN} denote \textit{batch normalisation} and \textit{BS} denote \textit{batch size}.}
    \label{tab:optcgan}
    \centering
    \begin{tabular}{|l|l|}
         \hline
    Model reference & Trained with:\\ 
   \hline
    Vanilla GAN &   ResNet-6, BN, BS = 3 and \textit{PixelGAN} discriminator\\
    LSGAN &  ResNet-6, BN, BS = 3 and \textit{PixelGAN} discriminator\\
    WGAN-GP &  ResNet-6, BN, BS = 3 and \textit{PixelGAN} discriminator\\
    \hline 
    \end{tabular}
\end{table} 

Lastly, we approach the sequential modelling strategy from a DL perspective by applying a cGAN for the second regression model, $g$. The cGAN-based model utilises convolutional filters in both the $\G$ and the $\D$ network. Therefore, the image-to-image translation requires the data we condition on, and the output data, to be represented by image patches instead of individual image pixels. Image patches were created from the input data, i.e.\ the processed \sen\ image,and the output dataset of 516,906 ALS-based AGB predictions, i.e.\ $\zhatyb$, similarly and simultaneously. For simplicity, we only describe the process for the \sen\ data.
Firstly, non-overlapping image patches of size $64\times64$ pixels were extracted in a grid manner from the \sen\ scene in \figr{fig:lidarsen}. Each patch corresponds to an area of approximately 289.6 ha on the ground. These non-overlapping image patches were randomly divided into five disjoint sets for 5-fold CV. For each of the five folds, one of the disjoint sets was considered the test set, while the remaining four folds were combined into a training set. To increase the number of image patches further, we extracted additional training patches in each training set by allowing a 50 \% overlap between adjacent patches.
Finally, we applied data augmentation with flipping and rotation to the training image patches. Since we do not allow overlap between test and training image patches, it implies that the final five training sets, after data augmentation, range between 2264 and 2424 patches. Each test set consists of 22 image patches  since no data augmentation was applied to the test sets.

\begin{table*}[htp]
\caption{Pearson correlation coefficients, R, RMSE and (CV-RMSE) computed between ALS predicted AGB, $\zhaty$, and model predicted AGB, $\zhatyx$, achieved from our sequential modelling approach. All metrics are in units of $\mathrm{Mg\,ha}^{-1}$.}
    \label{tab:res_seqmod}
    \centering
    \begin{tabular}{|l|l|c|l|l|c|}
         \hline
  Auxiliary data source & Modelling approach & Model & R & RMSE & CV-RMSE\\ 
   \hline
    \sen & Baseline sequential$^a$ &  $\widehat{AGB}  = \big(-1.61 -150.51VH -29.92VV +53.58\sqrt{VH}$ & 0.41  & 40.8 & 40.6 \\
        &                               &  $+25.64\sqrt{VV} + 271.36VH^2 +  9.64VV^2 \big)^2 +6.94$ & & & \\
    \sen & Sequential$^b$ &Vanilla GAN &  0.40  & 42.6 & 43.6\\
    \sen & Sequential$^b$ & LSGAN &  0.39  & 43.0 & 43.7\\
    \sen & Sequential$^b$ & WGAN-GP &  0.35  & 44.6 & 44.1\\
    \hline 
    \multicolumn{6}{l}{}\\
    \multicolumn{6}{l}{${}^\text{a}$ Baseline sequential model, see \Sec{sec:baseseq}} \\
    \multicolumn{6}{l}{${}^\text{b}$ cGAN-based sequential models, see \Sec{sec:cganseq}}\\
    \end{tabular}
\end{table*}

By condition on \sen\ image patches, we trained different cGAN-based models to generate realistic looking synthetic ALS-based AGB prediction image patches, $\zhatyxb$, of size $64\times 64$ pixels. Optimal translation would imply $\zhatyxb = \zhatyb$ or at least $\zhatyxb \approx \zhatyb$.
All models were trained for 200 epochs with a learning rate of $2\times10^{-4}$. We refer to \Sec{exp:data} and \Sec{exp:exp2} in the Appendix for an extensive evaluation of the impact that the choice of hyperparameters, objective function, and/or discriminator network have on the performance of the different cGAN models. For the remaining of this paper, we only report results for the three optimal cGAN-based models listed in \tabref{tab:optcgan}, which were identified from the extensive evaluation. Despite the selected objective function, these three models were trained with identical generator architecture, discriminator architecture and hyperparameters. We therefore refer to them by their objective function, i.e.\ as the Vanilla GAN, LSGAN or WGAN-GP model.

As the input and output to each of the optimal cGAN-based sequential models are of size $64\times 64$ pixels, we created synthetic ALS-based AGB prediction maps from the \sen\ scene as follows: the whole AOI  was first partitioned into $64\times 64$ image patches with 50\% overlap. For each of the optimal models, these patches were then fed into the trained $\G$ network to generate synthetic image patches with 50\% overlap. The generated synthetic image patches were then merged to construct a $\zhatyxb$ prediction map. Due to the overlap between the generated synthetic image patches, most pixels in this intermediate prediction map constitute of a weighted average of pixels from neighbouring image patches. Therefore, as a last step to the final $\zhatyxb$ prediction map, we apply mosaicking through linear image blending, using the $p$-norm with a heuristic value of $p=5$. Different norms were also considered, however we conclude that the specific choice of the norm has little impact on the blended result.

After training, we evaluated the performance of the Vanilla GAN, LSGAN and WGAN-GP model against each other and the baseline sequential regression model defined in \Sec{sec:baseseq}. We qualitatively and quantitatively compared $\zhatyx$ generated from the cGAN-based models against the 88 ground reference AGB plots, $z$, and the surrogate wall-to-wall map of AGB predictions, i.e.\ $\zhaty$. 

\begin{figure*}[htp]
    \centering
\subfloat[\label{als_agb2}ALS]{\includegraphics[width=0.5\textwidth]{images/agb_map/cmp2_agb_als.pdf}} \\
\subfloat[\label{s1aseq_agb2}Baseline sequential \sen]{\includegraphics[width=0.5\textwidth]{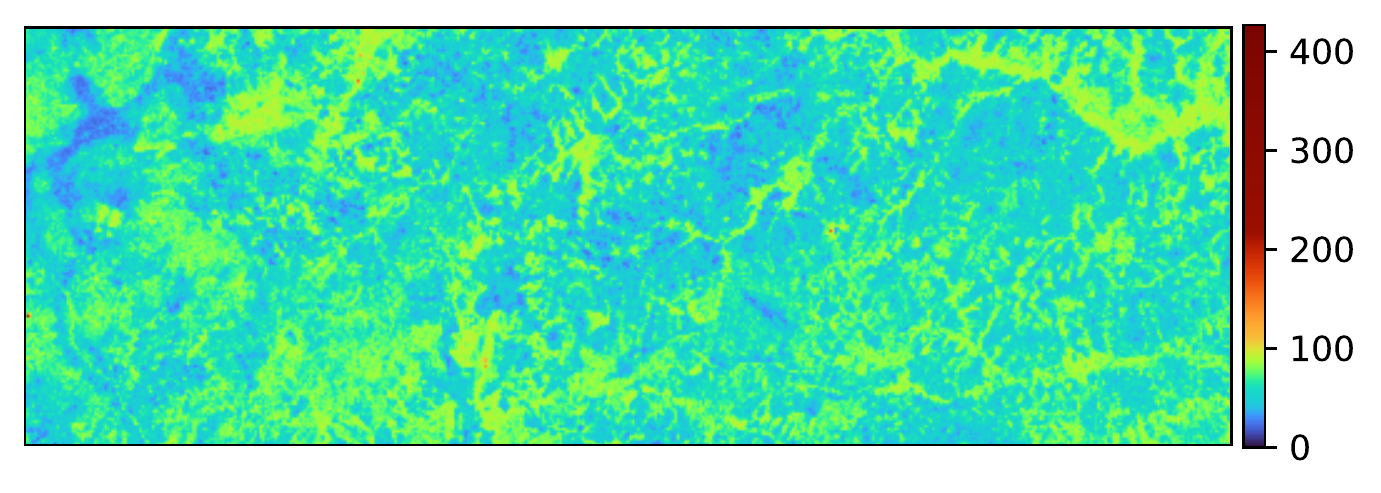}}
  \subfloat[\label{Vanilla_agb2}Vanilla GAN]{
        \includegraphics[width=0.5\textwidth]{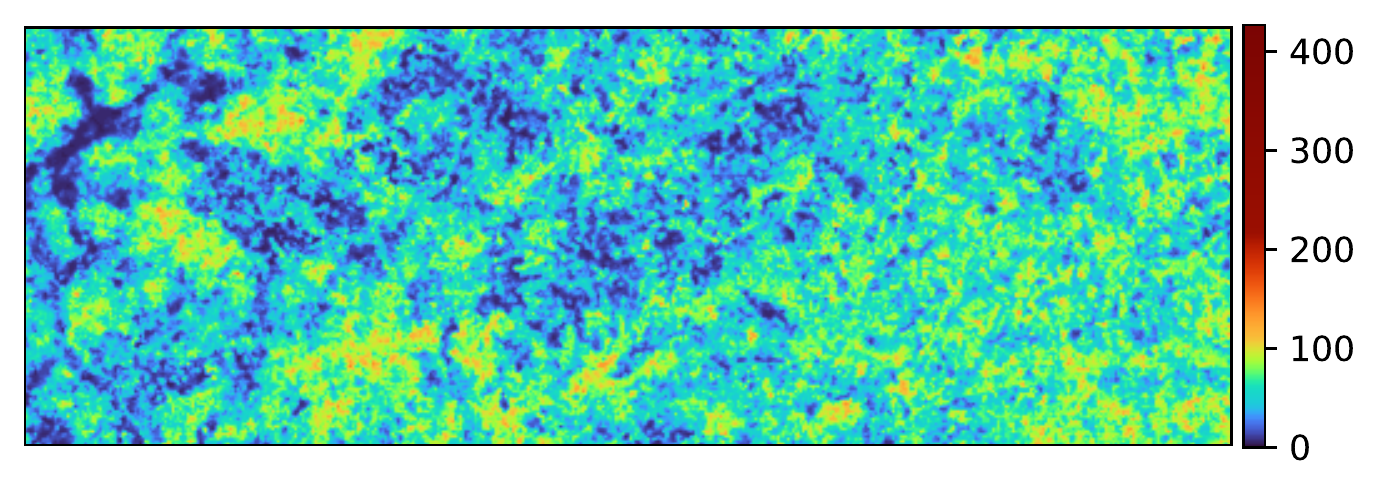}}\\
\subfloat[\label{lsgan_agb2}LSGAN]{
       \includegraphics[width=0.5\textwidth]{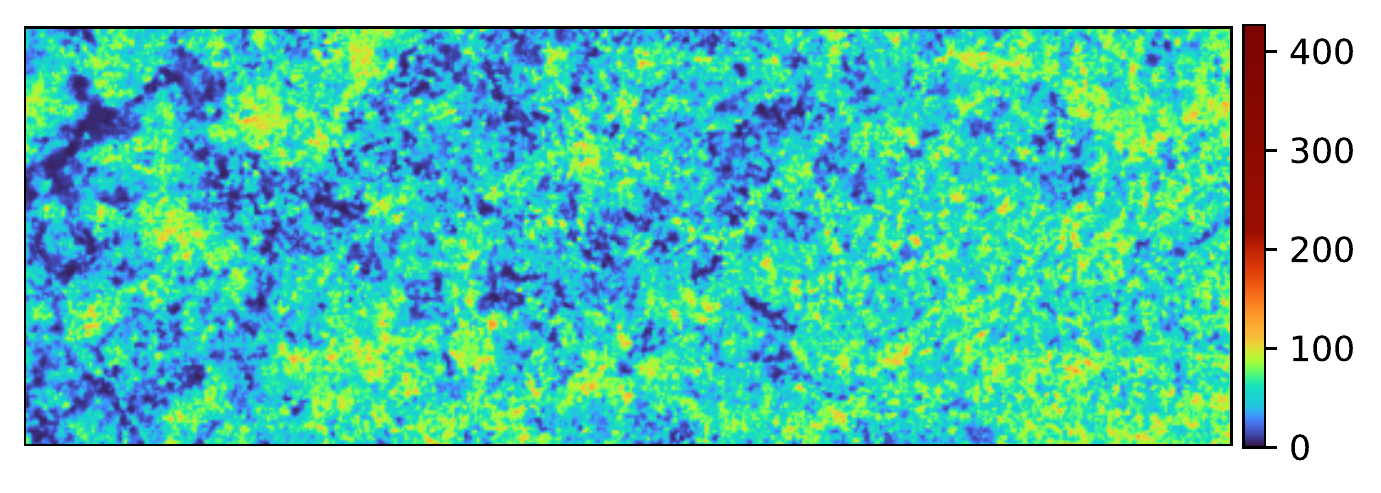}}
  \subfloat[\label{wgan_agb2}WGAN-GP]{
        \includegraphics[width=0.5\textwidth]{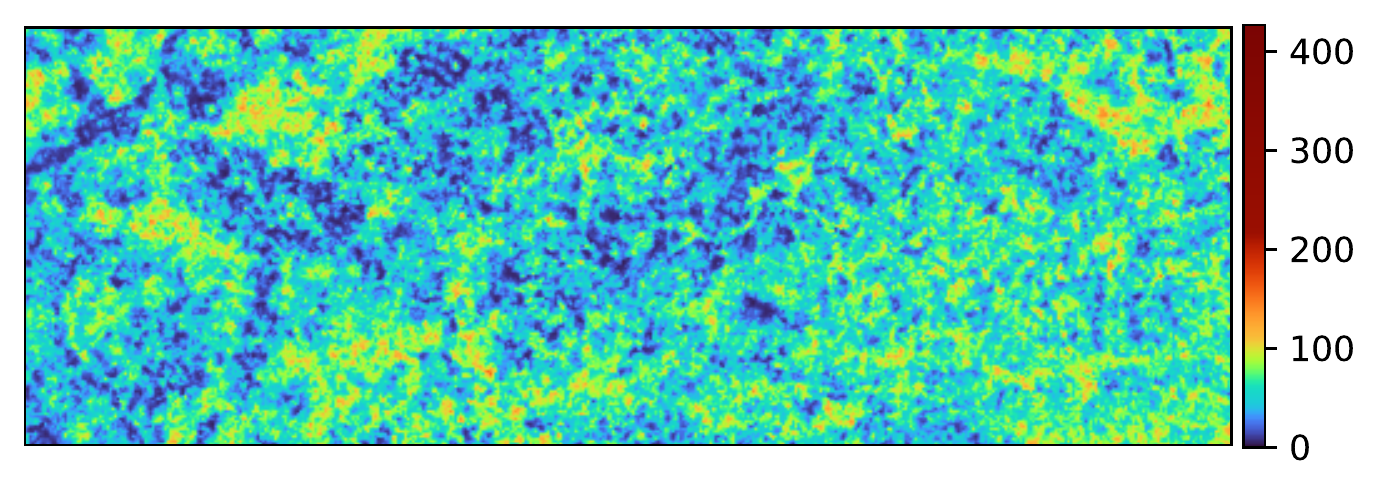}}
       
\caption{Generated synthetic ALS-based AGB prediction maps (in $\mathrm{Mg\,ha}^{-1}$) together with the  surrogate for ground reference plots, i.e. the ALS-based AGB map shown in \textbf{(a)} (this AGB map is the same as (a) in \figr{fig:AGBmaps_trad}). Fig. \textbf{(b)} represents a synthetic ALS-based AGB prediction map generated through the baseline sequential  \sen-model, (see \eq{eq:agb}). Fig.\ (b)-(e) are generated synthetic ALS-based AGB prediction maps generated through our proposed sequential regression models using \textbf{(c)} Vanilla GAN, \textbf{(d)} LSGAN and \textbf{(e)} WGAN-GP.}
  \label{fig:AGBmaps_seq}
\end{figure*}

\begin{figure*}[htb]
\hspace*{.3cm}
\subfloat[\label{als_agb_den}]{\includegraphics[width=0.25\textwidth]{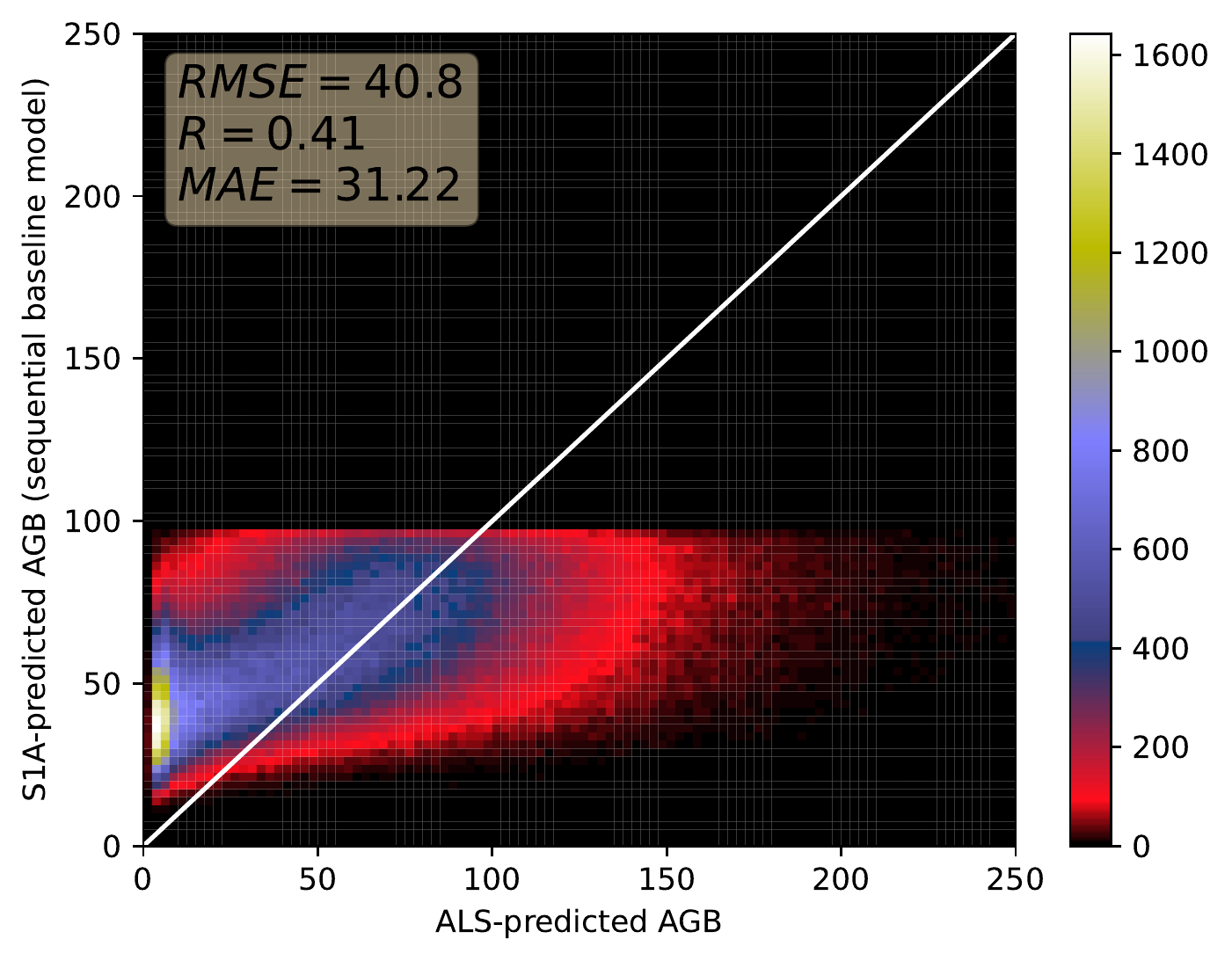}}
\hspace*{-.3cm}
\subfloat[\label{vanilla_agb_den}]{\includegraphics[width=0.25\textwidth]{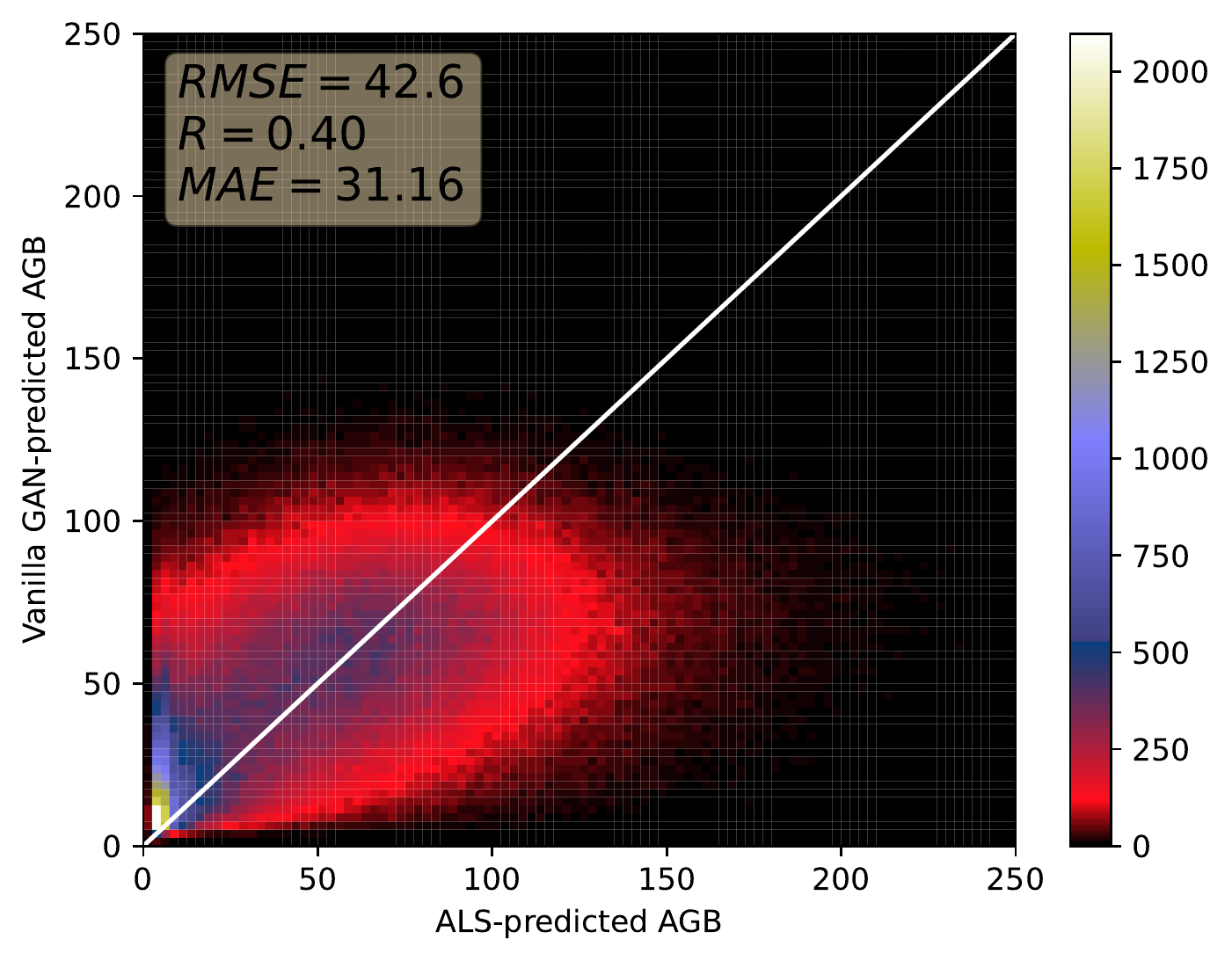}}
\hspace*{-.3cm}
 \subfloat[\label{lsgan_agb_den}]{\includegraphics[width=0.25\textwidth]{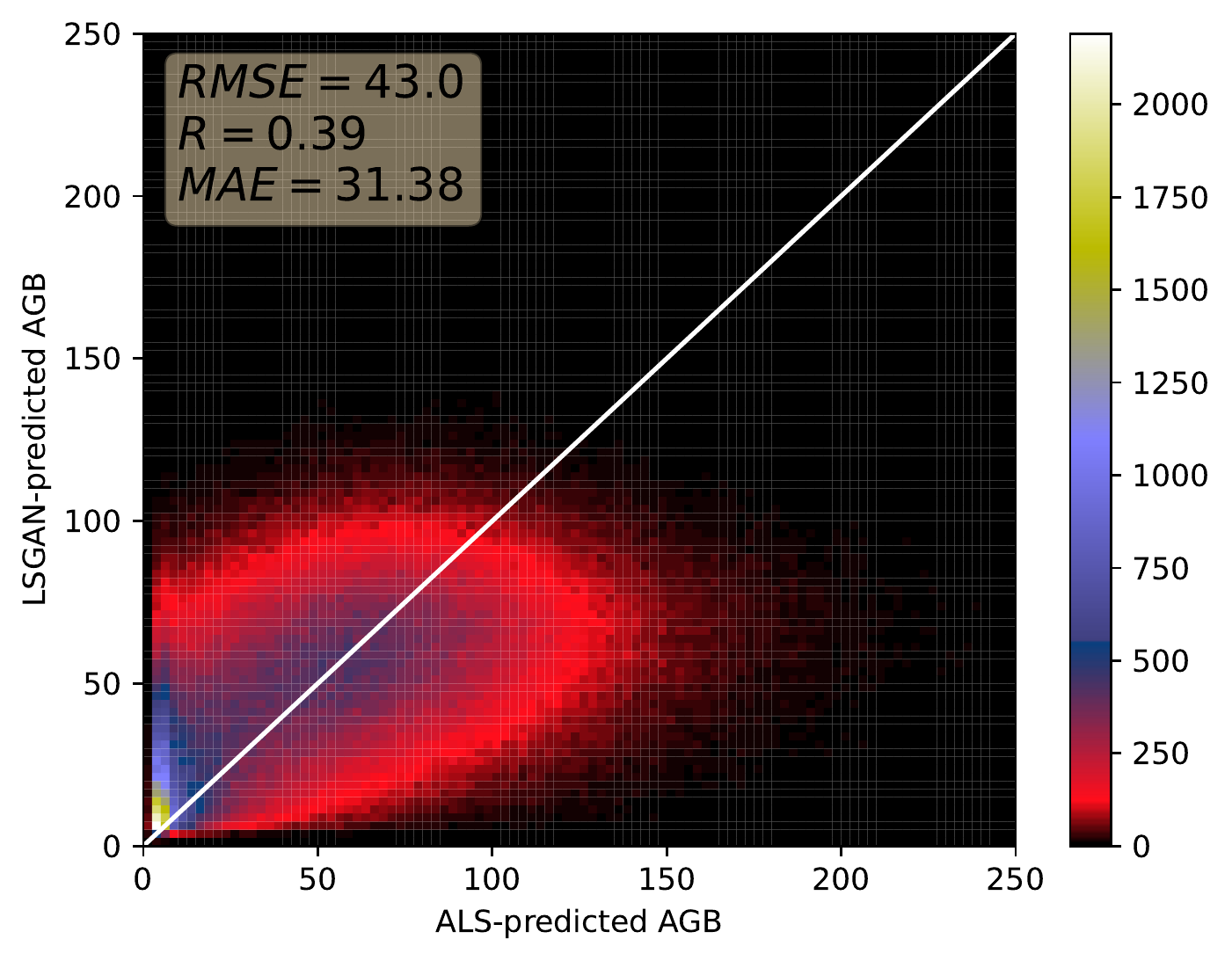}}
\hspace*{-.3cm}
\subfloat[\label{wgangp_agb_den}]{\includegraphics[width=0.25\textwidth]{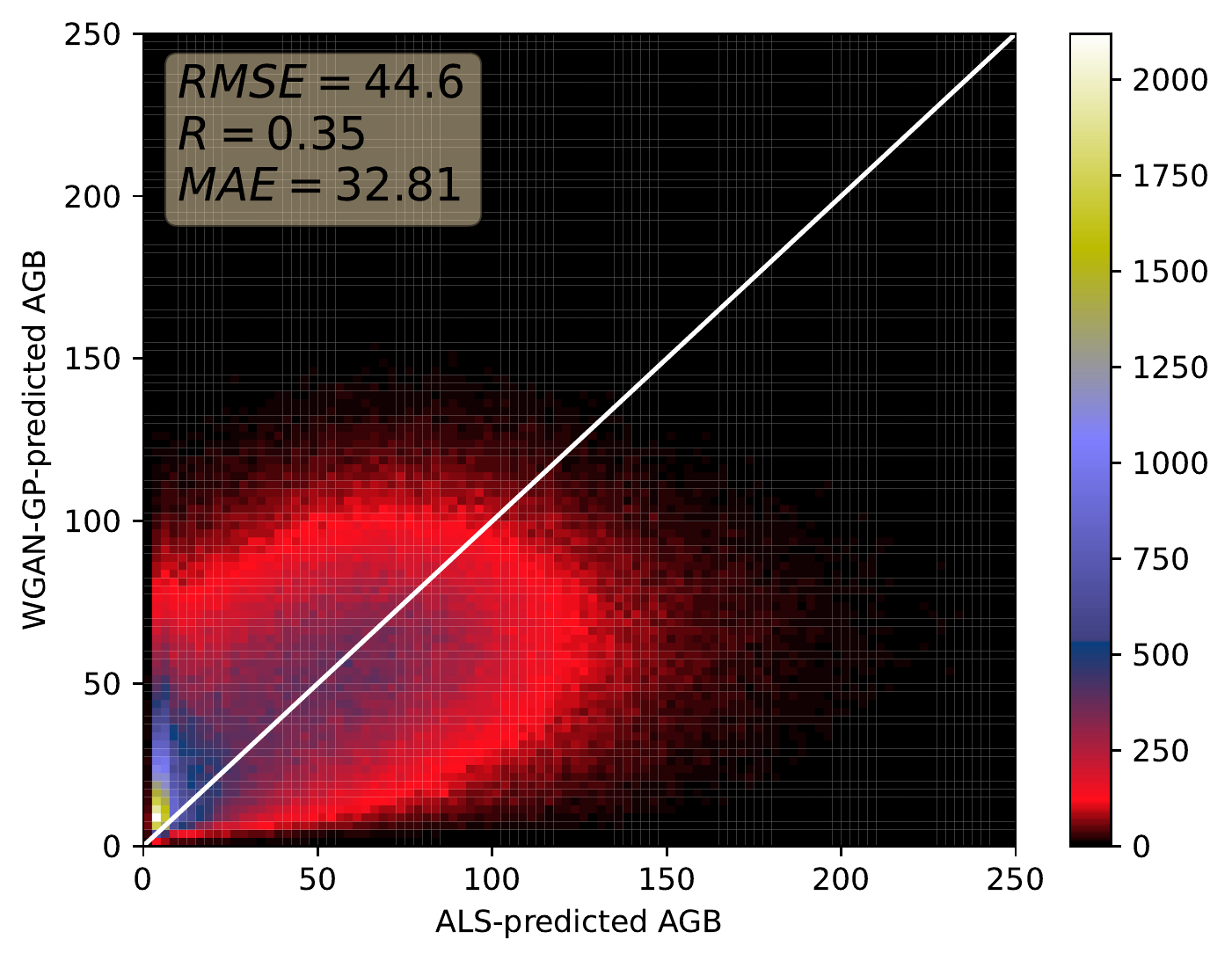}}
\vspace*{.2cm}

\caption{Density plots between constructed AGB maps and ALS-based AGB biomass predictions, $\zhaty$, for baseline sequential  model \textbf{(a)},  Vanilla GAN \textbf{(b)}, LSGAN \textbf{(c)} and WGAN-GP \textbf{(d)} models. Reported metrics are the RMSE, Pearson correlation coefficient (R) and the MAE between $\zhaty$ and the sequential model-based AGB predictions. The white lines are reference lines indicating 100\% correlation between $\zhaty$ and predictions.}
    \label{fig:density_seq}
\end{figure*}

\begin{figure}[htb] 
    \vspace*{-0.3cm}
\subfloat[]{\includegraphics[width=\columnwidth]{images/agb_map/cmp2_agb_als.pdf}}
\vspace*{-0.4cm}
    \subfloat[\label{s1_diff}]{\includegraphics[width=\columnwidth]{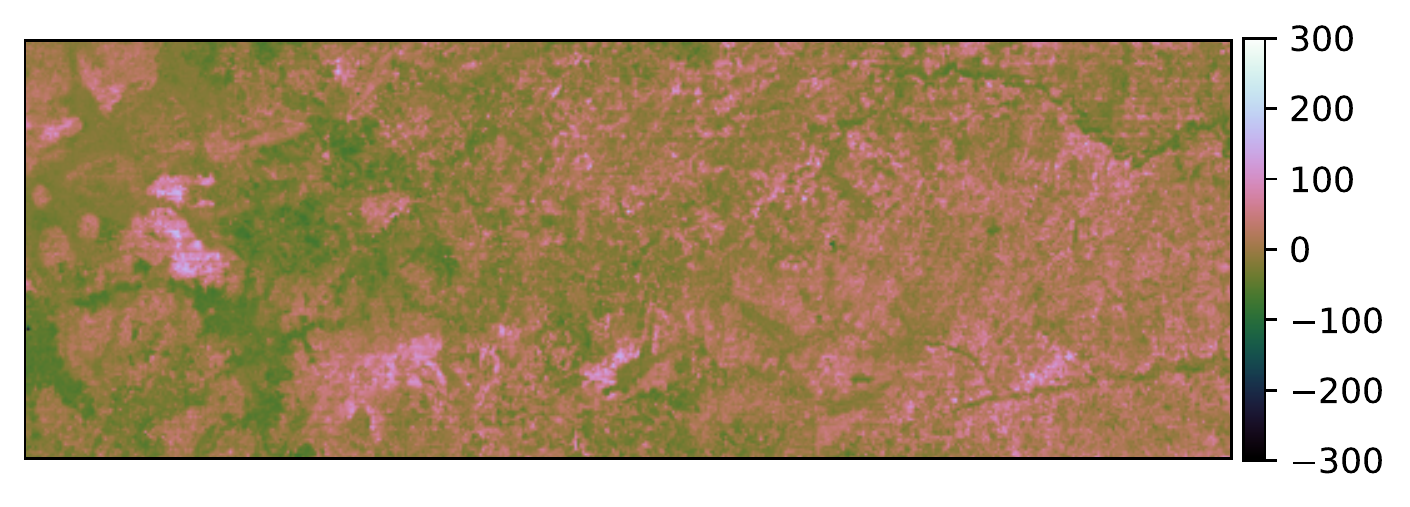}}
\vspace*{-0.4cm}
\subfloat[\label{van_diff}]{\includegraphics[width=\columnwidth]{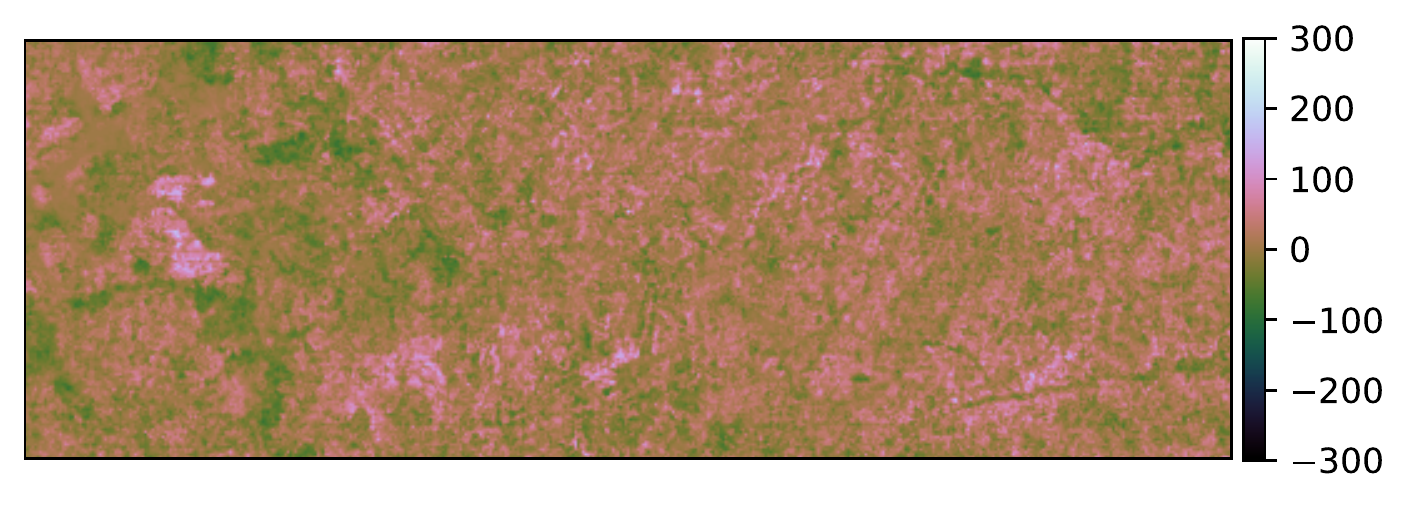}}
\vspace*{-0.4cm}
\subfloat[\label{ls_diff}]{
       \includegraphics[width=\columnwidth]{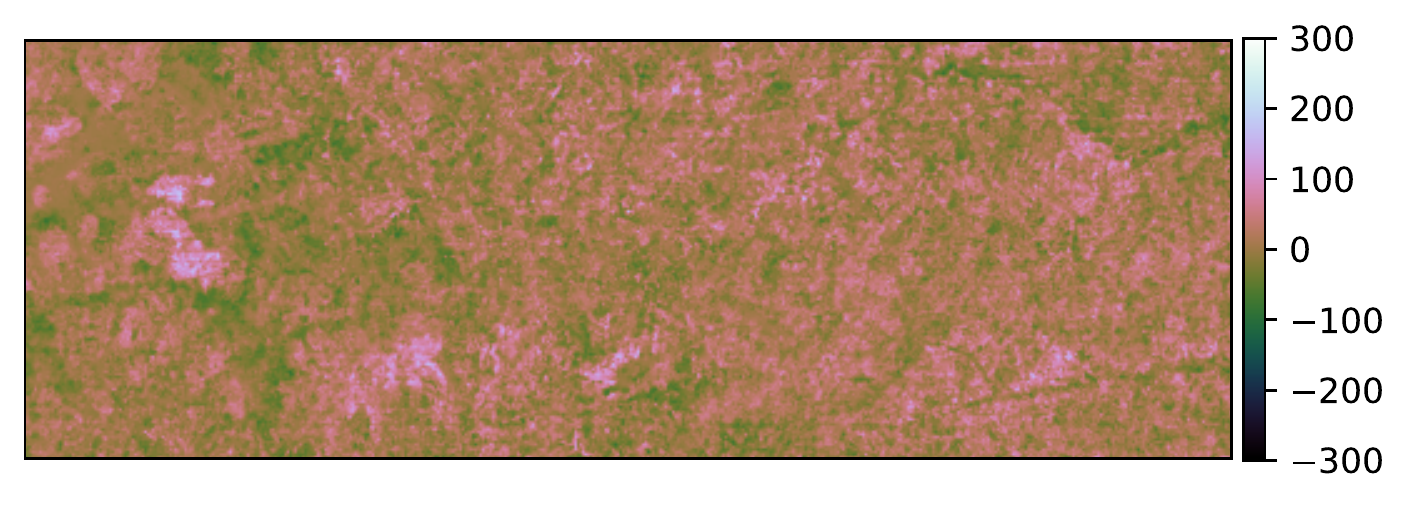}}
\vspace*{-0.4cm}
\subfloat[\label{wgan_diff}]{
       \includegraphics[width=\columnwidth]{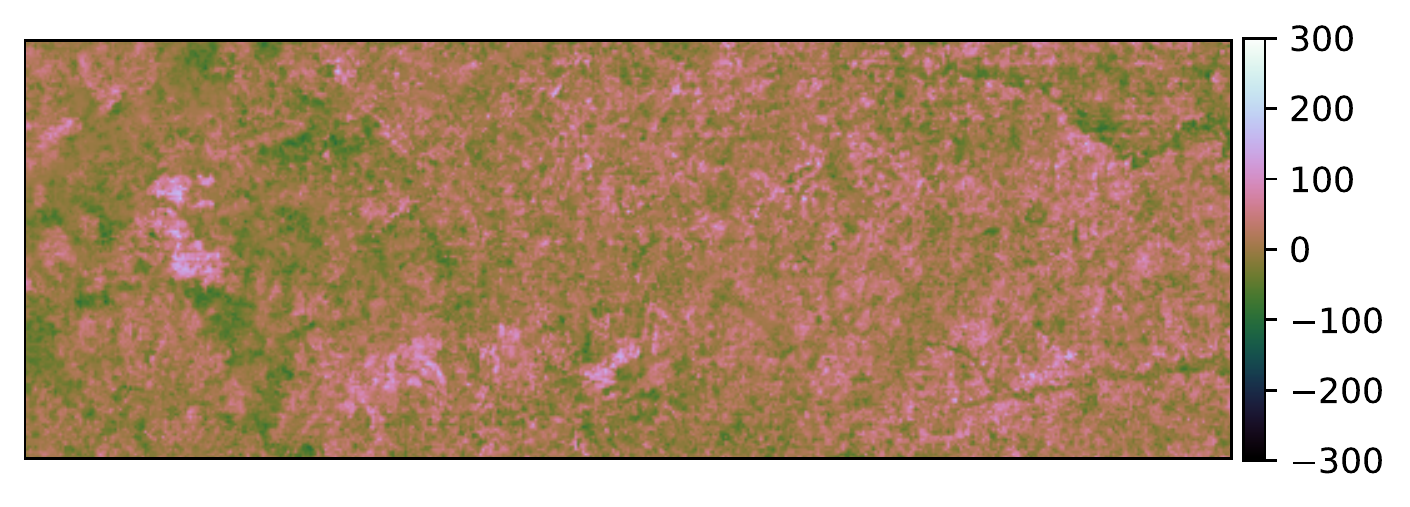}}
\caption{AGB difference maps (in $\mathrm{Mg\,ha}^{-1}$). Pixel-wise difference between the ALS-based AGB prediction map, shown in \textbf{(a)}, and constructed AGB prediction maps achieved form the four sequential models: baseline sequential  model \textbf{(b)}, Vanilla GAN \textbf{(c)}, LSGAN \textbf{(d)} and WGAN-GP \textbf{(e)}.}
  \label{fig:diffmaps_seq}
\end{figure}

\begin{figure*}[htb]
\hspace*{1.cm}
\subfloat[\label{s1aseq_scatter}]{\includegraphics[width=0.22\textwidth]{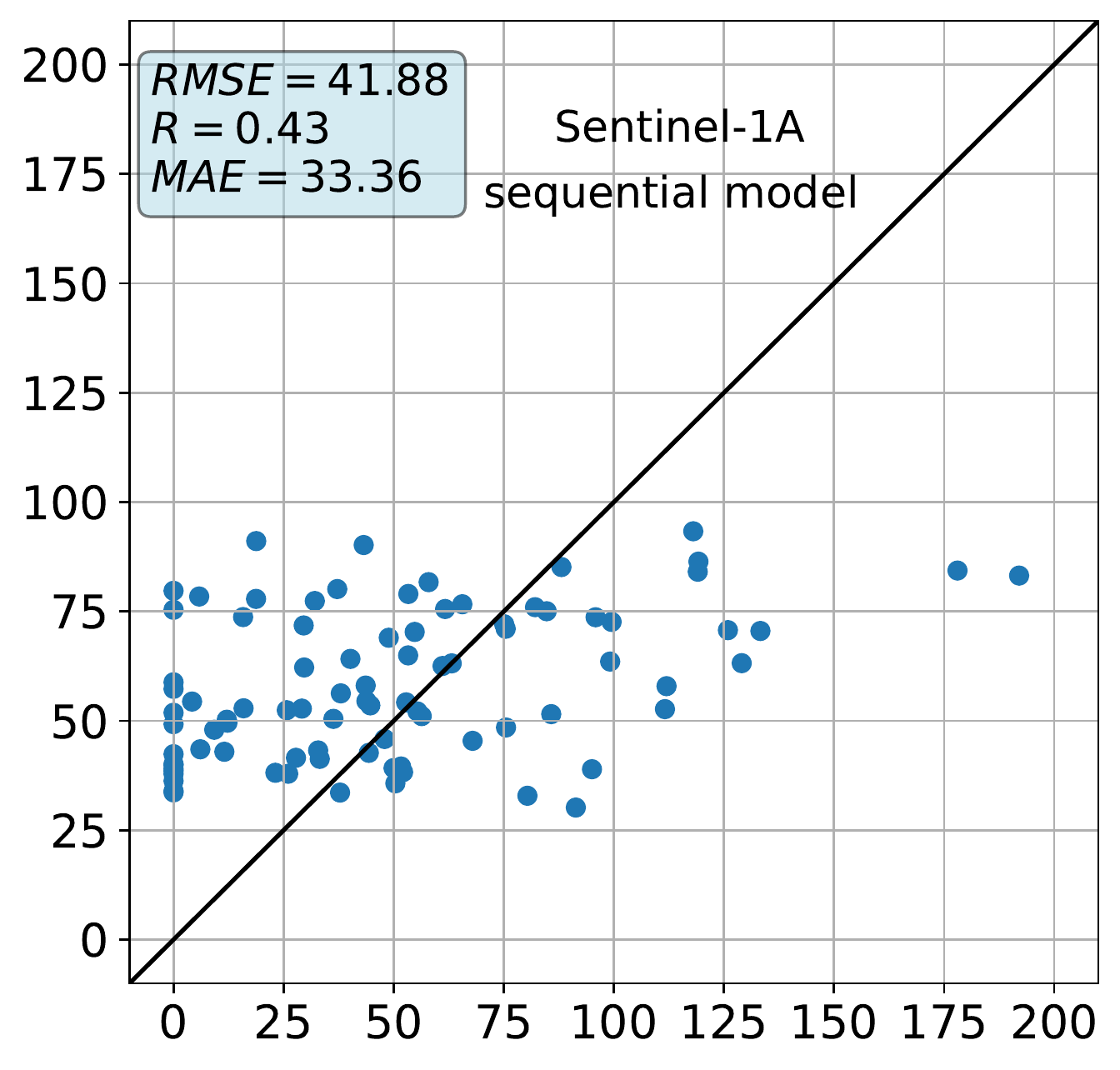}}
\hspace*{-.6cm}
\subfloat[\label{vanilla_scatter}]{\includegraphics[width=0.22\textwidth]{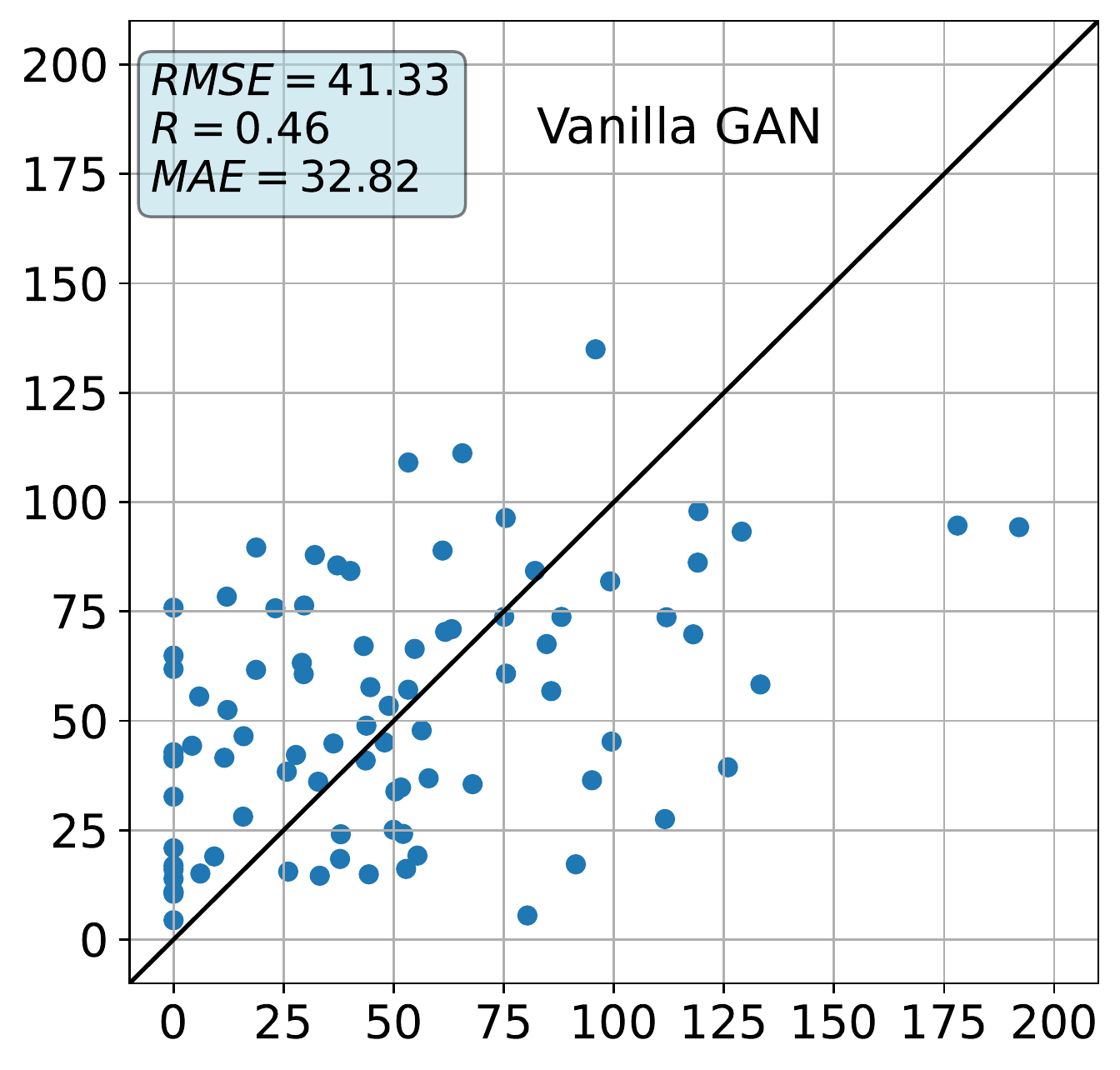}}
\hspace*{-.6cm}                 
 \subfloat[\label{lsgan_scatter}]{\includegraphics[width=0.22\textwidth]{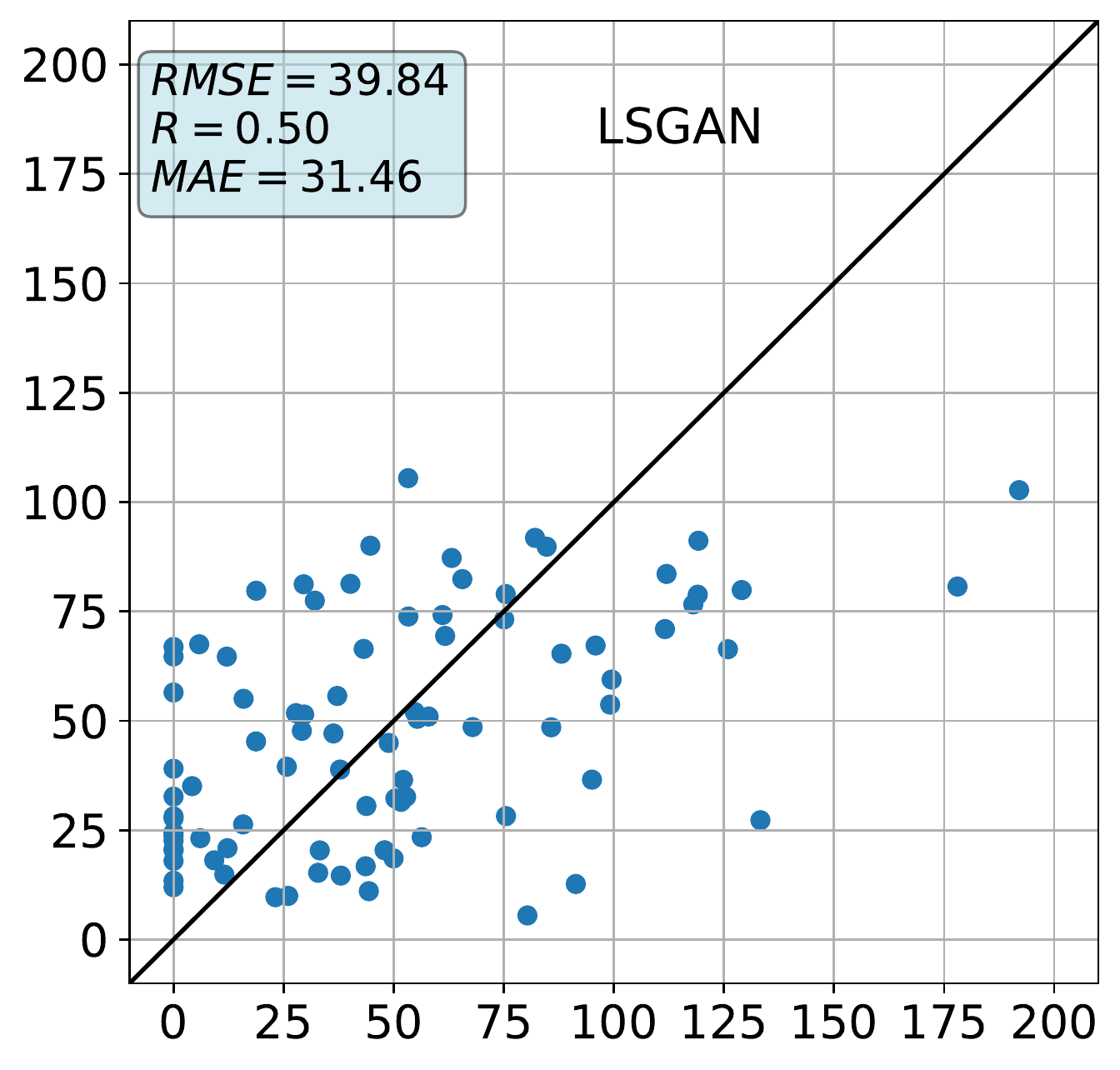}}
\hspace*{-.6cm}
\subfloat[\label{wgangp_scatter}]{\includegraphics[width=0.22\textwidth]{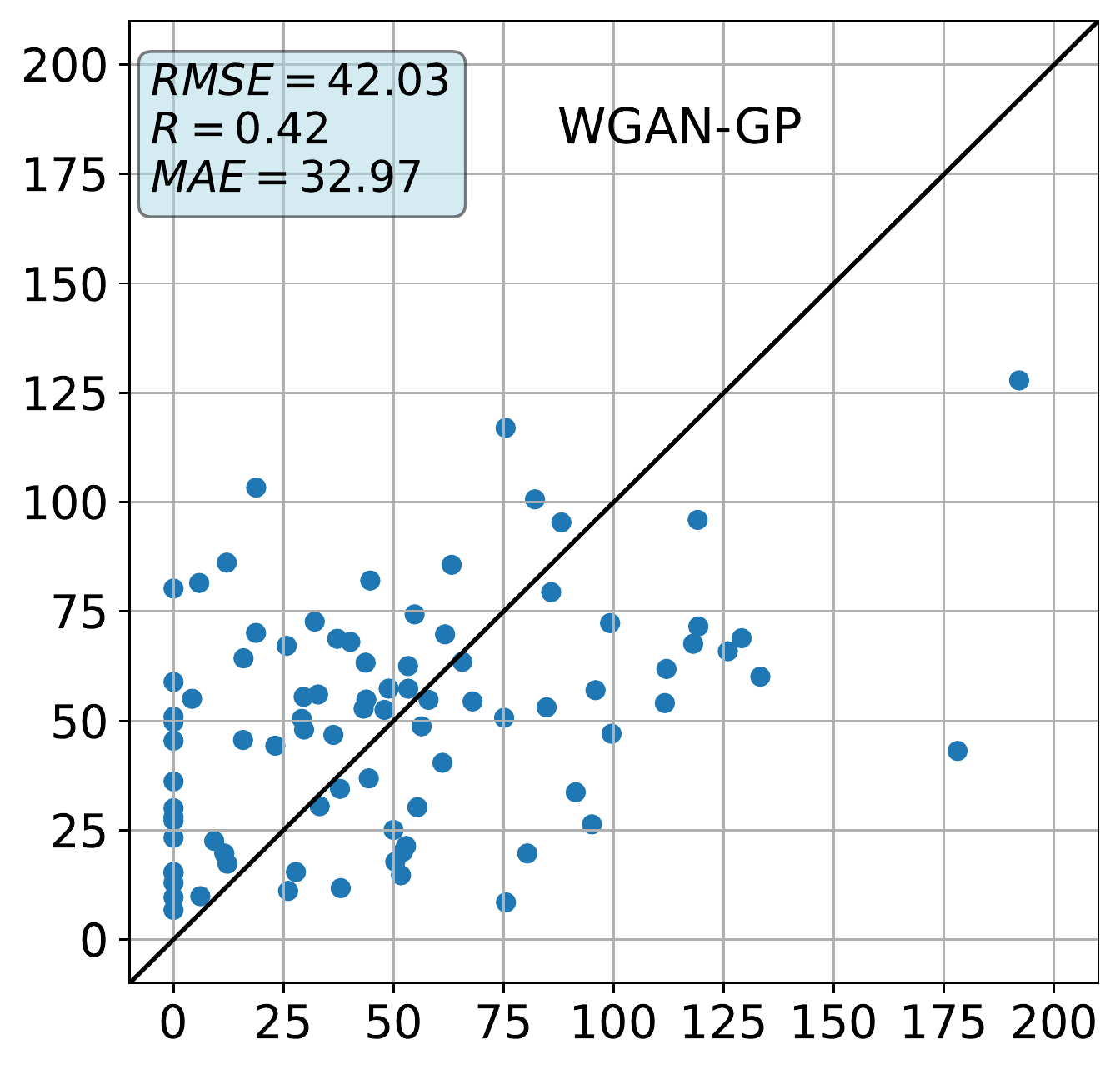}}
\vspace*{.2cm}
\leavevmode\smash{\makebox[0pt]{\hspace{-100em}         
  \rotatebox[origin=l]{90}{\hspace{-1.5em}
    Predicted AGB (Mg ha$^{-1}$)}
}} 
\vspace*{5cm}\hspace*{6.50cm}{Ground Reference biomass (Mg ha$^{-1}$)}\\
\vspace{-5cm}
\caption{Scatter plots between ground reference AGB, $z$, and model-predicted AGB. Model-predicted AGB values are retrieved from the baseline sequential regression model \textbf{(a)}, see \Sec{sec:baseseq}, or the proposed cGAN-based sequential models; Vanilla GAN \textbf{(b)} , LSGAN \textbf{(c)} and WGAN-GP \textbf{(d)}. See \Sec{sec:cganseq} for details on the cGAN-based methods. The black lines are reference lines indicating 100\% correlation between $z$ and predictions.}
    \label{fig:scatter_seq}
\end{figure*} 

\begin{table}[htb]
\caption{Computed Pearson correlation coefficients (R), RMSE and MAE between area-weighted means retrieved from regression models and ground reference plots of AGB ($\mathrm{Mg\,ha}^{-1}$).}
    \label{tab:res_gr_seq}
    \centering
     \begin{tabular}{|l|l|l|l|}
     \hline
    AGB prediction models based on: & R & RMSE & MAE\\ 
    \hline
    ALS$^{a}$ &  0.68 & 33.39 & 24.61 \\
    
    Baseline sequential $^b$ & 0.43 & 41.88 &33.36 \\
    Vanilla GAN$^c$ & 0.46 &  41.33 & 32.82\\
    LSGAN$^c$ & 0.50  & 39.84 & 31.46\\
    WGAN-GP$^c$ & 0.42 & 42.03 & 32.97\\
    \hline 
    \multicolumn{4}{l}{}\\
    \multicolumn{4}{l}{${}^\text{a}$ The non-sequential ALS-based regression model proposed in \cite{naessetMappingEstimatingForest2016}.} \\
    \multicolumn{4}{l}{${}^\text{b}$ The baseline sequential  regression model, proposed in \Sec{sec:baseseq}.} \\
    \multicolumn{4}{l}{${}^\text{c}$ The cGAN-based sequential regression models, proposed in \Sec{sec:cganseq}.}\\
    \end{tabular}
\end{table}

\subsubsection{Sequential model evaluation}\label{sec:seq_eval}
Here, we present results and evaluate the two subsequent models, \textit{g}, that were proposed in \Sec{sec:baseseq} and \Sec{sec:cganseq}. Note that the quantitative results in \tabref{tab:res_seqmod}, and \figr{fig:density_seq} are computed with respect to $\zhaty$, since it in the sequential modelling strategy replace $z$. 

Computed metrics between $\zhatyx$ and $\zhaty$, i.e.\ the Pearson correlation coefficient (R), RMSE and CV-RMSE, for all four sequential models are collected in \tabref{tab:res_seqmod}. Results in \tabref{tab:res_seqmod} indicate that the baseline sequential  model is preferred to the three cGAN-based models as it experiences both a smaller RMSE and CV-RMSE, and a higher R with respect to $\zhaty$. Among the cGAN-based models, the Vanilla GAN is preferred as it achieves the highest correlation and the lowest RMSE to $\zhaty$. However, the Vanilla GAN model also experiences the largest difference between RMSE and CV-RMSE, implying that AGB predictions retrieved from this model are less consistent. 

Generated synthetic AGB prediction maps for the proposed sequential models are shown in \figr{fig:AGBmaps_seq}. The prediction map from the baseline sequential  model is shown in (b), while (c)-(e) show corresponding prediction maps constructed from the cGAN-based models, i.e.\ the Vanilla GAN, LSGAN and WGAN-GP model.
The ultimate goal of the sequential model $g$ is to achieve AGB prediction maps that resemble the $\zhatyb$ prediction map in \figr{als_agb2}. Although the computed metrics for the baseline sequential regression model indicate that this model is prefered to the cGAN-based models, it is unable to capture the dynamic range of ALS-based AGB predictions, see \figr{s1aseq_agb2}. The model's inability to predict near-zero biomass is particularly severe which, similar to model model $h$, can be explained by the square root transform and the bias correction applied. The cGAN-based models are, however, able to predict zero biomass. Their constructed biomass maps also exhibit a higher dynamic range in levels of predicted biomass. All sequential AGB models are generally underpredicting $\zhaty$.

In \figr{fig:density_seq} we visualise density plots between $\zhaty$ and predicted AGB from the proposed sequential AGB regression models. The white lines indicate a reference line for 100\% correlation between $\zhaty$ and $\zhatyx$ . While the baseline model achieves better RMSE and R, the Vanilla GAN model achieves the lowest MAE. We note that all four sequential models struggle to predict $\zhaty$ correctly at low AGB levels. They are generally biased towards overpredicting at low $\zhaty$. While the cGAN based models manage to predict zero biomass, the baseline model can not. Since the baseline model only predicts AGB over 100 $\mathrm{Mg\,ha}^{-1}$ occasionally, it consequently underpredicts high $\zhaty$. The density plots of the three cGAN-based models indicate that they also underpredict high levels of $\zhaty$, but not to the same extent as the baseline sequential  model. 

We also compute the pixel-wise difference between $\zhaty$ and $\zhatyx$, i.e.\ $\zhaty - \zhatyx$, for each proposed sequential models. The pixel-wise differences are visualised in \figr{fig:diffmaps_seq}, where (b) is the difference for the baseline model, (c) for the Vanilla GAN model, (d) for the LSGAN model and (e) for the WGAN-GP model. By comparing the AGB difference maps in \figr{fig:diffmaps_seq} to the actual $\zhatyb$ prediction maps in \figr{fig:AGBmaps_seq}, we again show that all sequential models underpredict AGB in areas with high levels of $\zhaty$ (shown as pink or blue in (b)-(e)). We also highlight that at all sequential models overpredict AGB areas with low levels of $\zhaty$ (shown as green in (b)-(e)). The baseline sequential  model's inability to predict zero or low levels of biomass can probably explain the larger extent of green regions in (b), compared to (c)-(e).

For further comparison we provide sequential modelling results for the few ground reference AGB measurement we have available. We argue that achieving large-scale AGB maps that reflect the dynamic range of $\zhaty$ is one desired goal, but more important is the ability of the AGB predictions to match $z$ values. Thus, we computed the correlation between AGB predictions obtained with the proposed sequential modelling strategy and the 88 ground reference plots, shown with red markers in \figr{fig:map}. Since the physical area of each ground reference plot could intersect with several of the grids with pixel size 707m$^2$, we calculated the area-weighted mean of grid pixels intersecting with each separate ground reference plot. \figr{fig:scatter_seq} shows scatter plots of the correlation between $z$ and model-predicted AGB, retrieved from the sequential models, together with computed metrics: i.e.\ RMSE, R and MAE. Quantitative results derived from \figr{fig:scatter_seq} are also summarised \tabref{tab:res_gr_seq} together with computed metrics for model $f$. Similar to the scatter plot for model $h$, \figr{s1aseq_scatter} also indicate that AGB predictions from the baseline sequential model are bounded between 25-75 $\mathrm{Mg\,ha}^{-1}$. \tabref{tab:res_gr_seq} shows that neither of the proposed sequential models achieves as high correlation or low RMSE and MAE with respect to $z$ that model $f$ achieves. Nevertheless, it should be noted that $f$ \cite{naessetMappingEstimatingForest2016} was fitted against the available $z$. The sequential models, on the other hand, were optimised to achieve $\zhatyx \approx \zhaty$ as they were fitted against $\zhatyb$\footnote{In \Sec{sec:cal} in the Appendix, we experiment with an additional calibration step to further calibrate model $g$ against $z$. Results indicate that post-calibration of the output from $g$ with either gamma or linear calibration increases the accuracy and the correlation by a small amount. Nevertheless, the possible improvement is modest and we omit this additional step as the non-sequential \sen-based model still outperforms the post-calibrated sequential models on computed RMSE, MAE and R.\label{footnote:calibration}}.
While \tabref{tab:res_seqmod} indicates that the baseline sequential regression model predicts $\zhaty$ best, \tabref{tab:res_gr_seq} indicates that both the LSGAN model and the Vanilla GAN model perform better than the baseline sequential model on all three metrics. Additionally, all cGAN-based models obtain lower MAE with respect to $z$ than the baseline sequential model achieves. Among them, the LSGAN model performs best in predicting $z$. Additionally, all cGAN-based models obtain lower MAE with respect to $z$ than the baseline sequential model achieves. Interestingly, by comparing \tabref{tab:res_trad} to \tabref{tab:res_gr_seq} we identify the LSGAN model, in terms of R and RMSE, to perform better in predicting $z$ than the InSAR model. We therefore argue that the LSGAN and the Vanilla GAN model should be the first and second choice if one aims to achieve a model that reflects the dynamic range of the true AGB best.

\begin{figure}[htb!]
\hspace*{1.5cm}
\vspace{-.3cm}
\subfloat[\label{gr_agb_hist}]{\includegraphics[scale=0.3]{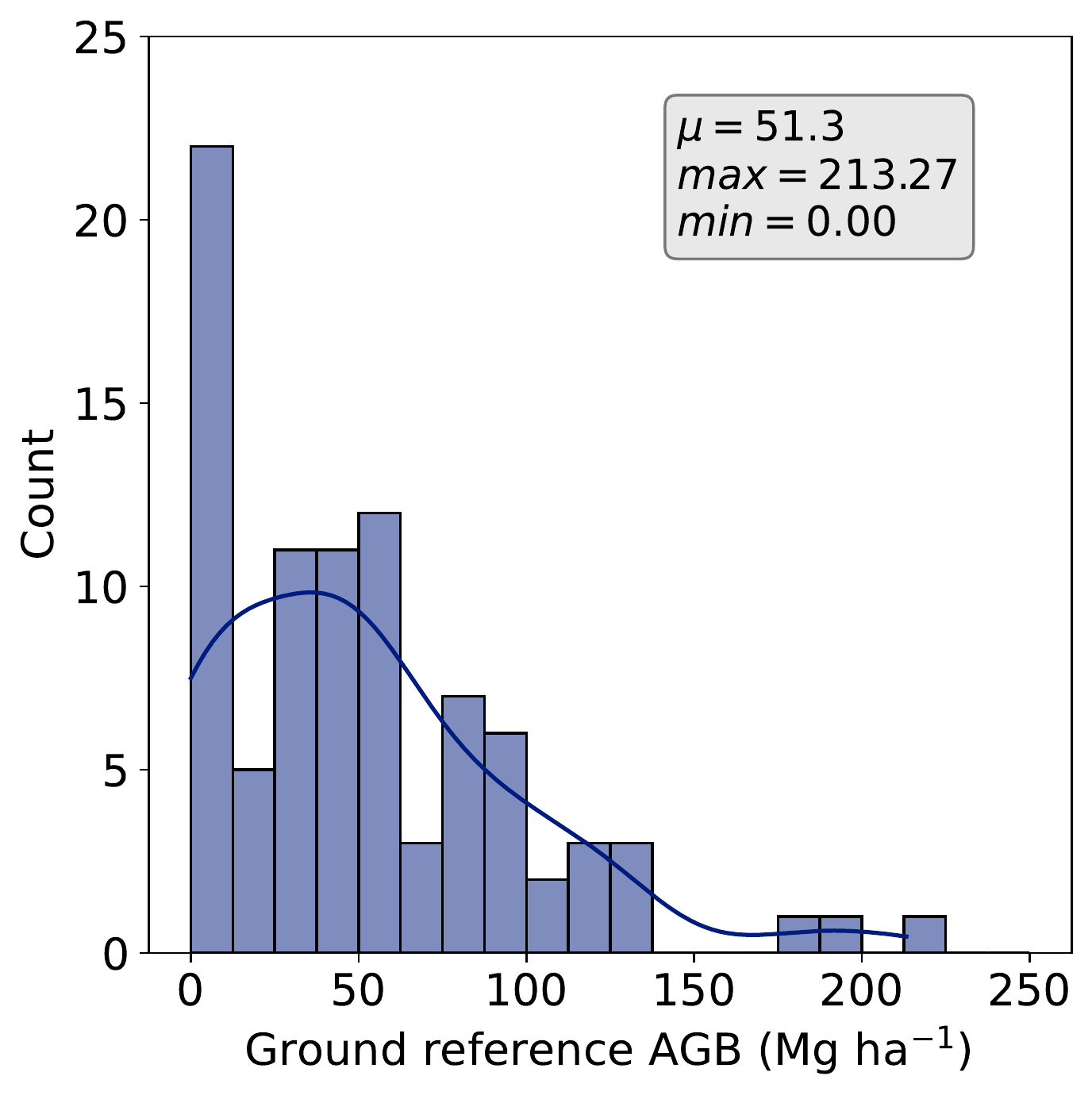}}\\
\vspace{-.3cm}
\centering
\subfloat[\label{s1anonseq_agb_hist}]{\includegraphics[scale=0.28]{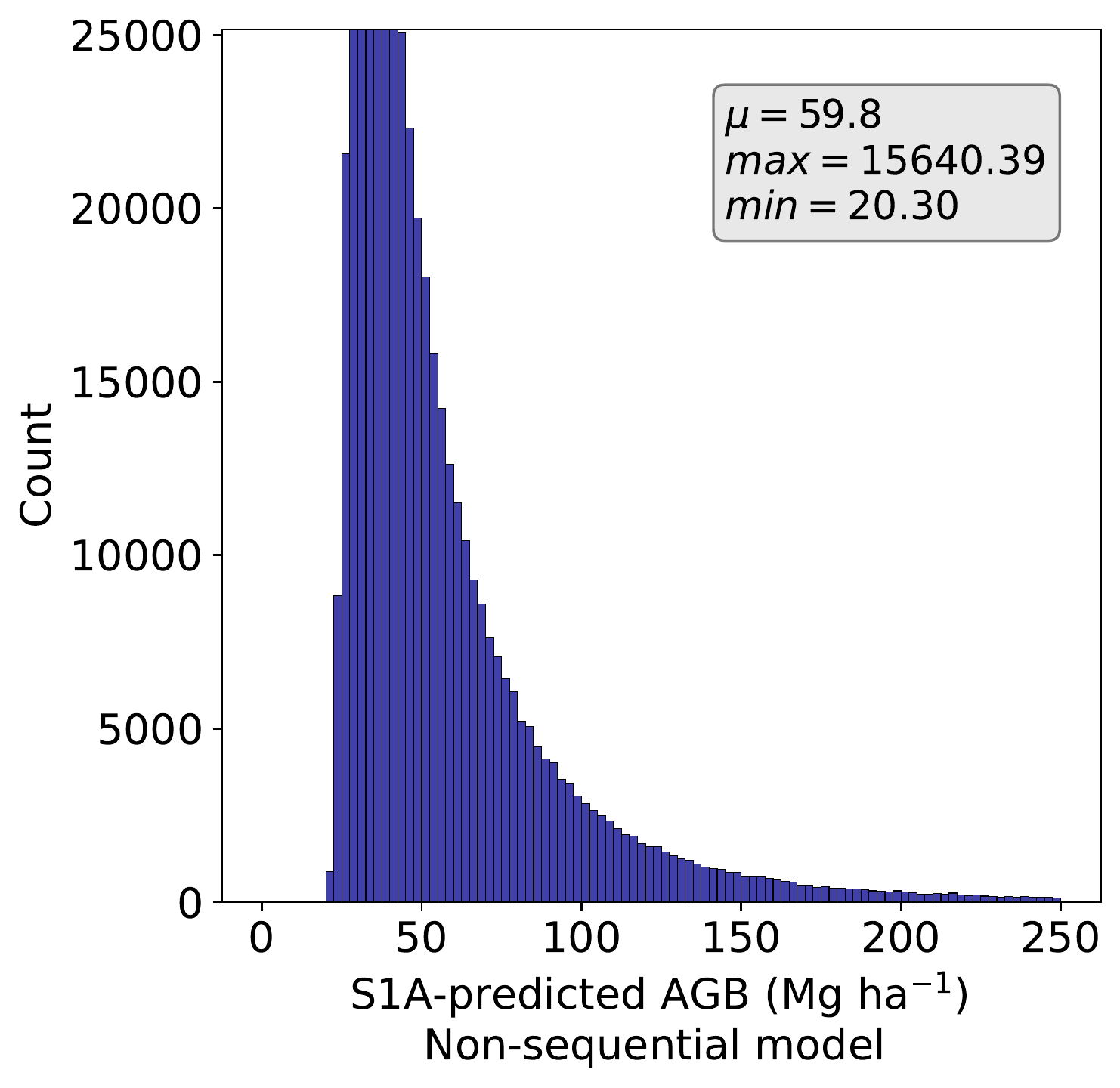}}
\subfloat[\label{s1aseq_agb_hist}]{\includegraphics[scale=0.28]{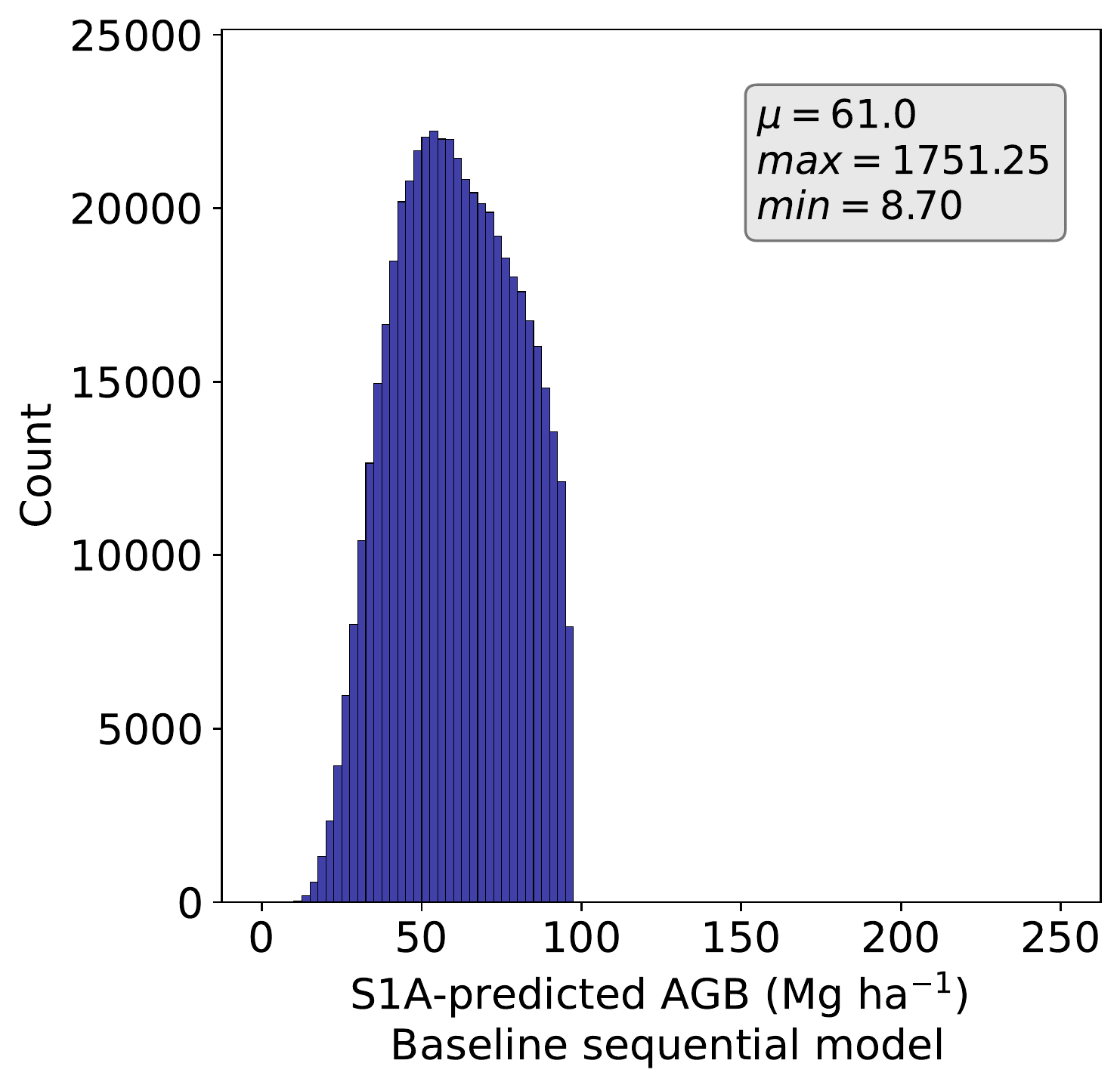}}\\
\subfloat[\label{als_agb_hist}]{\includegraphics[scale=0.28]{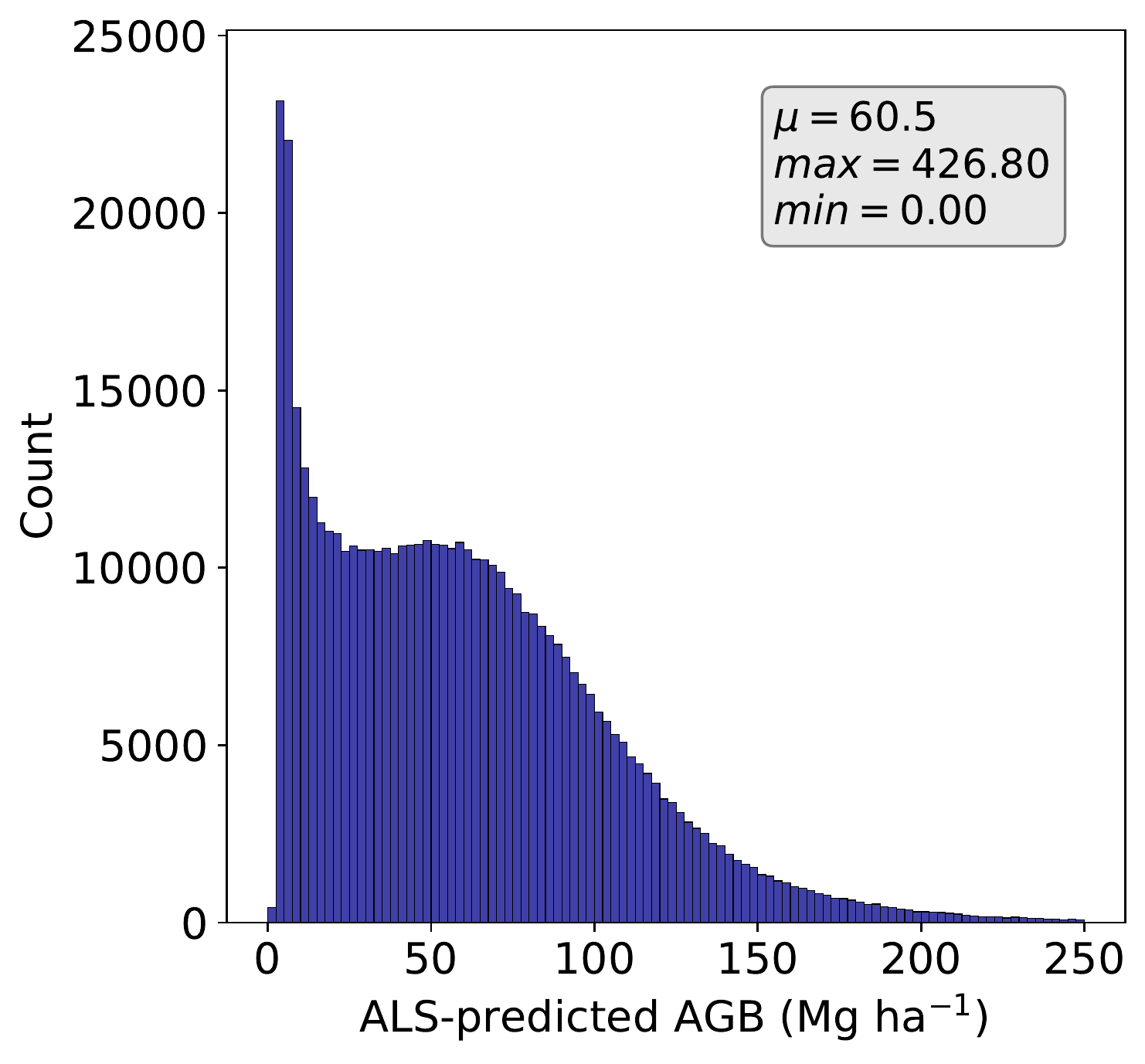}}
\subfloat[\label{vanilla_agb_hist}]{\includegraphics[scale=0.28]{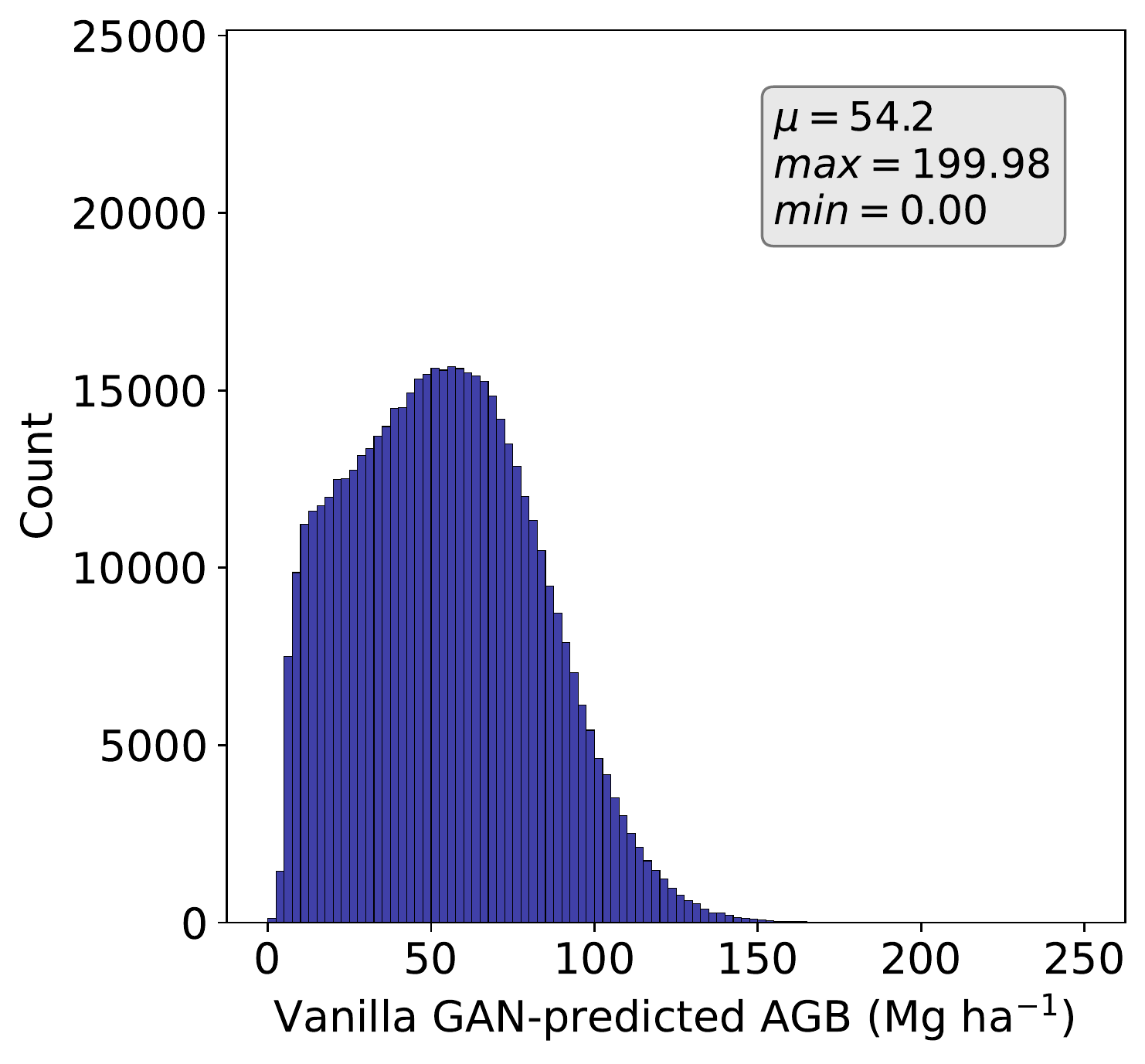}}\\
\subfloat[\label{lsgan_agb_hist}]{\includegraphics[scale=0.28]{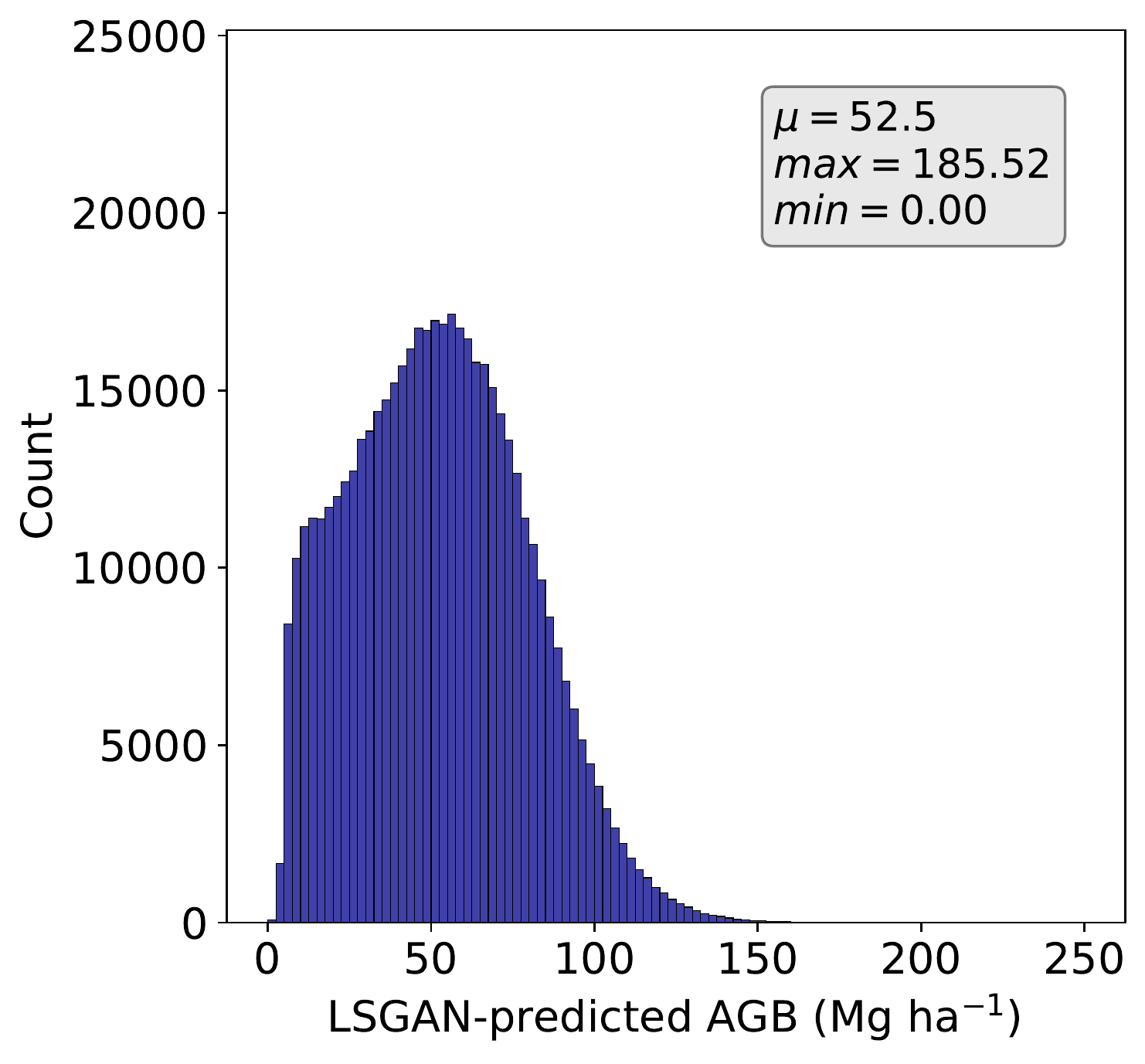}}
\subfloat[\label{wgan_agb_hist}]{\includegraphics[scale=0.28]{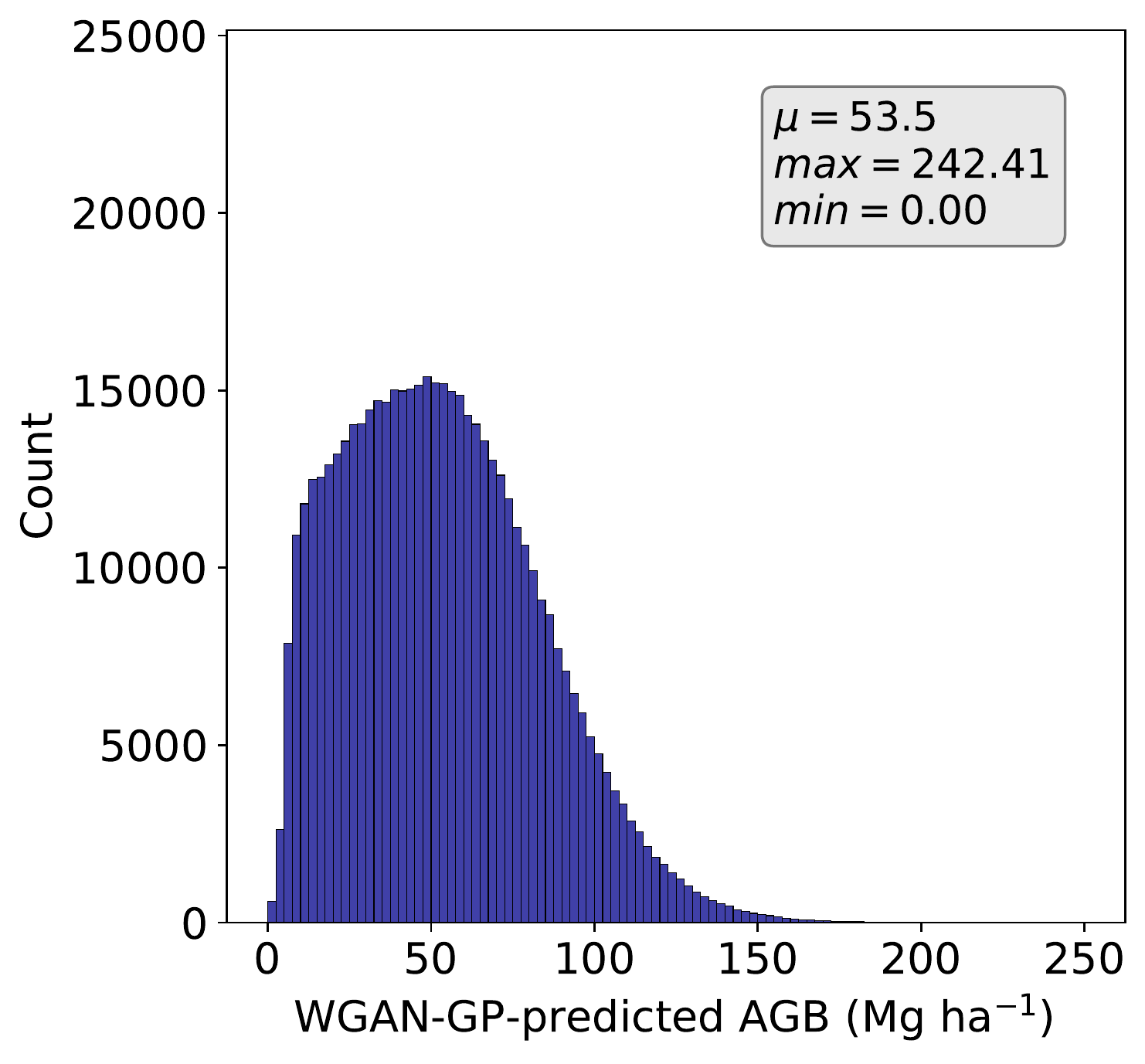}}\\
\caption{Histograms of AGB predictions from the proposed AGB models; non-sequential \sen\ \textbf{(b)}, baseline sequential \textbf{(c)}, Vanilla GAN \textbf{(e)}, LSGAN \textbf{(f)} and WGAN-GP \textbf{(g)}. A histogram over the collected ground reference AGB is shown in \textbf{(a)}, while \textbf{(d)} shows a histogram over ALS-based AGB predictions. Reported metrics are the sample mean, $\mu$, median and maximum and minimum predicted AGB (Mg ha$^{-1}$).}
    \label{fig:hist_seq}
\end{figure}  

\begin{table}[]
\caption{Overall RMSE and RMSE computed for each quartile, i.e. RMSE($Q_{0,1}$), RMSE ($Q_{1,2}$), RMSE($Q_{2,3}$) and RMSE($Q_{3,4}$) (lower is better). The RMSE metrics are computed between AGB prediction maps constructed in this work and the ALS-based AGB prediction map. All metrics are in units of $\mathrm{Mg\,ha}^{-1}$.}
    \label{tab:res_quantiles}
    \centering
    \begin{tabular}{|l|c|c|c|c|c|}
         \hline
   Model    &  RMSE & RMSE                          & RMSE                          & RMSE                          & RMSE                      \\
            &       &  ($Q_{0,1}$)  &  ($Q_{1,2}$)  &  ($Q_{2,3}$)  &  ($Q_{3,4}$)\\  
   \hline
    Non-sequential &  84.5 & 67.0 & 59.8 &86.3  & 114.3\\
    Baseline sequential& 40.8 & 40.0 & 27.2 &19.0  & 63.0\\
    Vanilla GAN &  42.6 & 32.4 & 29.0 &27.6  & 67.7\\
    LSGAN & 43.0 & 31.7 & 27.6 &27.3  & 69.8\\
    WGAN-GP & 44.6 & 34.0 & 30.7 &30.4  & 70.3\\
    \hline 
    \end{tabular}
\end{table}

\begin{table}[]
\caption{Pearson correlation coefficient computed between pixel-wise predictions for pairs of maps visualised in \figr{fig:AGBmaps_trad} (upper table) and \figr{fig:AGBmaps_seq} (lower table). Models referred to as ALS, InSAR, RapidEye, Landsat and PALSAR are retrieved from \cite{naessetMappingEstimatingForest2016}. Remaining models are developed for this work.}
    \label{tab:corr_models_seq}
    \centering
  \begin{tabular}{|l|c|c|c|c|c|}
  \hline
  &{InSAR}&{RapidEye}&{Landsat}&PALSAR&  {\sen}\\
  &       &          &         &             &  {(non-sequential)}\\
  \hline
{ALS}& 0.65 & 0.48 & 0.42 & 0.28&  0.16\\
{InSAR}&  & 0.38 & 0.57 & 0.38 &  0.15 \\ 
{RapidEye}&  &  & 0.31 &0.22 &   0.15\\
{Landsat}&  &  &  & 0.34 &  0.13 \\ 
{PALSAR}&  &  &  &  &   0.02 \\ 
\hline
\hline
\multicolumn{2}{|l|}{} & {Vanilla} & {LSGAN} & {WGAN-GP} & {\sen}\\
\multicolumn{2}{|l|}{}    & {GAN}  &  & &  {(sequential)}  \\
\hline
\multicolumn{2}{|l|}{ALS}   &  0.4 & 0.39 & 0.35 &  0.41        \\
\multicolumn{2}{|l|}{Vanilla GAN} &     & 0.59 & 0.55  &  0.62  \\
\multicolumn{2}{|l|}{LGAN} &     &       & 0.53      &   0.60 \\
\multicolumn{2}{|l|}{WGAN-GP} &     &       &       &  0.59 \\
    \hline
\end{tabular}
\end{table}

\subsection{Non-sequential and sequential modelling}
To broaden the discussion, evaluate the suitability of the \sen\ sensor as a data source for AGB regression models and enable further comparison of the non-sequential and sequential modelling strategies, we provide three additional results: \figr{fig:hist_seq}, \tabref{tab:res_quantiles} and \tabref{tab:corr_models_seq}.

In \figr{als_agb_hist} we show histogram plots over predicted AGB values derived from the ALS-based regression model $f$ together with AGB predictions from models proposed in this work: the non-sequential \sen\ model \figr{s1anonseq_agb_hist}, the baseline sequential model \figr{s1aseq_agb_hist}, the Vanilla GAN model \figr{vanilla_agb_hist}, the LSGAN model \figr{lsgan_agb_hist} and the WGAN-GP model \figr{wgan_agb_hist}. We also show a histogram of measured ground reference AGB, $z$, in \figr{gr_agb_hist} overlaid with a nonparametric estimate of the underlying probability density function. Note the similarities between the distributions of $z$ and $\zhaty$ \cite{naessetMappingEstimatingForest2016} in (b).
Besides not being able to predict low AGB values, see \figr{s1anonseq_agb_hist} and \figr{s1aseq_agb_hist}, both the non-sequential \sen\ model and the baseline sequential model predict some extreme AGB values of 15,640 $\mathrm{Mg\,ha}^{-1}$ in \figr{s1anonseq_agb_hist} and 1,751 $\mathrm{Mg\,ha}^{-1}$ in \figr{s1anonseq_agb_hist}, which neither of the cGAN-based models do. Instead, the maximum predicted AGB from the three cGAN-based AGB models are rather close to the maximum measured AGB in the field plots, i.e.\ 213.4 $\mathrm{Mg\,ha}^{-1}$ \cite{naessetMappingEstimatingForest2016}. Also, all cGAN-based models behave more similar to $z$ and $f$ for middle-to-high levels of AGB, see \figr{vanilla_agb_hist}-g compared to \figr{gr_agb_hist} and \figr{als_agb_hist}. This could indicate that the more complex cGAN-based models have learned AGB dynamics of $z$ and $f$ better in middle-to-high levels of AGB, than the simpler non-sequential and baseline sequential model manages.

To emphasise where the proposed models are more or less consistent with the ALS-based AGB prediction map, we evaluate AGB predictions from the five models against $\zhatyx$ in terms of overall RMSE, and RMSE computed for each quartile. Results provided in \tabref{tab:res_quantiles}  clearly show that AGB predictions from the non-sequential \sen\ model deviate most from $\zhatyx$, both overall and in each quartile. The baseline sequential  model is most similar to  $\zhatyx$ in the second and third quartile and achieves the smallest RMSE among all five proposed models in the fourth quartile. As expected from the histograms in \figr{fig:hist_seq} and the constructed AGB prediction maps in \figr{fig:AGBmaps_seq}, \tabref{tab:res_quantiles} show that all cGAN-based models produce low RMSE in the first quarter quartile, with the LSGAN model being better than the Vanilla GAN model. Among the cGAN-based models, the Vanilla GAN model only receives the smallest RMSE in the fourth quartile. Once again, it is shown in \tabref{tab:res_quantiles} that the WGAN-GP model is the worst among the cGAN-based models.

\tabref{tab:corr_models_seq} shows the Pearson correlation coefficient computed between pixel-wise AGB predictions for pairs of maps from either \figr{fig:AGBmaps_trad} or \figr{fig:AGBmaps_seq}. Correlations computed between AGB predictions retrieved from the non-sequential models are listed in the upper part of \tabref{tab:corr_models_seq}, while correlations computed between AGB predictions from the sequential models and the surrogate regression target, i.e.\ $\zhaty$, are combined in the lower part of the table. Previous results from \cite{naessetMappingEstimatingForest2016} identified the ALS-based AGB prediction map and the InSAR-based AGB map to have the greatest correlation with each other, see \tabref{tab:corr_models_seq}, and with $z$, see \tabref{tab:res_trad}. AGB predictions from the Landsat- and PALSAR-based models achieved the smallest correlation with $z$, see \tabref{tab:res_trad}. The proposed non-sequential model $h$ achieves by far the lowest correlations with any of the other five non-sequential AGB models, see \tabref{tab:corr_models_seq}. This is probably a consequence of the \sen-based model's inability to predict low biomass levels. In e.g.\ the left part of the AOI, see \figr{fig:AGBmaps_trad}, the ALS-, InSAR- and the RapidEye-based AGB models predict AGB around 0 $\mathrm{Mg\,ha}^{-1}$ in approximately the same areas, while predicted AGB levels retrieved from the non-sequential \sen-based model deviates highly in the same areas. Note that all sequential models achieve a much higher correlation with model $f$ than what model $h$ achieves. Logically, this could be explained by the fact that all sequential models were fitted against $f$. While the non-sequential \sen-based model $h$, the InSAR model and the ALS model $f$ achieve the highest correlations and lowest RMSE with respect to $z$, the surprisingly low correlation between $h$ and $f$ is notably. It could imply that model $h$ is overconfident on the small set of $z$ measurements.  Among the sequential models, \tabref{tab:res_seqmod} and \tabref{tab:corr_models_seq} show that the proposed baseline model achieves the lowest RMSE and highest correlation coefficient with respect to $\zhaty$. Furthermore, the cGAN-based model trained with the WGAN-GP objective function achieves the smallest correlation with $\zhaty$, see \tabref{tab:corr_models_seq}. Overall, the correlations between the sequential models and the ALS-based model $f$ are all higher than the corresponding correlation between AGB  predictions from $f$ and the PALSAR model, and similar to the correlation between AGB  predictions from $f$ and the Landsat model. In addition to the discovery that the LSGAN model performs better than the non-sequential InSAR model in predicting $z$, these results suggest that the cGAN-based sequential modelling approach and the use of \sen\ data for AGB prediction are worth pursuing further.

\section{Discussion}\label{sec:dis}
The focus of this work has been to develop non-sequential and sequential regression models based on C-band SAR for AGB prediction in Tanzania. One main advantages of utilising \sen\ data as regressors is that it enables frequent and affordable updates of an AGB map with extensive coverage. This approach has a low cost compared to keeping the most accurate prediction model from \cite{naessetMappingEstimatingForest2016} up-to-date by repeated acquisition of ALS data. Our results show that the proposed non-sequential \sen-based model $h$ and the sequential LSGAN model best provide AGB predictions close to measured ground reference AGB, $z$. Only the ALS and the RapidEye-based model in \cite{naessetMappingEstimatingForest2016} perform  better on this task. Noteworthy, in terms of R and RMSE, both the model $h$ and the sequential LSGAN model were identified to be more accurate than the InSAR-based non-sequential model on the same task. Since the InSAR-based model provides estimates of canopy height that are highly correlated with AGB \cite{Kaasalainen2015, Galidaki}, we expected it to be superior in predicting $z$.
We emphasise that we are training all our models using C-band SAR intensity data, which have previously been shown to suffer from much lower saturation levels than e.g\ the L-band ALOS PALSAR sensor. As C-band data neither penetrates as deeply into the forest volume as L-band data, nor can it compete with the accuracy of AGB estimates produced from optical data \cite{dobson1992dependence, le1992relating, luSurveyRemoteSensingbased2016, zolkos}, it has traditionally been considered an inferior information source for AGB estimation. Thus, we have in this work demonstrated the potential of using \sen\ data for AGB prediction and suggest further research on \sen-based models for AGB retrieval.

Formally, the proposed models were assessed in terms of their relative accuracy on AGB prediction with respect to model $f$, \cite{naessetMappingEstimatingForest2016}, and available AGB in situ measurements. However, whenever a certain methodology is implemented for operational purposes in an MRV system, the ultimate goal is to produce estimates of carbon stocks and changes. Among these, estimates for the AGB pool are essential. Further, the Intergovernmental Panel on Climate Change specifies that results should be reported as inferences in the form of confidence intervals \cite{ipcc} (p. 1.10). Thus, although the maps themselves can be useful, for example, to identify critical areas of carbon loss, the prediction map is just an intermediate product on the way to estimating the carbon budget.
AGB can easily be estimated from the prediction maps constructed by the current methods by aggregating individual pixel values. Estimating the uncertainty of AGB estimates in the form of variances or confidence intervals for non-parametric methods such as ANNs, support vector machines, random forest regression and other techniques is a current research issue. To provide such estimates was beyond the scope of the current study. Recent applications of e.g.\ bootstrap resampling for random forest-based prediction models demonstrate that such variance estimators may easily be adopted for ANN models as well, see e.g.\ \cite{Esteban2019Bootstrap}. However the computational burden will be substantial.

By approaching AGB prediction through sequential modelling with ALS-based predictions as a surrogate for $z$, deep contextual models could be utilised for the regression task. As far as we know, this is the first time that contextual cGAN models have been used to simulate ALS-based AGB prediction maps from \sen\ data. 
A natural question is whether DL-based approaches for AGB predictions are worth further investigation, especially since they are more complex to train than traditional statistical regression models. We would argue that more research is needed in utilising contextual DL models to retrieve biophysical parameters from RS data. We have shown that the LSGAN model performs well and reproduces dynamic AGB levels more realistically than simpler non-contextual models. Despite this, the cGAN-based models fall behind the traditional sequential and non-sequential models on RMSE with respect to ground reference data. The trade-off between perceptual quality and reconstruction accuracy is known from the research field of single image super-resolution (SR), \cite{Blau2015, yang_deep_2019, ESRGAN, Ledig, Soh2019SR}. SR in RS data has been studied in e.g.\ \cite{SR_RS1, SR_RS_3, SR_RS_2}. For future work on AGB prediction by DL regression, it appears relevant to incorporate ideas from the field of SR and investigate additional architectures and balancing of GAN losses against traditional $L_1$ and $L_2$ losses for reconstruction. The purpose would be to obtain a model that focuses on the reconstruction loss, yet produces AGB prediction maps that maintain local dynamics.

\subsection{Error discussion}
The accuracy of the proposed models is influenced by many factors, such as the radiometric accuracy of radar images, time lag, and error propagation through the model sequence. The latter was also pointed out in  \cite{englhartAbovegroundBiomassRetrieval2011}. We refer to the time lag as the time difference between collecting the field inventory data in January-February 2014, the acquirement of ALS measurements in 2014 and the acquisition of the Sentinel-1 scene in September 2015. Possible inaccuracies may propagate, first when the ALS model upscales the field inventory data to a $\zhatyb$ map, and secondly when the sequential models are trained. Additional factors that may affect the overall accuracy is resampling of the \sen\ scene to the same grid as the ALS-based AGB prediction map or the image blendingprocess which is applied to construct the full cGAN-based AGB prediction maps from a set of patches. Despite this, the advantage of using a sequential modelling approach on \sen\ data is the ability to achieve biomass prediction maps with high update frequency on a national level. 
Our sequential approach also has potential use in biomass change detection, where the relative change of biomass from one time to another is of higher interest than the absolute AGB values.

As previously mentioned, $z$ was collected within circular sample plots, the most common plot shape in boreal and temperate forest sampling \cite{Tomppo2014ASampling}. However, all remotely sensed datasets used in \cite{naessetMappingEstimatingForest2016} and this work are represented by square pixels. Therefore, using circular field plots is suboptimal, as each model's correspondence to $z$ needs to be computed by an area-weighted mean of neighbouring pixels. The sequential models are not directly related to the circular plots, but through the $\zhaty$, which was trained against $z$. Nevertheless, when computing the correspondence between the sequential model's AGB predictions and $z$, the above challenge arises when the area-weighted mean between square pixels intersecting a circular pixel is computed. In the end, this will influence the overall accuracy of the models. Note that the sampling design in \cite{Tomppo2014ASampling} was optimised for field-based estimation of AGB given a limited budget for inventories, not for upscaling supported by RS, in which case the species diversity and spatial variability of AGB in the miombo woodlands imply that larger sample plots should be used. We sustain \cite{Naesset2016PlotSize}, which concludes that decisions regarding the sample plot size, and thereby its shape, is one of many parameters that have to be considered in future field-based surveys if one aims to enhance estimation through the use of remotely sensed data.

\section{Conclusion}\label{sec:conclusion}
The focus of this work was to investigate the suitability of \sen-based models for AGB prediction in an MRV system for miombo forests in Tanzania. Previously, N\ae sset~\emph{et~al.} \cite{naessetMappingEstimatingForest2016} developed traditional non-sequential regression models for AGB in a Tanzanian AOI using either ALS, TanDEM-X InSAR, RapidEye, Landsat or PALSAR data with a limited amount of ground reference AGB data. The ALS-based AGB predictions achieved the highest accuracy, but the cost and infrequent update of ALS data prevent this model from being of practical use in an MRV system. Therefore, we turned to freely available and easily accessible \sen\ data for this work and developed three different models for AGB prediction from this source: a traditional non-sequential model, a  baseline sequential model and a deep learning-based sequential model. We compared each model's accuracy on the AGB prediction task. Additionally, maps of AGB predictions were compared and evaluated with respect to their ability to recreate realistic biomass dynamics. The model performances and most important results are summarised below.

\subsubsection{Non-sequential \sen\ model} This model was, as the models in \cite{naessetMappingEstimatingForest2016}, trained against the limited ground reference data. Its performance can therefore be directly compared to the results in \cite{naessetMappingEstimatingForest2016}. Among all models proposed for this work, this model achieves the lowest RMSE and highest correlation coefficient (R) against ground reference data. Although this model cannot predict AGB levels between 0-20 $\mathrm{Mg\,ha}^{-1}$, it performs better than the InSAR-based model in terms of R and RMSE. It is only beaten by the ALS-based and the RapidEye-based models \cite{naessetMappingEstimatingForest2016}. However, the non-sequential \sen\ model achieves the highest RMSE in a pixel-by-pixel comparison with the ALS-predicted AGB map. Hence, we conclude that it sacrifices a more realistic prediction of the dynamic range and local variability of AGB values to meet the goal of producing a low RMSE against ground reference data.

\subsubsection{Sequential models}These were developed to enable AGB prediction on a larger scale through a modelling strategy with two subsequent regression models. We propose to employ the ALS-based model \cite{naessetMappingEstimatingForest2016} as the first model. The second model in the chain is trained to relate SAR backscatter images to ALS-based AGB prediction maps, which are used as a surrogate for ground reference data. The baseline sequential model applies a traditional statistical regression model also in the second stage. The alternative sequential model instead uses a deep neural network for cross-modal image-to-image translation, i.e.\ the Pix2Pix cGAN architecture \cite{isola} with some modifications warranted by the application. This cGAN architecture generates synthesised ALS-based AGB predictions during model fitting by conditioning on SAR backscatter data. In contrast to the other models, it uses contextual information from pixel neighbourhoods in its predictions. The baseline sequential model, followed by the Vanilla GAN model, achieve the highest R and lowest RMSE against the ALS-based AGB predictions. Conversely, the LSGAN model is best among the sequential models at reproducing ground reference data, and is only beaten by the ALS-based and RapidEye-based models from \cite{naessetMappingEstimatingForest2016}. In this respect, the LSGAN model achieves slightly higher RMSE and lower R than the non-sequential \sen-based model. However, the LSGAN model can predict AGB levels around 0 $\mathrm{Mg\,ha}^{-1}$ and also achieves higher correlation with the ALS-based predictions. Thus, the contextual cGAN-based models seem to better capture the dynamic range and local variability of AGB.

We have in this research demonstrated the potential of utilising \sen\ data for AGB prediction in Tanzania. Although C-band \sen\ data traditionally have been considered an inferior information source for AGB estimation due to low penetration of the canopy, our results show that \sen-based models are a viable alternative for forest AGB retrieval, especially considering that the data are freely available.

\section*{Acknowledgments}
\label{sec:ack}
We gratefully acknowledge employees of the Tanzania Forest Services Agency, Sokoine University of Agriculture, Norwegian University of Life Sciences, and the Swedish University of Agricultural Sciences (SLU) for participation in field work and provision of in situ measurements, remote sensing data and derived AGB products. Special thanks to professor Håkan Olsson for providing access to ALS data acquired by SLU and for comments on the manuscript. S.\ Björk and S.N.\ Anfinsen further acknowledge discussions with and input from colleagues at the UiT Machine Learning Group.

\ifCLASSOPTIONcaptionsoff
  \newpage
\fi



\bibliographystyle{IEEEtran}
\begin{footnotesize}
\bibliography{bibtex/bib/references.bib}

%
\vskip -2\baselineskip plus -1fil
\begin{IEEEbiography}[{\includegraphics[width=1in,height=1.3in,clip,keepaspectratio]{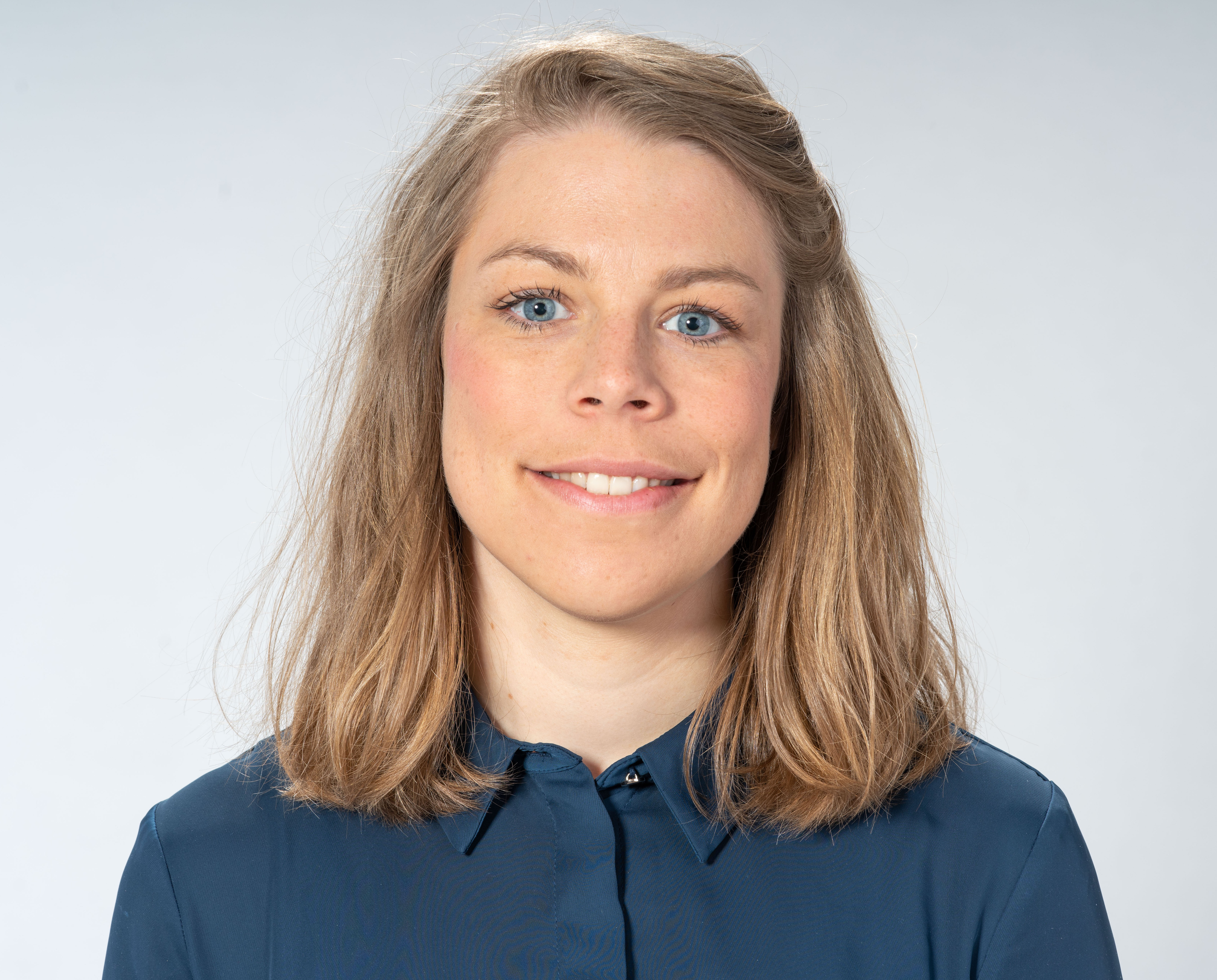}}]{Sara Björk}
received the M.Sc.\ in Applied Physics and Mathematics from UiT The Arctic University of Norway, Troms\o, Norway, in 2016. Since 2017 she has been working towards the Ph.D.\ degree in the Machine Learning Group at the Department of Physics and Technology, UiT. Since May 2022 she is also affiliated with KSAT Kongsberg Satellite Services as a system developer in the DevOps team Applied Deep Learning. Her research interests include image processing, machine learning, deep learning, and generative methods, with emphasis on information extraction from remote sensing data.  
\end{IEEEbiography}




\vskip -2\baselineskip plus -1fil
\begin{IEEEbiography}[{\includegraphics[width=1in,height=1.25in,clip,keepaspectratio]{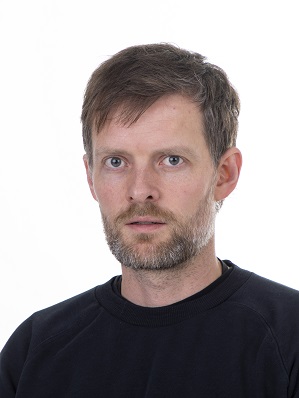}}]{Stian Normann Anfinsen}
received the M.Sc.\ degree in communications, control and digital signal processing from the University of Strathclyde in Glasgow, UK (1998) and the Cand.mag.\ (1997), Cand.scient.\ (2000) and Ph.D.\ degrees (2010) from UiT The Arctic University of Norway in Tromsø, Norway (UiT). Since 2014 he has been a faculty member at the Department of Physics and Technology at UiT, formerly with the Earth Observation Group and currently as a Professor in the Machine Learning Group. His main affiliation since August 2021 is with NORCE Norwegian Research Centre as a Senior Researcher.  His research interests are in statistical modelling, pattern recognition and machine learning algorithms for image, graph and time series analysis in earth observation and energy analytics.
\end{IEEEbiography}

\vskip -2\baselineskip plus -1fil
\begin{IEEEbiography}[{\includegraphics[width=1in,height=1.25in,clip,keepaspectratio]{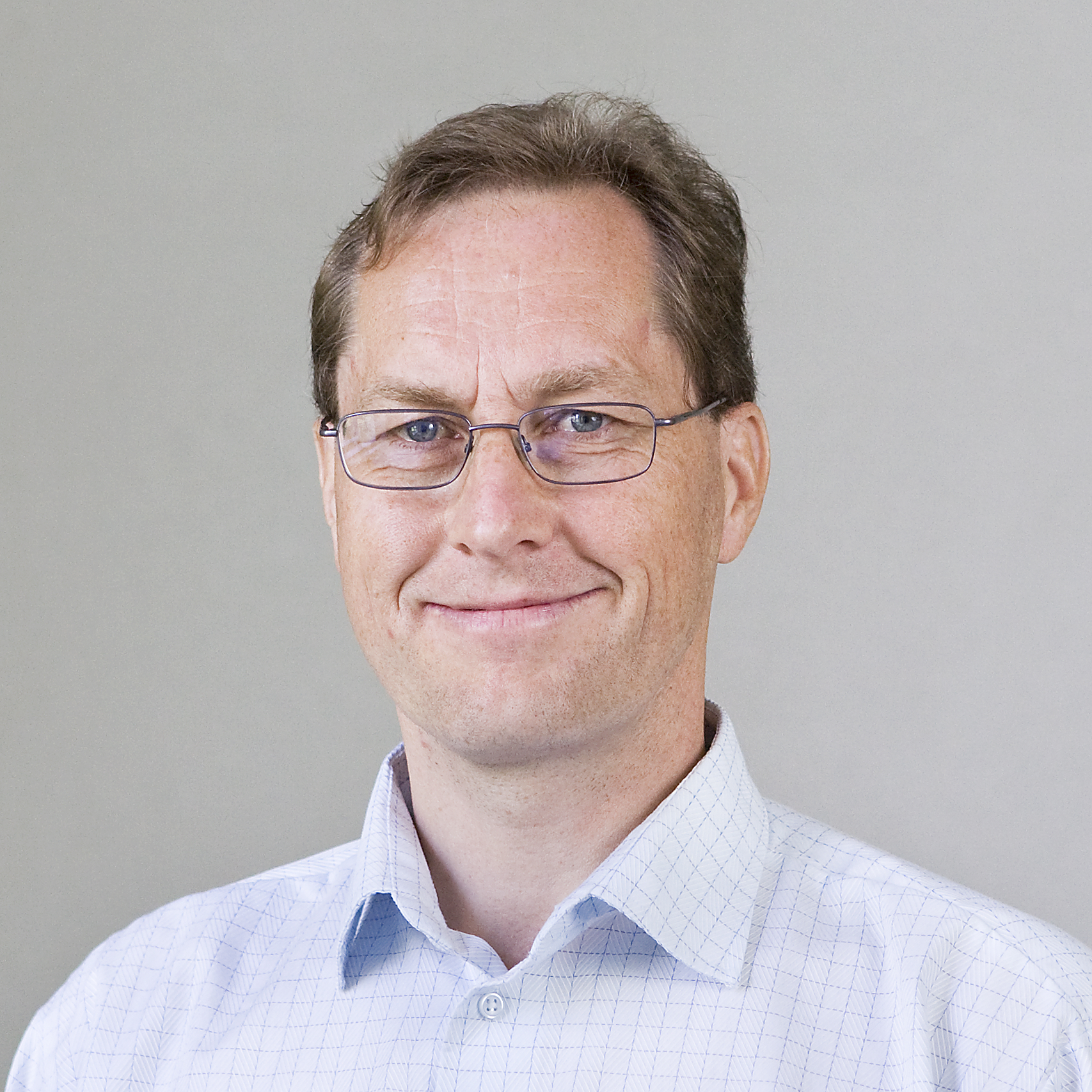}}]{Erik N\ae sset}
received the M.Sc. degree in forestry and the Ph.D.\ degree in forest inventory from the Agricultural University of Norway, Ås, Norway, in 1983 and 1992, respectively. His major field of research is forest inventory and remote sensing, with particular focus on operational management inventories, sample surveys, photogrammetry, and airborne LiDAR. He has played a major role in developing and implementing airborne LiDAR in operational forest inventory. He has been the Leader and Coordinator of more than 60 research programs funded by the Research Council of Norway, the European Union, and private forest industry. He has published around 250 papers in international peer-reviewed journals. His teaching includes lectures and courses in forest inventory, remote sensing, forest planning, and sampling techniques.
\end{IEEEbiography}

\vskip -2\baselineskip plus -1fil
\begin{IEEEbiography}[{\includegraphics[width=1in,height=1.25in,keepaspectratio]{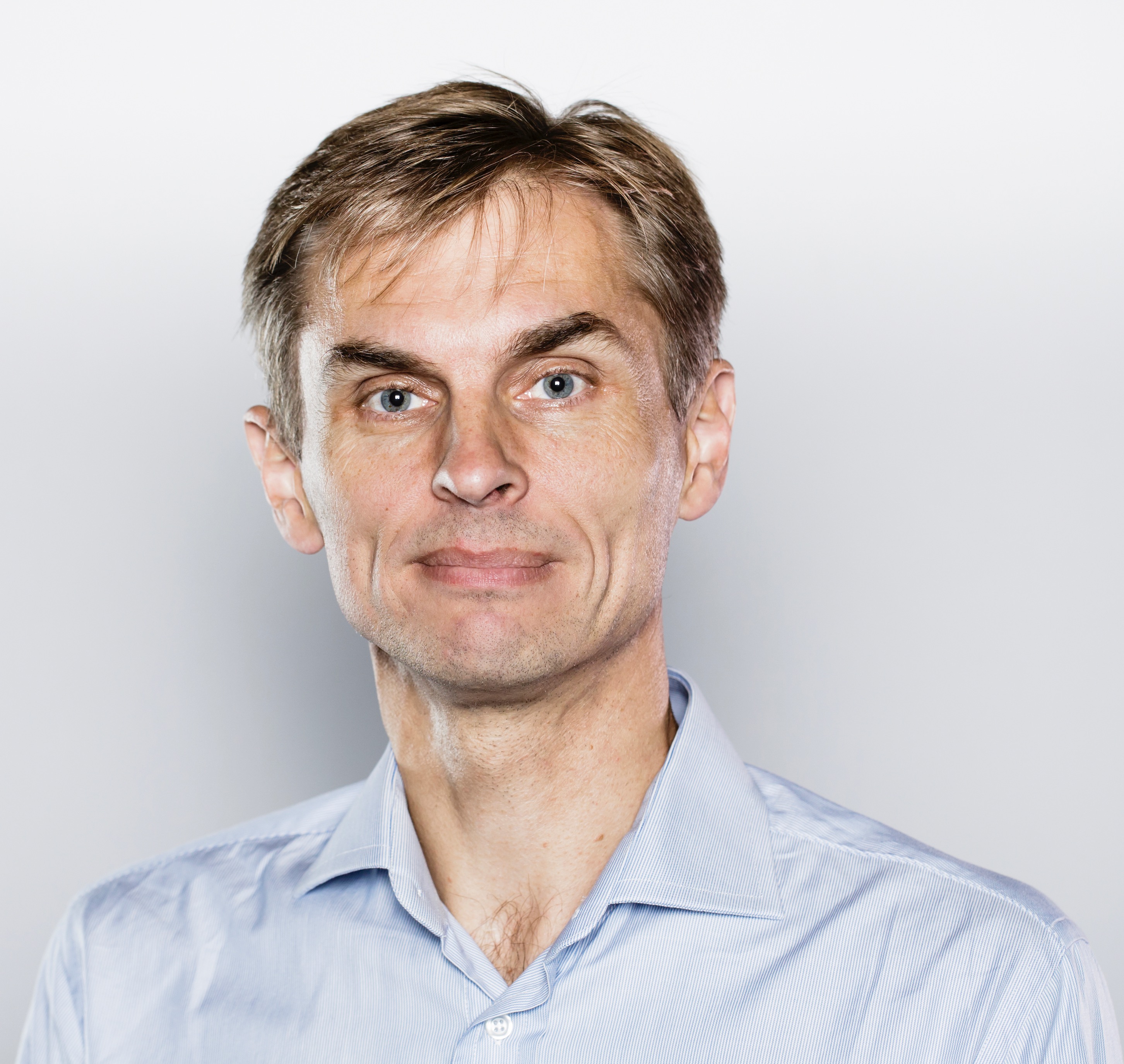}}]{Prof. Terje Gobakken}
is professor in forest planning and he has published more than 190 peer-reviewed scientific articles related to forest inventory and planning in international journals. He has been working at the Norwegian National Forest Inventory and participated in compiling reports of emissions and removals of greenhouse gases from land use, land-use change and forestry in Norway. He has coordinated and participated in a number of externally funded projects - including international projects funded by for example NASA and EU (FP6 and FP7), and has broad practical and research-based experience with development of big data and information infrastructures for forest inventory, planning and decision support.
\end{IEEEbiography}

\vskip -2\baselineskip plus -1fil
\begin{IEEEbiographynophoto}
{Eliakimu Zahabu}
\end{IEEEbiographynophoto}


\vfill

\end{footnotesize}
\newpage
\appendix
\section*{} \label{sec:app}
This appendix includes a specification of the modifications done to the \textit{Pix2Pix} architecture \cite{isola} to make it suitable for generation of synthetic ALS-based AGB image patches in our sequential modelling strategy. It also provides additional experiments and results that were conducted for this work.

\subsection{Modified Pix2Pix architecture}\label{sec:mod_pix2pix}
The cGAN-based sequential model used for generation of synthetic ALS-based AGB image patches, $\zhatyxb$, is based on the image-to-image translation framework \textit{Pix2Pix} \cite{isola}. To meet our needs, we modified it in the following ways:
\begin{enumerate}
     \item We enable the use of calibrated pixel values read from image files in GeoTIFF format. This is necessary since we work with images with pixel values that carry information about physical entities and represent either calibrated $\sigma_0$ values (backscatter coefficients) or AGB predictions measured in $\mathrm{Mg\,ha}^{-1}$.
     \item We change the activation function in the output layer from a hyperbolic tangent (tanh) function used in \cite{isola} to a rectified linear unit (ReLU) activation function. In an earlier phase of this work \cite{Sara2020}, we noticed that the tanh activation function we used in the output layer generated AGB values that overestimated the ALS-based AGB predictions from \cite{naessetMappingEstimatingForest2016}, and particularly failed to predict AGB values close to zero. An essential criterion for our cGAN regression model is that it should be able to predict zero biomass to correlate well with AGB ground reference data, $z$, in non-vegetated areas. The overprediction observed in \cite{Sara2020} can be explained by the nature of the tanh activation function. As the range of the tanh function is [0.0, 1.0], it implies that all data introduced to the cGAN need to be normalised to the same range. The tanh function must output exactly zero to predict zero biomass, which only happens when the action potential goes to $-\infty$. This explains why prediction with the tanh function seems to clip the AGB values at a level higher than zero.
 \end{enumerate}
In conclusion, by substituting the tanh activation function with a ReLU function in the output layer and allowing the regression target to be calibrated AGB values in $\mathrm{Mg\,ha}^{-1}$ units, instead of being normalised to [0.0, 1.0], our modified Pix2Pix architecture no longer overestimates AGB that should be close to zero.  

\subsection{Experiment 1: A study of the impact of speckle filtering and choice of discriminator network}\label{exp:data}

A common preprocessing step for SAR products is speckle filtering. Speckle filters reduce the effects of the inherent speckle phenomenon on the product and smooths the pixel values. In this experiment, we evaluate if speckle filtering of the \sen product affects the accuracy and the quality of cGAN-generated AGB predictions. To this end, we created two different datasets from the \sen\ GRD product: The first was produced by following the SAR processing workflow defined in \Sec{sec:preprosar}; For the second dataset, we used the refined Lee filter \cite{refinedlee} with SNAP's default window size of $7 \times 7$ to apply speckle filtering between step 4) and 5) in the same workflow. We refer to them as the \sen\ dataset with and without speckle filtering. A separate cGAN network was trained on each.

Additionally, we evaluated the three discriminator networks $\D$ presented in \Sec{sec:D} against each other to assess their effect on cGAN performance for data generation. For all experiments in this section, we trained the cGAN for 200 epochs using a ResNet-6 network, WGAN-GP objective function, batch size (BS) of 2, layer normalisation (LN) for $\D$ and batch normalisation (BN) for $\G$. These settings were determined by the model validation results presented in \cite{Sara2020}.  

\begin{figure}[htb]
    \centering
    \includegraphics[scale=.4, width=\columnwidth]{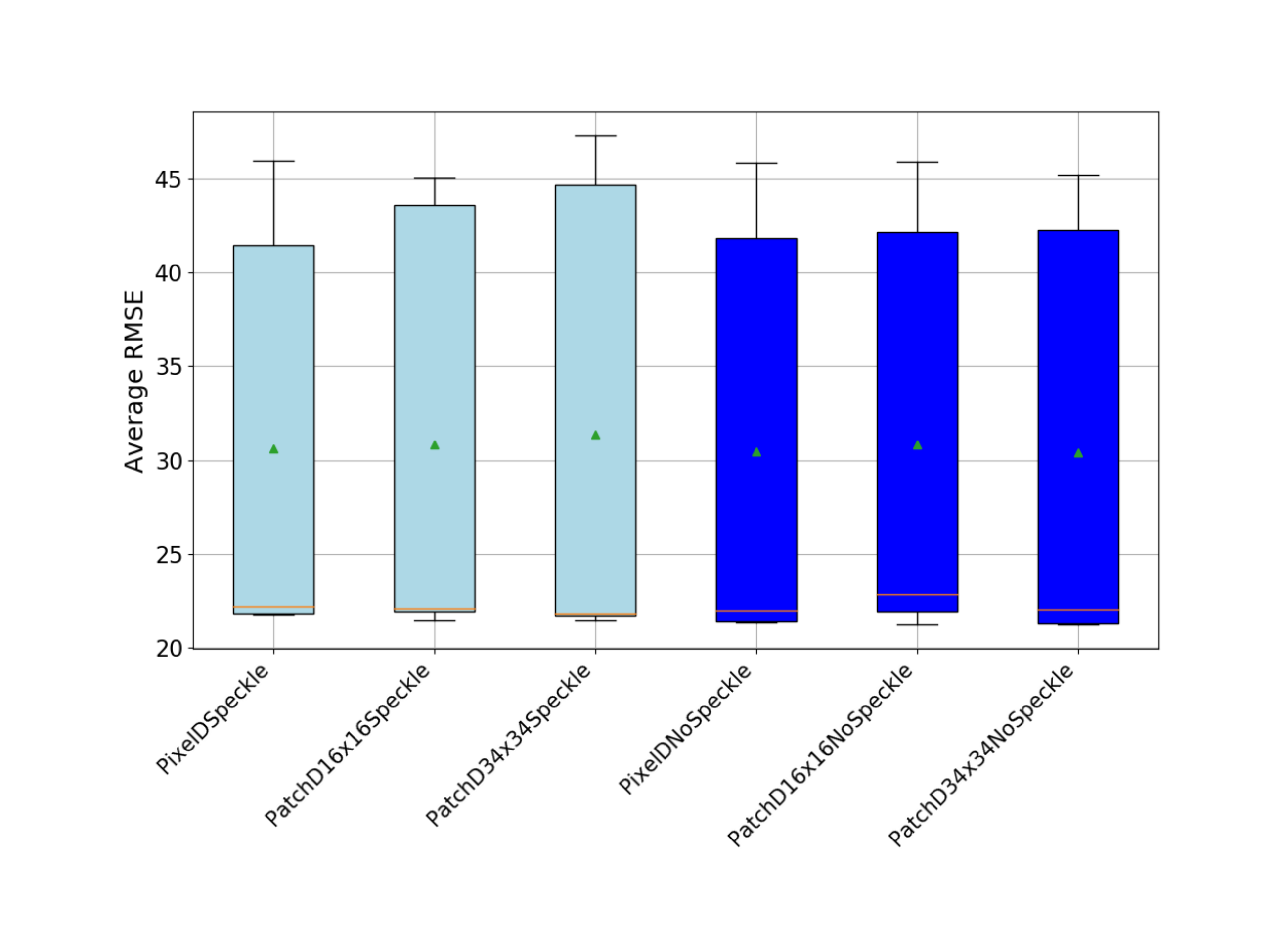}
    \caption{Boxplot comparison between models trained with different types of $\D$ on datasets produced with or without speckle filtering. Green triangles indicate the mean value computed over the five folds, while orange horizontal lines indicate the median.}
    \label{fig:exp1}
\end{figure}

\textbf{Results: }
A boxplot of average RMSE, computed between $\zhaty$ and $\zhatyx$ for the different models trained with 5-fold cross-validation (5-fold CV), is shown in \figr{fig:exp1}. Light blue bars indicate results obtained with models trained on speckle filtered data, while dark blue bars represent models trained on unfiltered data. Within a specific colour, the left, middle and right-most bar represent models trained with a \textit{PixelGAN}, a $16 \times 16$ \textit{PatchGAN}, and a $34 \times 34$ \textit{PatchGAN}, respectively. Overall, \figr{fig:exp1} shows less spread and tighter boxes for models trained on the dataset where speckle filtering was omitted. Thus, during preprocessing of the \sen\ product, speckle filtering should be skipped to achieve slightly smaller RMSE between $\zhaty$ and $\zhatyxb$. 
In general, \figr{fig:exp1} also show that the specific type of discriminator has little impact on the average RMSE for the dataset without speckle filtering. As the \textit{PixelGAN} discriminator produces slightly less spread than the two other discriminator networks, we applied it to all remaining experiments in this work and  choose to omit speckle filtering in the processing of the \sen\ product.

\begin{figure*}[htb]
    \centering
  \includegraphics[width=0.6\textwidth,height=5.8cm, angle=270]{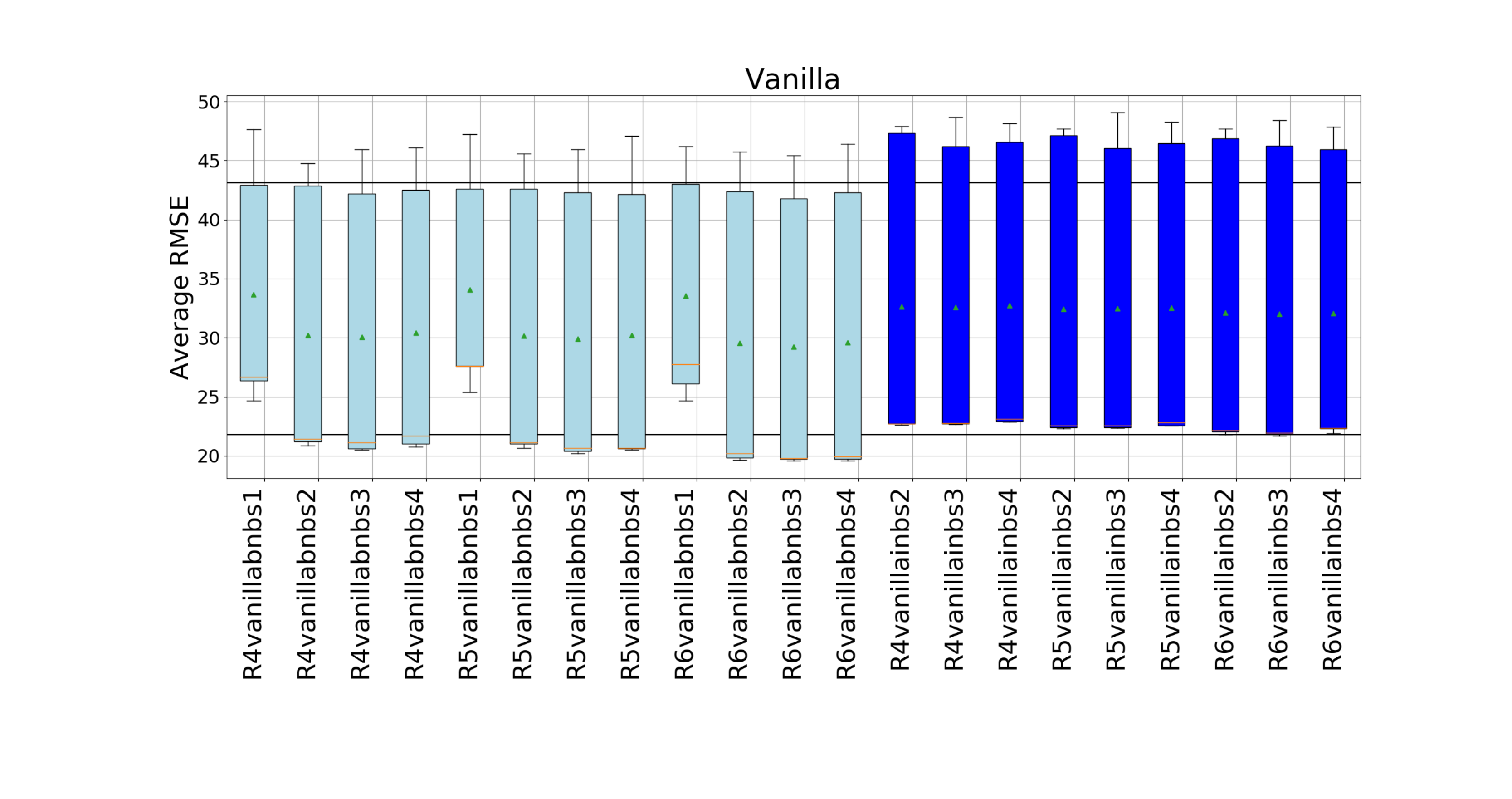}
\hspace{-1.2cm}
  \includegraphics[width=0.6\textwidth, height=5.8cm, angle=270]{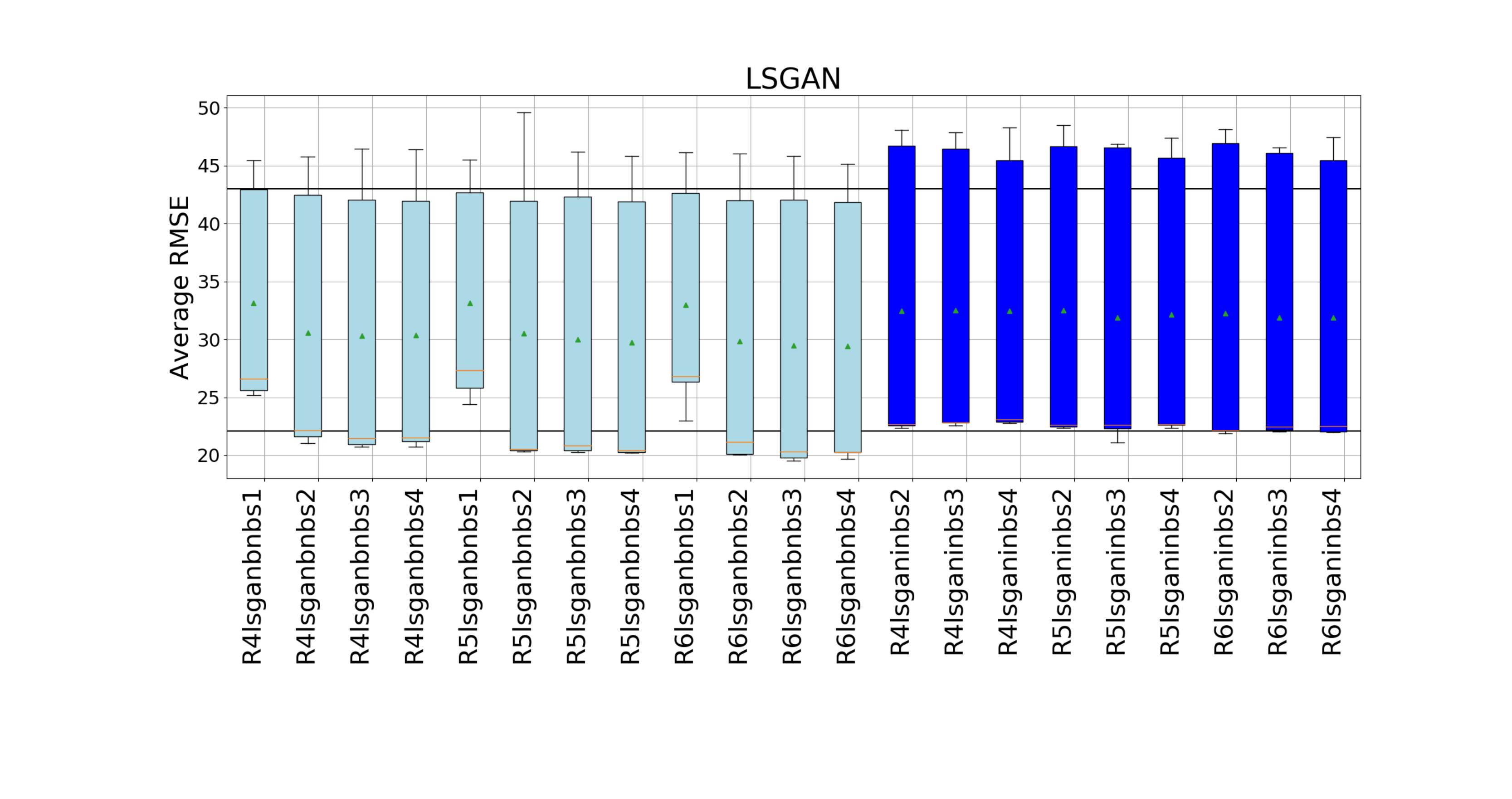}
\hspace{-1.2cm}
  \includegraphics[width=0.6\textwidth, height=5.8cm,angle=270]{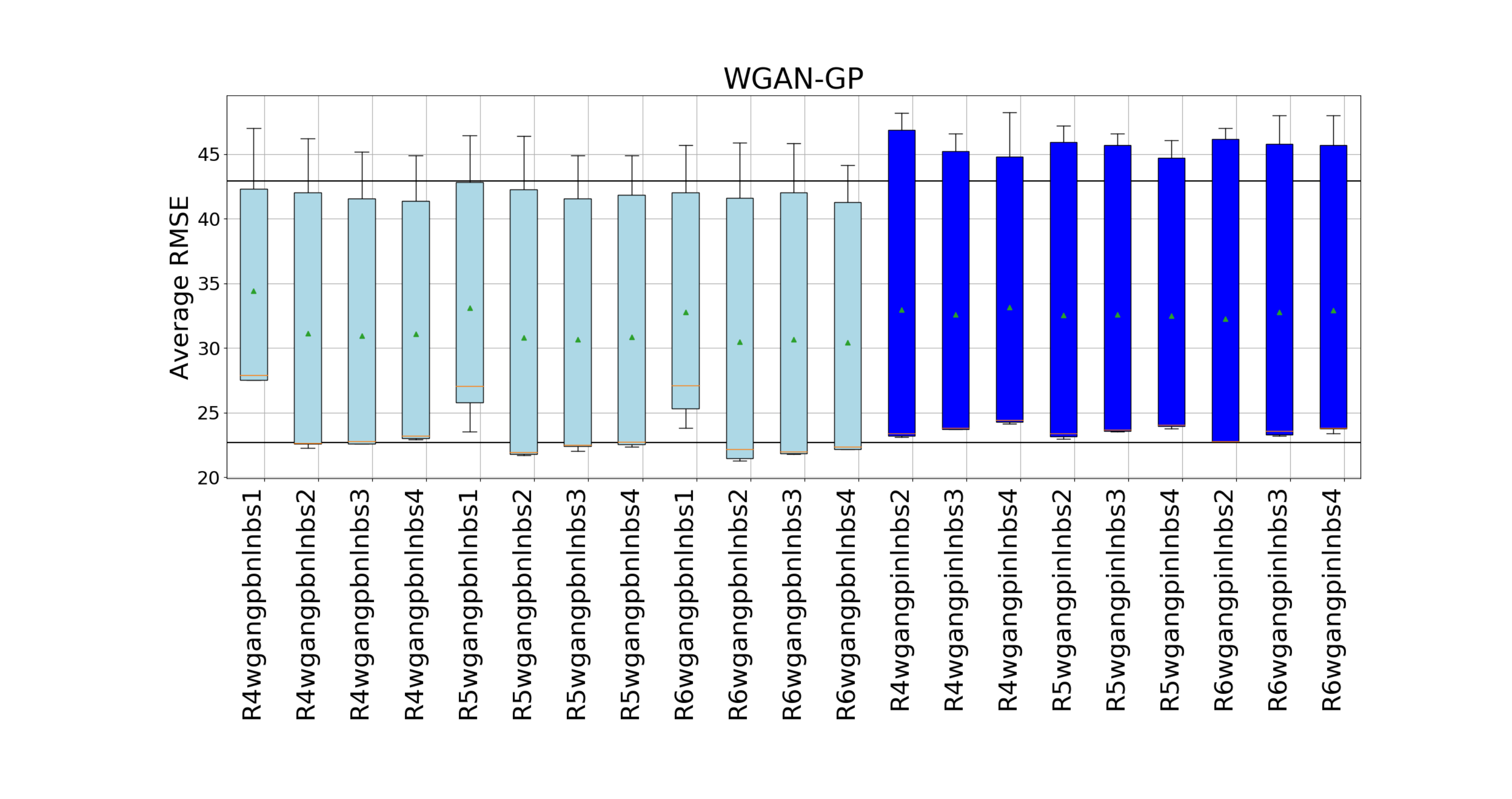}
\caption{Impact of normalisation method, BN or IN, on model performance: Light blue bins and dark blue bins represent models trained with BS and IN, respectively. Green triangles are mean values computed over the five folds, while orange vertical lines are medians. The two vertical black lines in each column are arbitrary reference lines to easen visual comparison. \textbf{Columns:} Models trained with Vanilla GAN (left), LSGAN (middle), and WGAN-GP (right).}
\label{fig:exp2_inworse}
\end{figure*}

\begin{figure*}[htb]
    \centering
     \hspace{-1.1cm}
    \includegraphics[width=0.67\textwidth,height=5.87cm, angle=270]{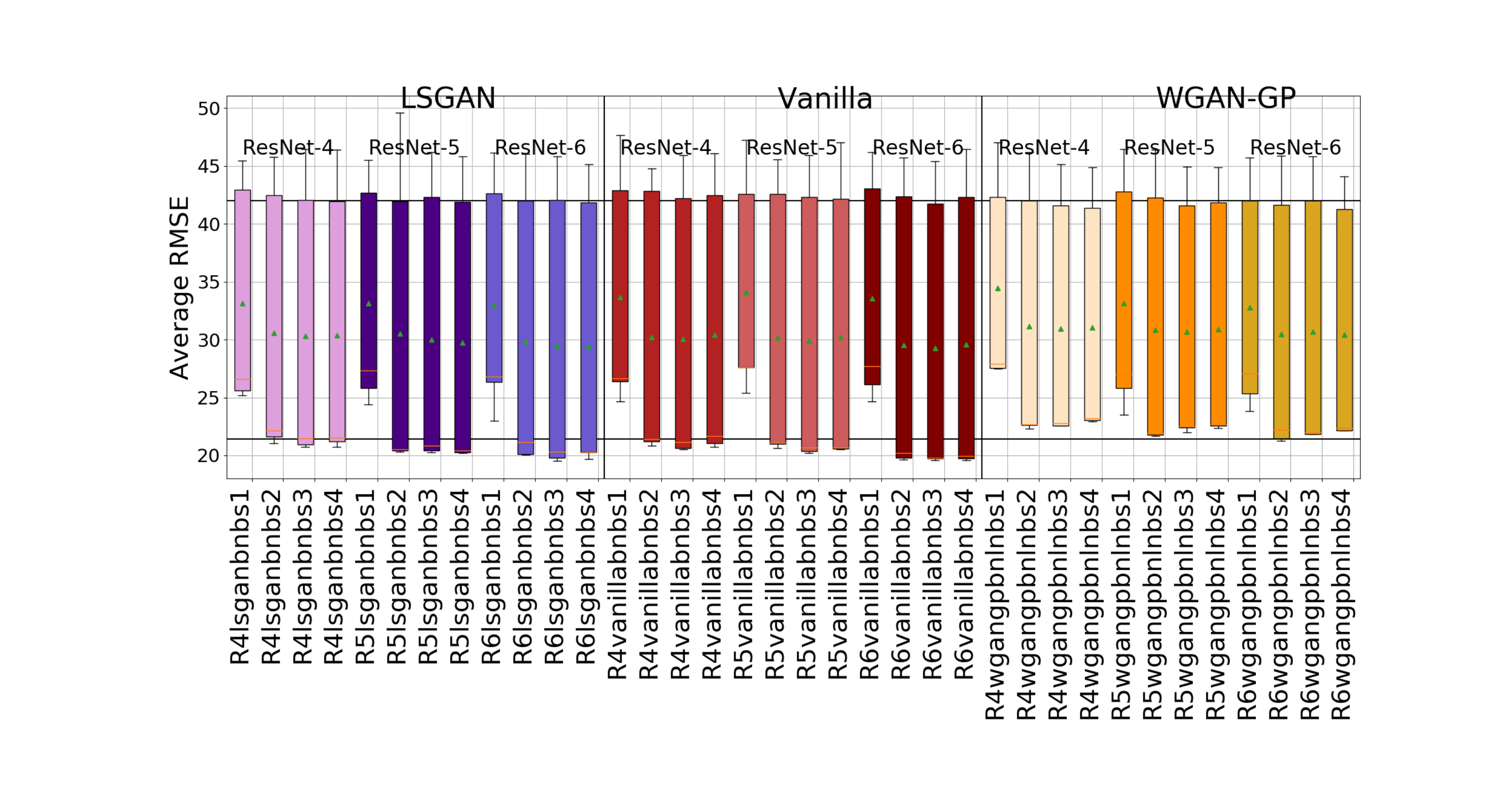}
    \hspace{-1.3cm}
    \includegraphics[width=0.67\textwidth,height=5.87cm,angle=270]{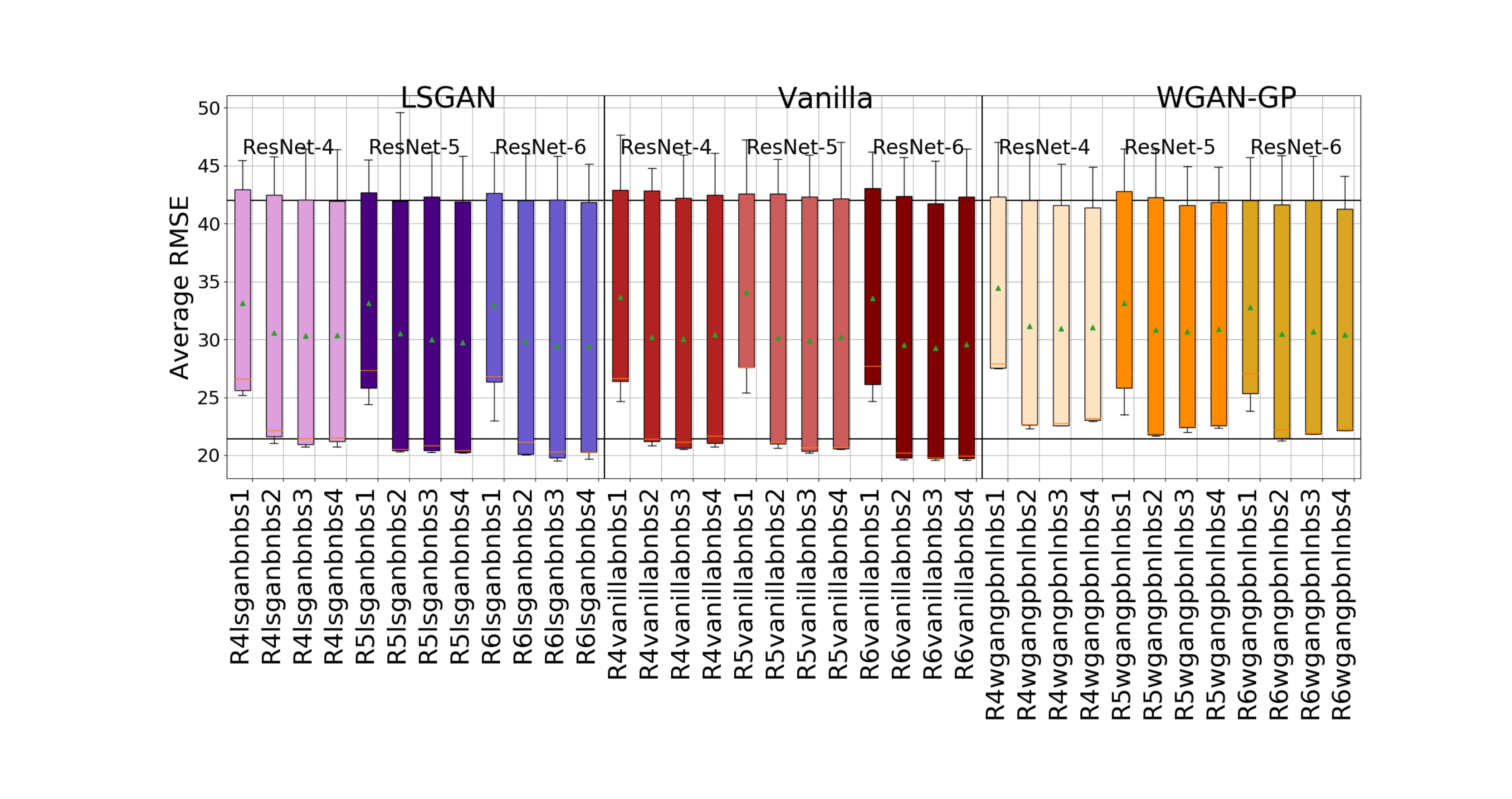}
    \hspace{-1.3cm}
    \includegraphics[width=0.67\textwidth,height=5.87cm,angle=270]{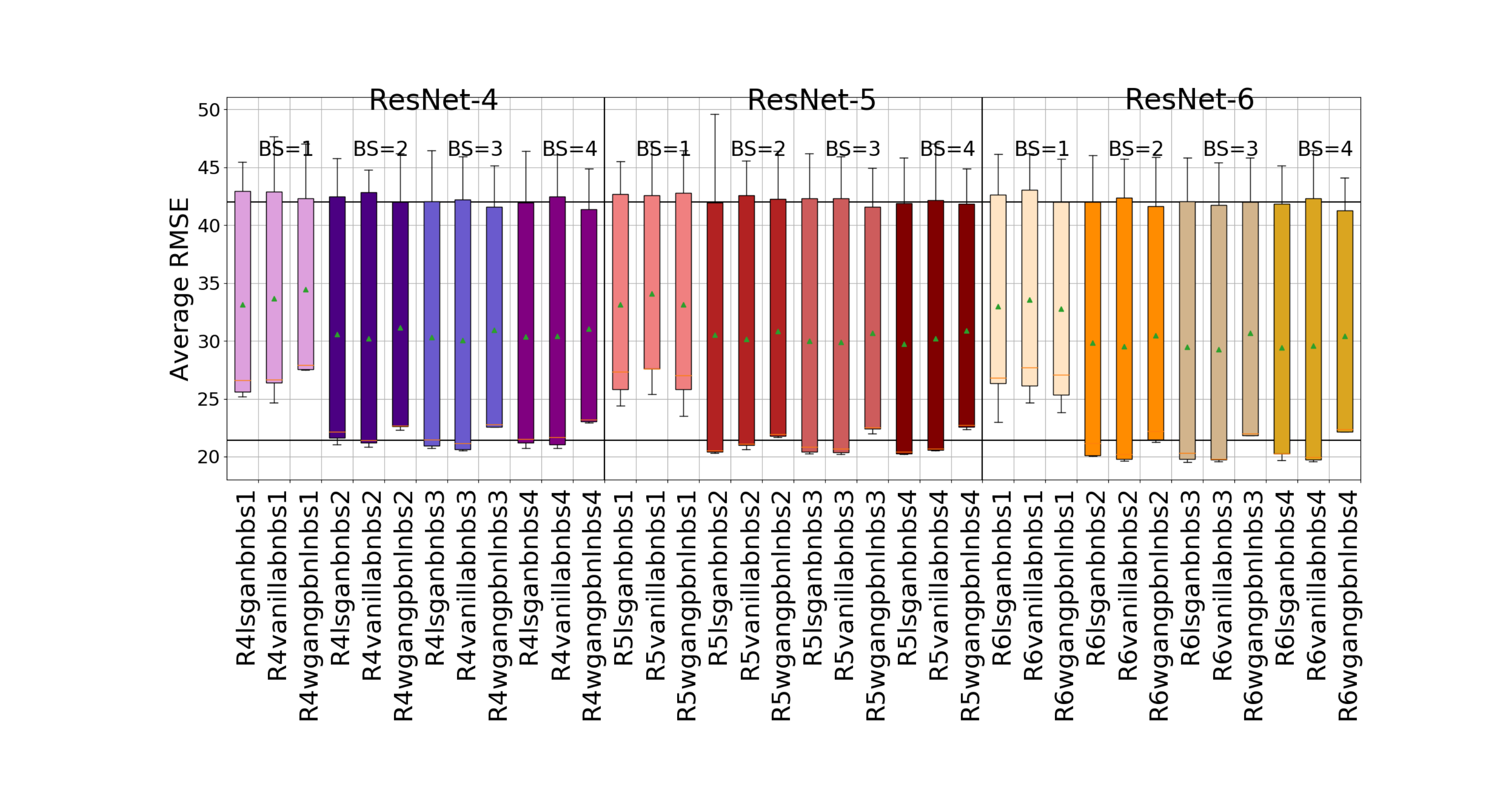}
    \caption{Boxplot of average RMSE for models trained with all three objective functions, ResNet-4, 5 or 6, with BS varying between 1-4 and BN only. Green triangles indicate the mean value computed over the five folds, while orange vertical lines indicate the median. The two vertical black lines are arbitrary chosen reference lines to easen visual comparison. \textbf{Left column:} Grouped by BS in ascending order (top to bottom). \textbf{Middle column:} Grouped by ResNet in ascending order (top to bottom) in addition to BS in ascending order (top to bottom). \textbf{Right row:} Grouped by objective functions together with similar hyperparameters, sorted by ResNet and BS in ascending order (top to bottom).}
    \label{fig:exp2_sortedAll3}
\end{figure*}

\subsection{Experiment 2: A comparison of model architectures, normalisation methods and objective functions}\label{exp:exp2}

Here, we investigated if any combination of model architecture, normalisation method and cGAN objective function improves the accuracy of $\zhatyx$ with respect to $\zhaty$. Based on the results in \Sec{exp:data} we kept the dataset fixed, i.e.\ we used the \sen product processed without speckle filtering and applied the $1\times 1$ \textit{PixelGAN} discriminator for all models trained in this section. Nine different cGAN generator architectures $\G$ were trained by combining the three ResNet networks and the three objective functions from Sec.~\ref{sec:method}. We also applied BN or instance normalisation (IN) for Vanilla GAN  and LSGAN, while for WGAN-GP we applied LN for $\D$ and either BN or IN the $\G$ network, as suggested in \cite{wgangp}. We additionally experimented with a BS between 1 and 4. For each model, we applied 5-fold CV, and trained it for 200 epochs. We evaluate the different models on the 5-fold CV test sets by visualising boxplots of average RMSE computed between $\zhaty$ and $\zhatyx$.

\textbf{Results: }\figr{fig:exp2_inworse} visualises models trained on the three different objective functions in separate columns, i.e.\ Vanilla GAN in the left column, LSGAN in the middle column and WGAN-GP in the right column. We show models trained with BN in light blue colour, while models trained with IN are shown in dark blue colour. For all three objective functions, models trained with BN achieve a smaller average RMSE. Additionally, \figr{fig:exp2_inworse} shows that most models trained with BN also experience a smaller spread in average RMSE over the 5-fold CV dataset. Thus, we conclude from \figr{fig:exp2_inworse} that applying BN is preferable to produce $\zhatyx$ predictions with smaller average RMSE. 

In \figr{fig:exp2_sortedAll3} we compare models trained with different ResNet architectures and BS values to each other. In the left column, the models are first sorted by objective function, then by ascending BS, and finally by ascending ResNet model order. The grouping by BS is indicated with colours. In the middle column, models are again first sorted by objective function, but then by ascending ResNet model order (colour coded groups), and finally by ascending BS. In the right column, models are first sorted by ascending ResNet model order, then by ascending BS (colour coded groups), and lastly by objective function. Overall, \figr{fig:exp2_sortedAll3} shows that the choice of objective function has little influence on the average RMSE, as the group of bins for the different objective functions look very similar to each other. Neither does the choice of ResNet model order have a significant impact on the average RMSE, although the positions of the green triangles in the left column of \figr{fig:exp2_sortedAll3} indicate that ResNet-6 has a slightly smaller mean value than ResNet-5 and ResNet-4. What influences the average RMSE the most, is the choice of BS. All columns show that BS = 1 yields a smaller spread of average RMSE, but also a higher mean value. Models trained on BS = 2, 3 or 4 achieve a similar spread of average RMSE for all three objective functions, although the WGAN-GP shows slightly less spread.  
To summarise, the choice of normalisation method and batch size has the largest impact on the RMSE, compared to other hyperparameters and the objective functions for the $\G$ or $\D$ networks. We recommend that BN should be chosen instead of IN, and that BS = 1 should be avoided.

\subsection{Image patch generation}\label{patchgen}
In this experiment, we evaluated the correspondence between generated image patches $\zhatyxb$ to $\zhatyb$. \tabref{tab:optcgan} lists the implementational choices for each cGAN variant used in this experiment, these are based on the validation described above, see \Sec{exp:data} and \Sec{exp:exp2}. Each cGAN variant were trained on a training set, while the test set were kept aside. After training, we allowed the trained $\G$ network of each model to generate $\zhatyxb$ image patches from \sen\ image patches. These \sen\ image patches were from the test set, and had therefore not been seen by the network during training. Since the test set also contain the corresponding target , i.e. $\zhatyb$ image patches, these were used to evaluate the generator's performance quantitatively and qualitatively.

\begin{figure*}[!htb]
    \centering
\includegraphics[scale=0.5]{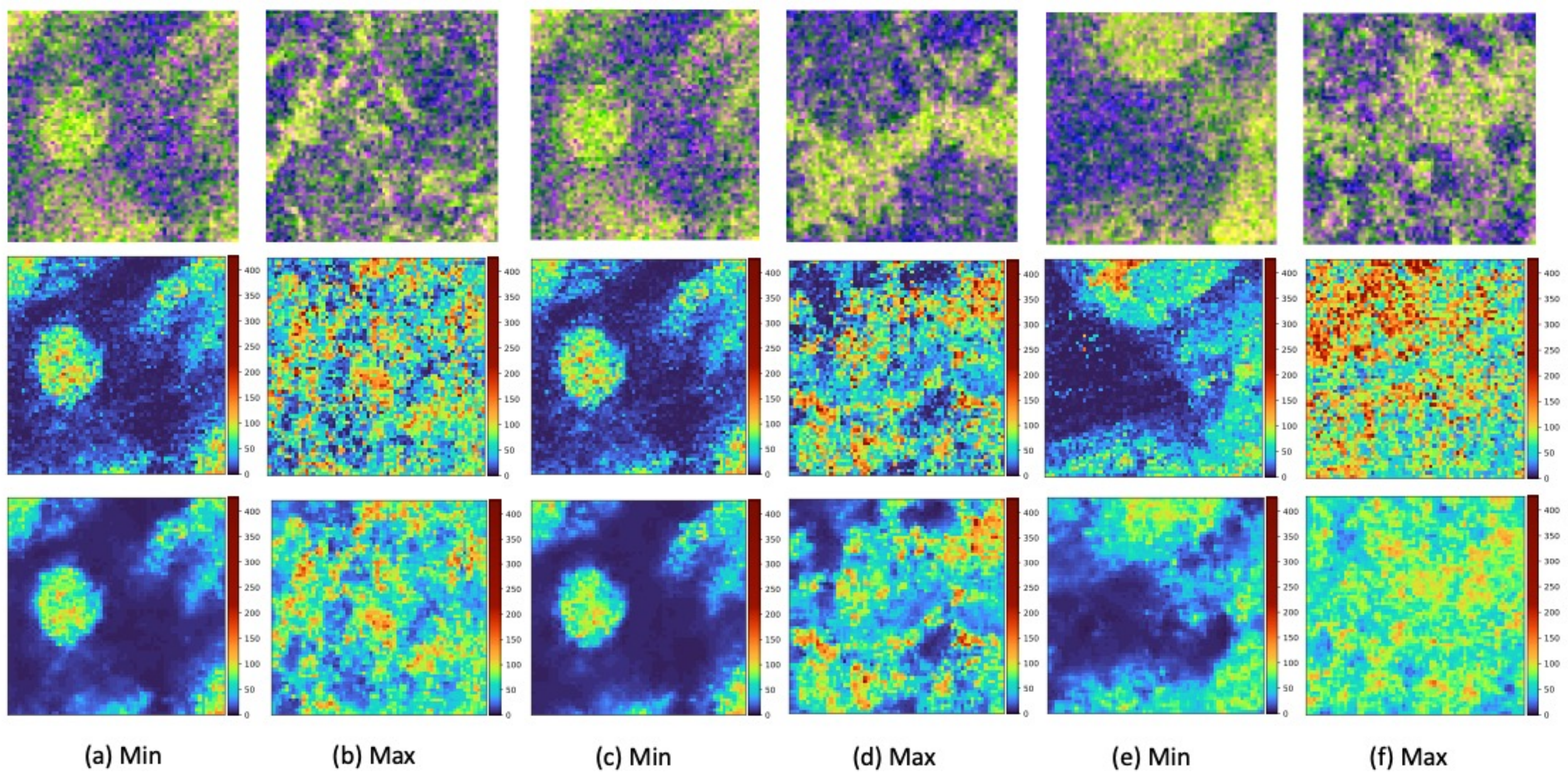}
\caption{\textbf{First row:} \sen\ patches. \textbf{Second row:} Target image patches, i.e.\ ALS-based AGB predictions $\zhatyb$. \textbf{Third row:} Generated synthetic image patches, i.e.\ $\zhatyxb$. Column \textbf{(a)} and \textbf{(b)}: Vanilla GAN; \textbf{(c)} and \textbf{(d)}: LSGAN; \textbf{(e)} and \textbf{(f)}: WGAN-GP. Columns with caption \textit{Min} and \textit{Max} respectively refer to an image patch within the test set that achieves minimum and maximum RMSE, computed over all $64\times 64$ pixels in the test patch ($\mathrm{Mg\,ha}^{-1}$).}
  \label{fig:minmax}
\end{figure*}

\textbf{Results: }For each of the models in \tabref{tab:optcgan}, we select test patches, i.e.\ $\zhatyxb$ and corrsponding $\zhatyb$, having the smallest and greatest RMSE ($\mathrm{Mg\,ha}^{-1}$) to investigate the worst and best case scenarios. The RMSE is computed over all pixels within the image test patch. \figr{fig:minmax} shows a qualitative comparison of the identified test patch with the smallest and greatest RMSE for the three models. The first row of \figr{fig:minmax} visualises patches from the input domain i.e.\ \sen, the middle row from the target domain, i.e.\ $\zhatyb$, and the third row from the generated domain, i.e.\ $\zhatyxb$. Columns with caption \textit{Min} indicate an image patch with the smallest RMSE for a specific model, while caption \textit{Max} instead indicates an image patch with the largest RMSE. Columns (a) and (b) correspond to patches from the optimal Vanilla GAN, (c) and (d) from the optimal LSGAN, while (e) and (f) are from the optimal WGAN-GP. Quantitative comparisons of RMSE for the patches in \figr{fig:minmax} are shown in \tabref{tab:minmax}. 

As the same patch was identified as easiest to translate by both the Vanilla GAN and the LSGAN model, these two cGAN variant must have learned similar translation dynamics between the input and output domains. See columns (a) and (c) of \figr{fig:minmax}.
The results provided in \Sec{sec:seq_eval} also points to the same direction; overall the Vanilla GAN and the LSGAN models perform more similar to each other and achieves higher accuracy than the WGAN-GP model. \tabref{tab:minmax} clearly shows that the WGAN-GP model is the worse among the three, having an RMSE which is almost twice as high as for the Vanilla GAN or LSGAN. From \figr{fig:minmax}, it can be noted that all three objective functions seem to be approximately equally appropriate for translating from $\xcal$ to $\zhatyxb \in \zcal$ when patches from the two domains have similar appearance, but struggle when the $\xcal$ and $\zhatyxb \in \zcal$ domains deviate from each other in appearance. Visually, all three objective functions generate synthetic patches which are somewhat more blurry than $\zhatyb$ predictions. Blurriness is a known weakness with generative models, such as GANs \cite{Khayatkhoei2020SpatialFreq, Durall2020Watch}. Several possible explanations to it exists, for example that blurriness can be related to the transposed convolution up-sampling method used in the second part of the $\G$ network. These up-sampling methods affect the model's ability to correctly reproduce the spectral distribution in images, or to generate new images with sharp high frequent components such as edges \cite{Durall2020Watch}. 

\begin{table}[!htb]
    \caption{List of minimum and maximum RMSE for the test image patches shown in \figr{fig:minmax}. The listed models are from \Sec{patchgen} and only differ from each other by the objective functions.}
    \label{tab:minmax}
    \centering
  \begin{tabular}{|c|c|c|}
    \hline
 Model & Min [$\mathrm{Mg\,ha}^{-1}$] &  Max [$\mathrm{Mg\,ha}^{-1}$]\\
    \hline
    Vanilla GAN; ResNet-6 BN, BS=3 &  11.03 & 27.04\\
    LSGAN; ResNet-6 BN, BS=3 &  10.92 &  27.34\\
   WGAN-GP; ResNet-6 BN, BS=3 & 22.23 &  57.27\\
    \hline
\end{tabular}
\end{table}

\subsection{Comparison of linear or dB scale SAR input}\label{sec:lindb}
In the \sen\ processing workflow we settled for, see \Sec{sec:preprosar}, conversion to dB scale was only applied if the \sen\ scene was used by the cGAN-based sequential models. The use of dB scale on the \sen\ data for these models was decided by the results of the experiments provided in this section.
We evaluated the impact of keeping the \sen\ input data on linear scale versus to transform it to a logarithmic decibel (dB) scale. This was done by creating two versions of the \sen\ dataset, where conversion to dB was applied to one of these. Except for this step, both \sen\ datasets, referred to as \sen\ linear or \sen\ dB, were identically processed. For each of the optimal model implementations listed in \tabref{tab:optcgan}, we trained one model on the \sen\ linear dataset and another on the \sen\ dB dataset. This yielded six different possibilities to generate $\zhatyx$, i.e.\ three different linear cGAN-based models and three different dB cGAN-based models. From each of these six models, we extracted $\zhatyxb$ predictions corresponding to the position of each AGB ground reference measurement $z$. 

\begin{figure*}[!htb]
  \begin{picture}(0,250)
 \renewcommand\thefigure{\alph{figure}}
 \renewcommand\caption[1]{\refstepcounter{figure}
                          \par\centering(\thefigure)\par}
  \put(30,60){\begin{minipage}[b]{.22\textwidth}
   \subfloat[\label{s1aseq_scatter_linear}]{\includegraphics[scale=0.30]{images/scatter/scatter_s1a_sequential_rmse_p.pdf}}
    \end{minipage}}
  \put(145,120){\begin{minipage}[b]{.22\textwidth}
   \subfloat[\label{vanilla_scatter_linear}]{\includegraphics[scale=0.3]{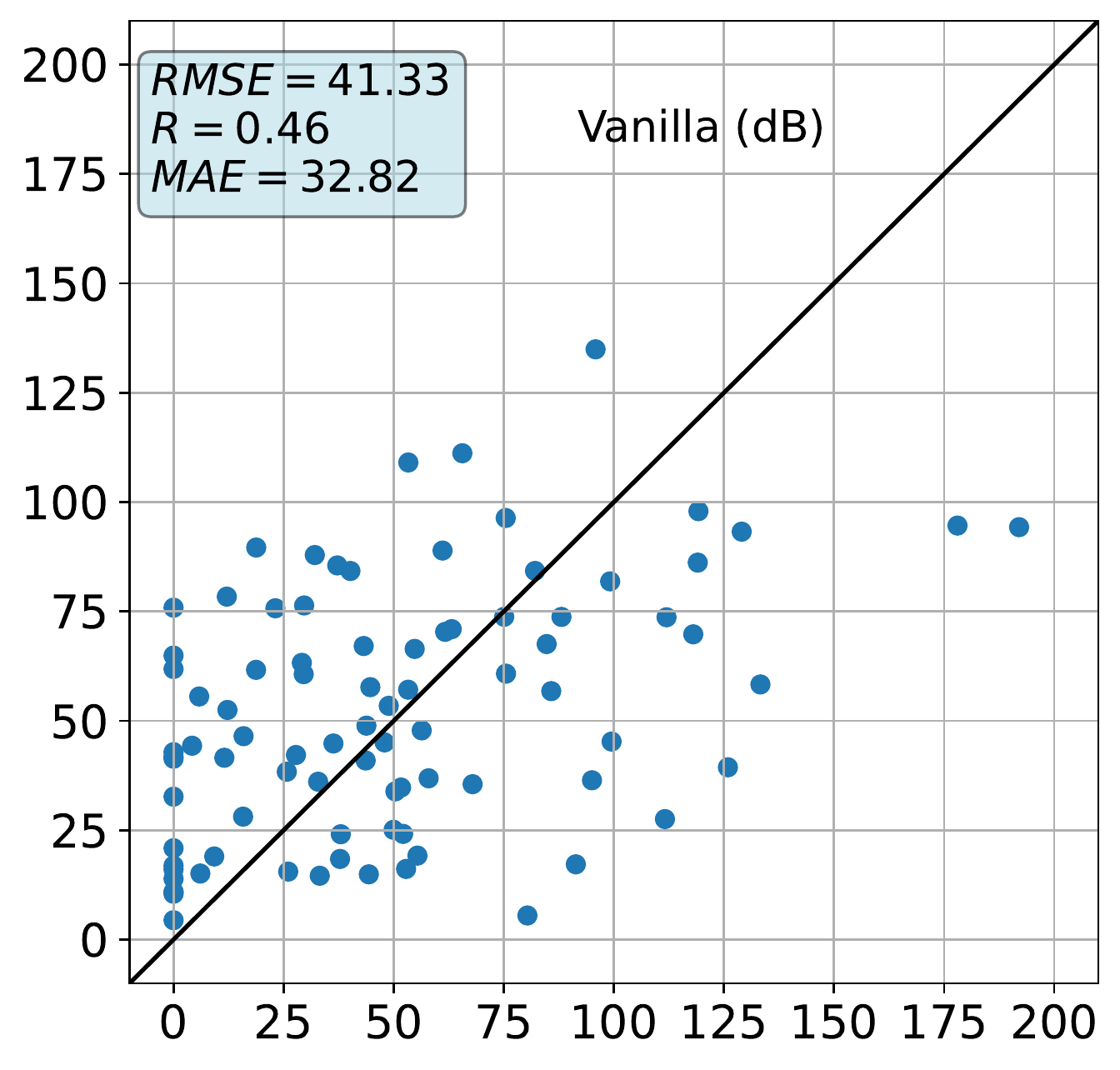}}
    \end{minipage}}
    \put(260,120){\begin{minipage}[b]{.22\textwidth}
   \subfloat[\label{lsgan_scatter_linear}]{\includegraphics[scale=0.3]{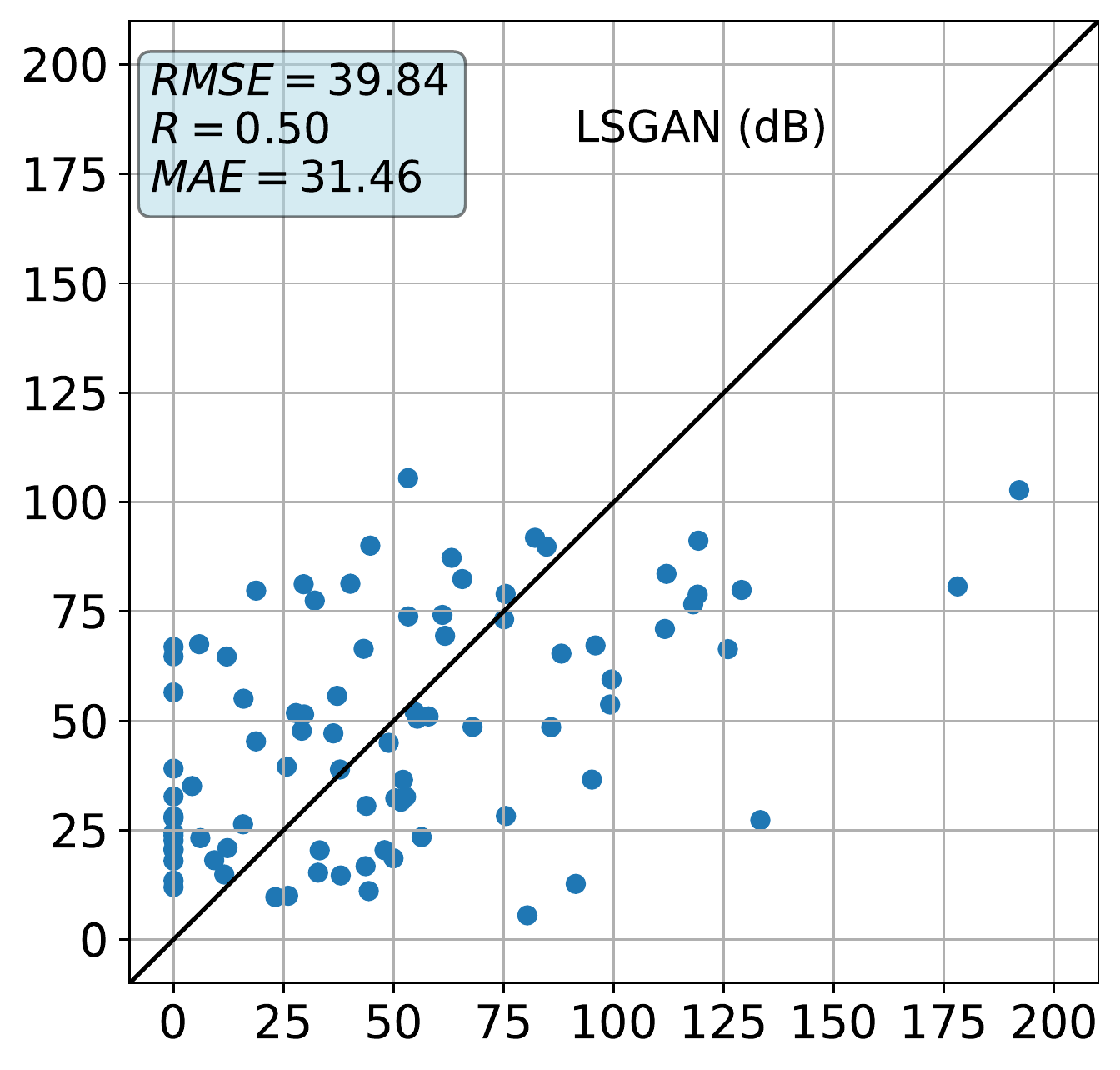}}
    \end{minipage}}
      \put(375,120){\begin{minipage}[b]{.22\textwidth}
   \subfloat[\label{wgangp_scatter_linear}]{\includegraphics[scale=0.3]{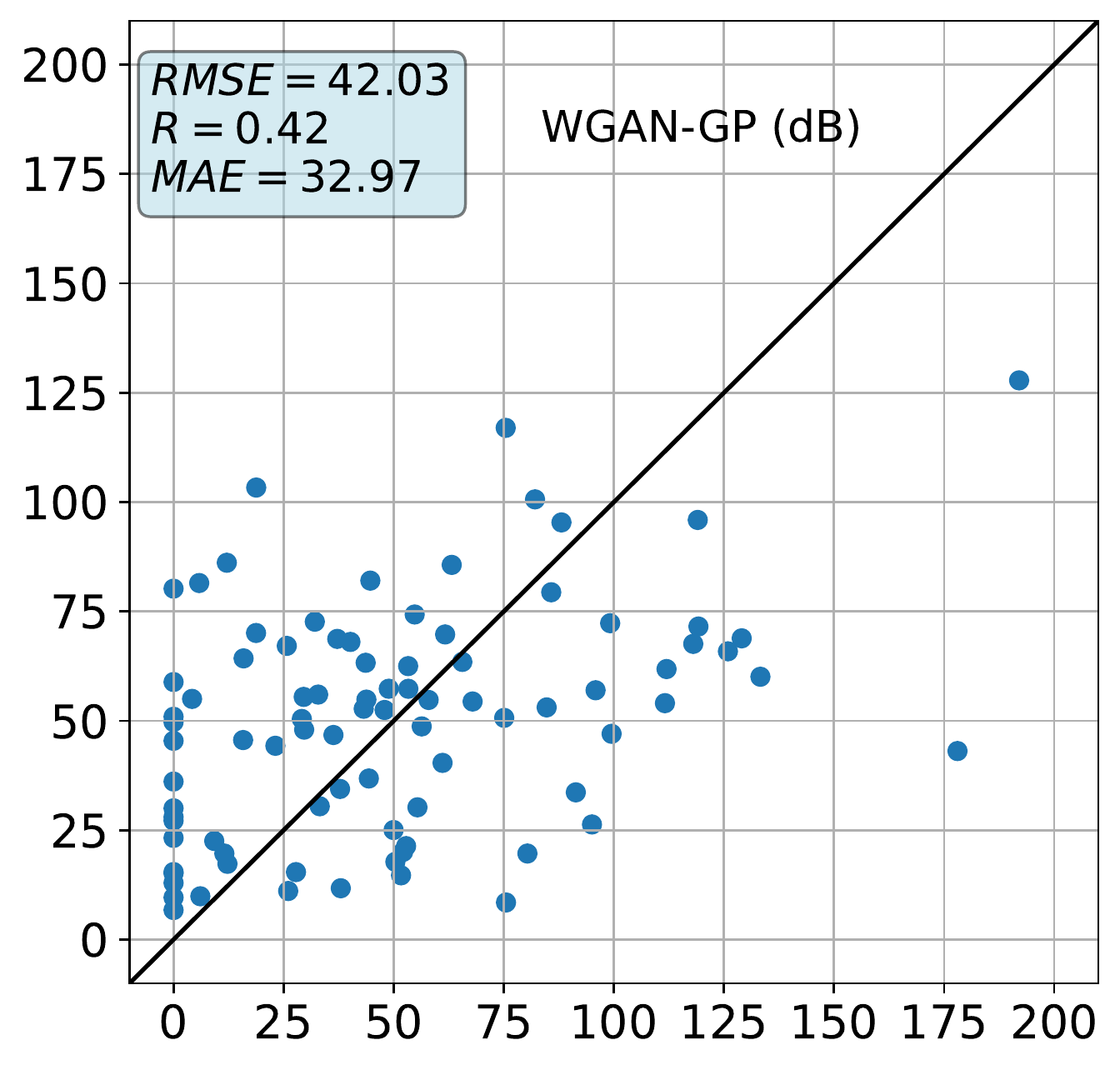}}
    \end{minipage}}
  \put(145,-8){\begin{minipage}[b]{.22\textwidth}
   \subfloat[\label{vanilla_scatter_db}]{\includegraphics[scale=0.3]{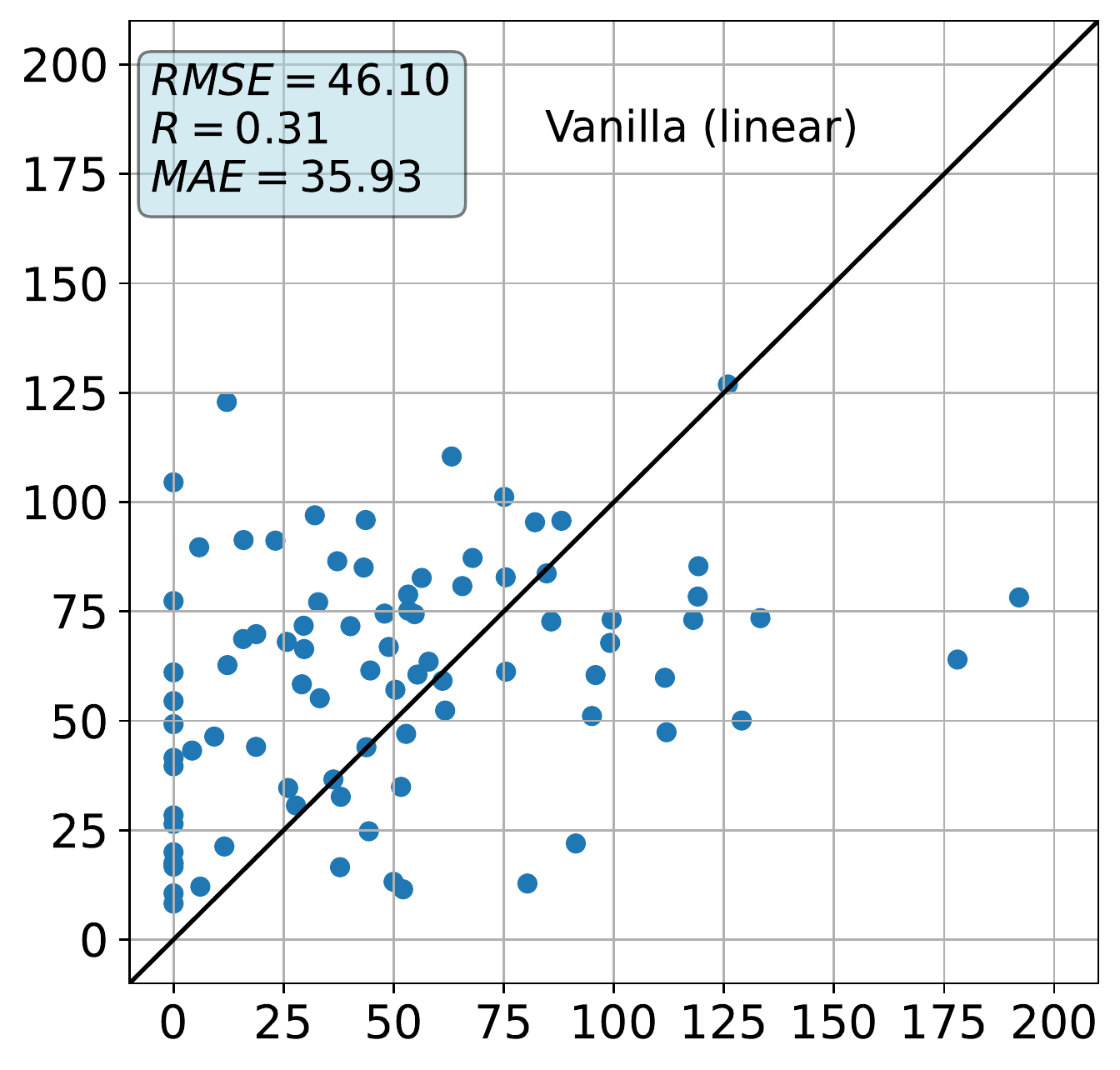}}
    \end{minipage}}
    \put(260,-8){\begin{minipage}[b]{.22\textwidth}
   \subfloat[\label{lsgan_scatterdb}]{\includegraphics[scale=0.3]{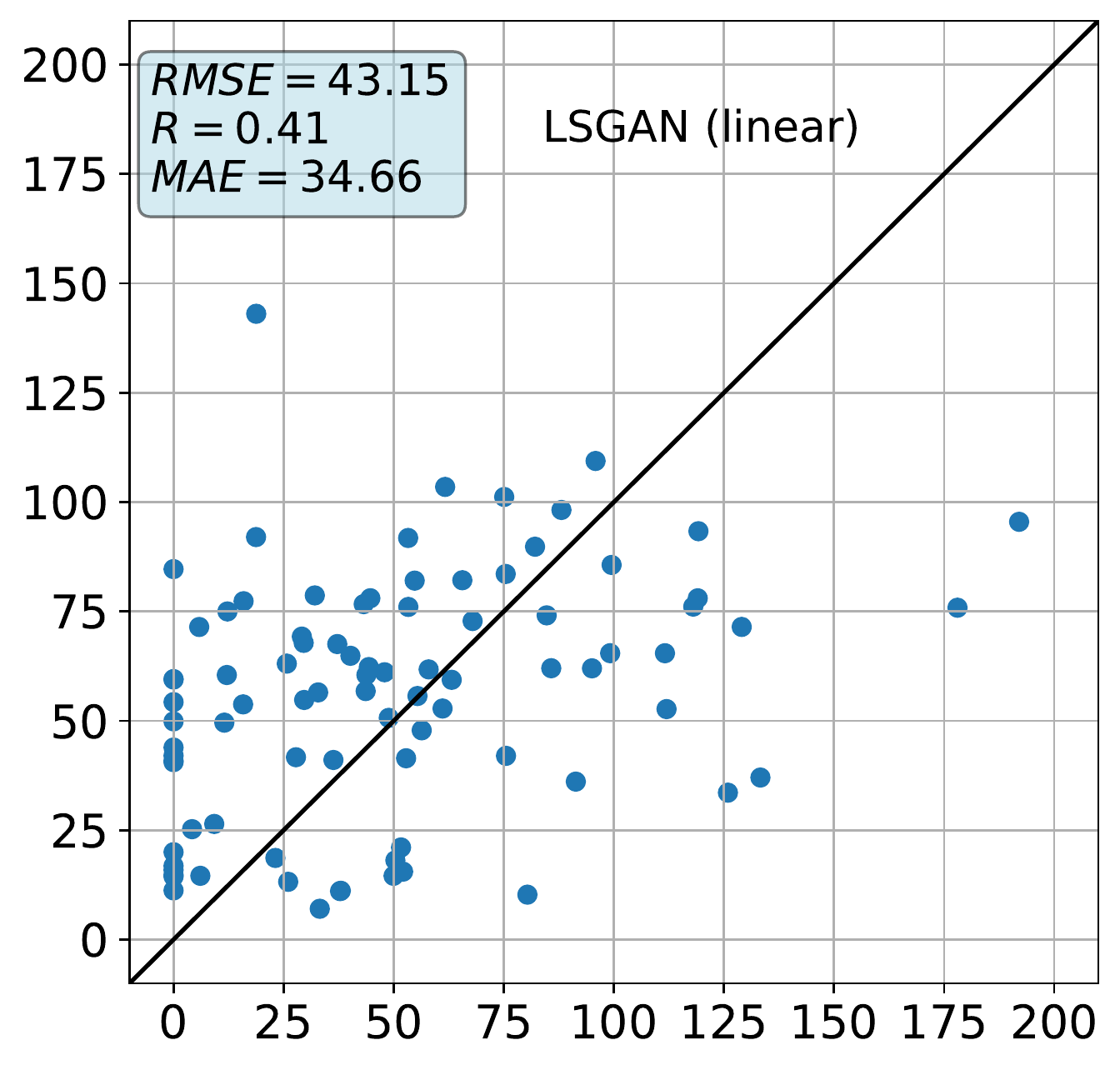}}
    \end{minipage}}
      \put(375,-8){\begin{minipage}[b]{.22\textwidth}
   \subfloat[\label{wgangp_scatter_db}]{\includegraphics[scale=0.3]{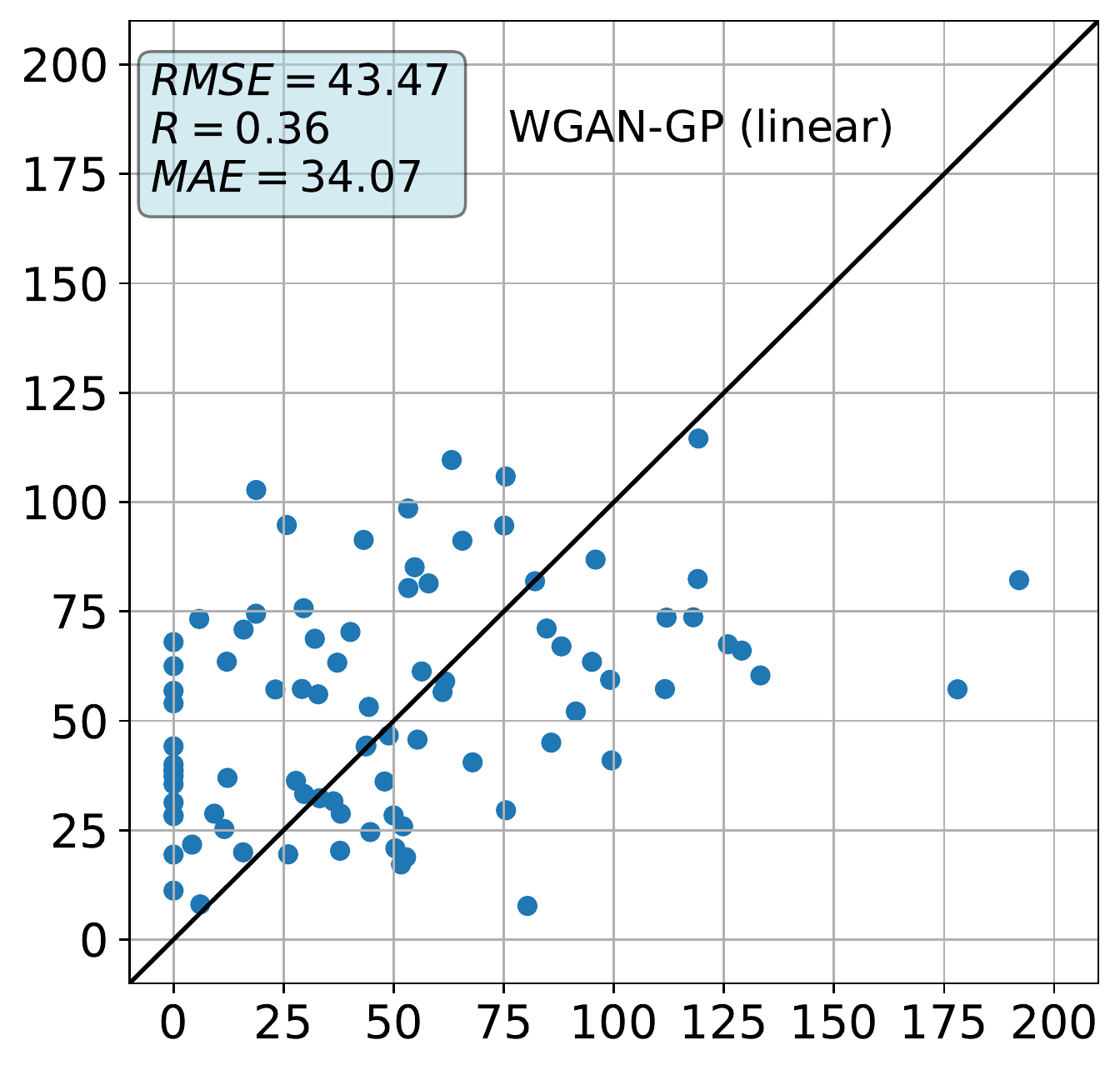}}
    \end{minipage}}
  \end{picture}
\vspace*{.2cm}
\leavevmode\smash{\makebox[0pt]{\hspace{2em}         
  \rotatebox[origin=l]{90}{\hspace{6em}
    Predicted AGB (Mg ha$^{-1}$)}
}} 
\put(240,-30){\begin{minipage}[b]{.5\textwidth}
   {Ground Reference biomass (Mg ha$^{-1}$)}
    \end{minipage}}

\caption{Scatter plots between AGB ground reference data, $z$, and model-predicted AGB. \textbf{Upper row}, models trained with \sen\ data on dB scale, i.e. the proposed Vanilla GAN \textbf{(b)}, LSGAN \textbf{(c)} and WGAN-GP \textbf{(d)} models. \textbf{Lower row}, same models as above, but trained with \sen\ on linear scale. Fig.\ \textbf{(a)} model-predicted AGB values from the baseline sequential  regression model given in \eq{eq:agb} trained with \sen\ data on linear scale. The black lines are reference lines indicating 100\% correlation between $z$ and AGB predictions. Units are in Mg ha$^{-1}$.}
    \label{fig:scatterseq_db_linear}
\end{figure*}

\textbf{Results:} We provide scatter plots of $\zhatyx$ predictions and $z$ in \figr{fig:scatterseq_db_linear}, where (b)-(d) represent results from the cGAN models trained on linear scale, while (e)-(g) represent corresponding results from the cGAN models trained on dB scale. For comparison with the baseline sequential \sen\ model, we also show a corresponding scatter plot of it in (a) (it is the same figure as in \figr{fig:scatter_seq} (a)). We also provide computed RMSE, R and MAE in each scatter plot.  
Overall, \figr{fig:scatterseq_db_linear} shows that R decreases while both RMSE and MAE increase if any of the cGAN models are trained on linear scale as compared to dB scale. We conclude that the conversion of calibrated $\sigma_0$ values to dB scale, which increases the dynamic range of the pixel values in the image, is advantageous for achieving more accurate image-to-image translation through the cGAN architecture.

\subsection{Post-calibration of sequential models}\label{sec:cal}
Although the non-sequential \sen\ model cannot predict AGB between 0-20.3 $\mathrm{Mg\,ha}^{-1}$, it still achieves a higher correlation coefficient R and a lower RMSE/MAE with respect to $z$ than any of the proposed sequential models. One explanation can be that the non-sequential model had access to the ground reference data $z$ during model fitting. By contrast, the sequential models were only using $\zhaty$ during model fitting and have therefore not been calibrated against $z$. In this experiment, we investigated if the accuracy of the sequential regression models could improve if we after constructing the synthetic AGB prediction maps calibrated them against $z$. As the original LSGAN model achieved the highest correlation with $z$, we focus the experiments in this section on this model and the baseline sequential \sen\ model. Furthermore, for the LSGAN we considered both \sen\ data on linear scale and dB scale. Overall, we investigated five common calibration methods, i.e.\ \textit{linear, exponential, gamma, nth-root} and \textit{logarithmic} calibration. Among these, we choose to show gamma and linear calibration results, as we obtained the best results with these methods.

\begin{figure*}[htb]
\centering
\subfloat[\label{als_scatter_ref}]{\includegraphics[width=0.25\textwidth]{images/scatter/scatter_als_rmse_p.pdf}}
\subfloat[\label{s1a_scatter_ref}]{\includegraphics[width=0.25\textwidth]{images/scatter/scatter_s1a_rmse_p.pdf}}
\\
\hspace*{1.cm}
 \subfloat[\label{lsgan_orig_scatter}]{\includegraphics[width=0.25\textwidth]{images/appendix/scatter/db_linear/scatter_lsgan_db_spat_rmse_p.pdf}}
 \subfloat[\label{lsgan_gamma_scatter}]{\includegraphics[width=0.25\textwidth]{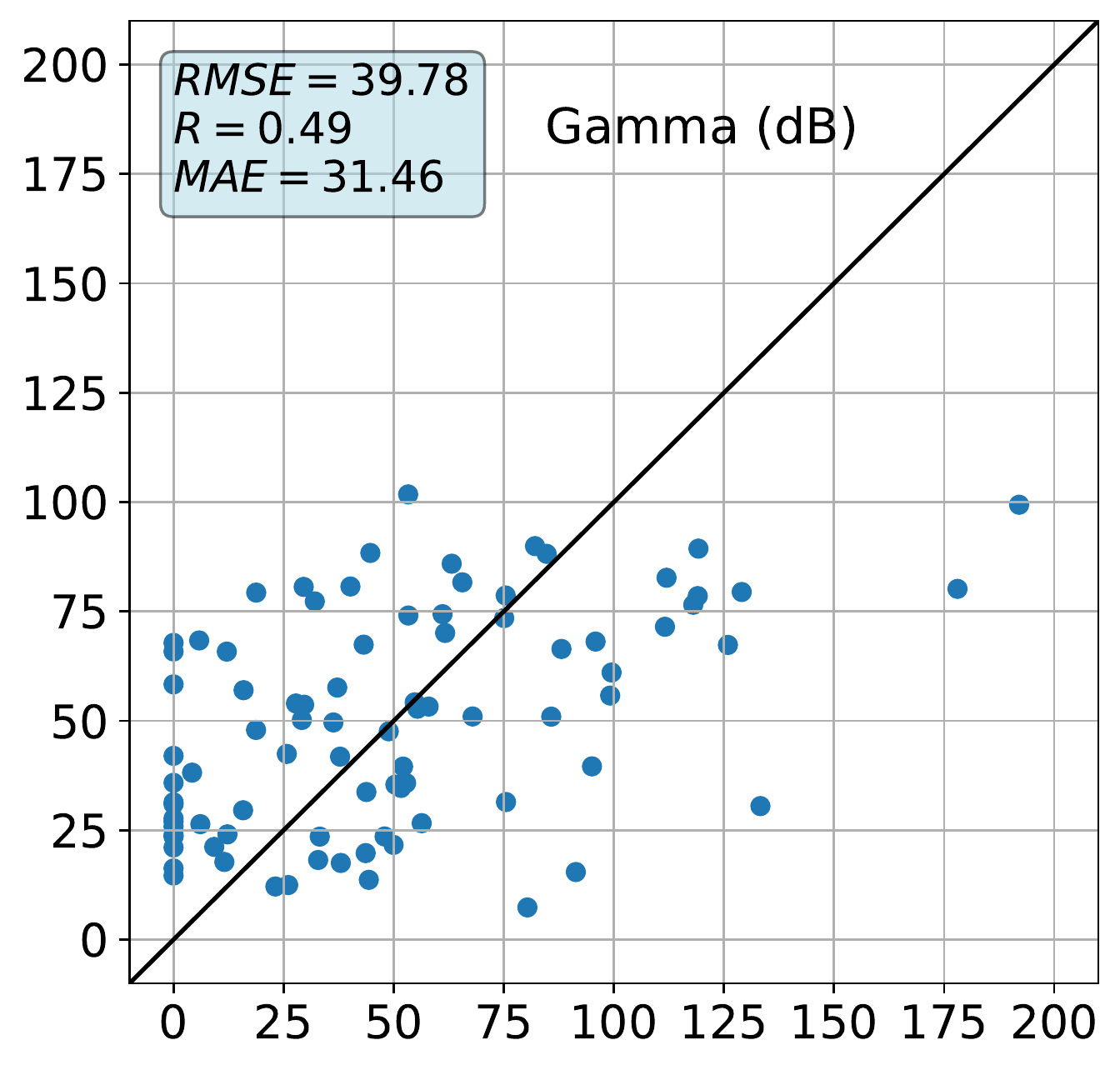}}
\subfloat[\label{lsgan_linear_scatter}]{\includegraphics[width=0.25\textwidth]{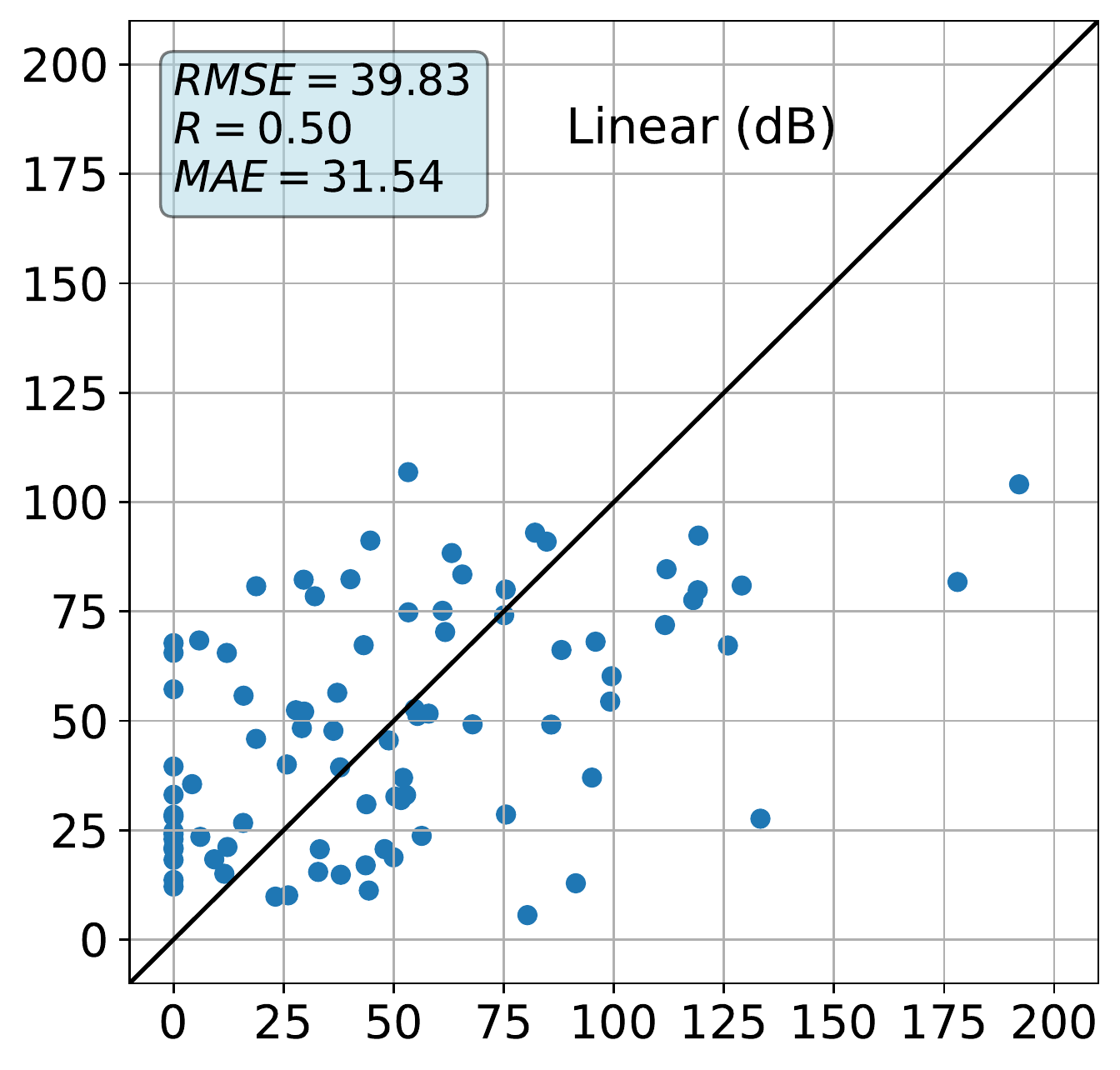}}
\vspace*{.2cm}
\\
\hspace*{1.cm}
 \subfloat[\label{lsgan_linspat_scatter}]{\includegraphics[width=0.25\textwidth]{images/appendix/scatter/db_linear/scatter_lsgan_lin_spat_rmse_p.pdf}}
 \subfloat[\label{lsgan_gamma_linear_scatter}]{\includegraphics[width=0.25\textwidth]{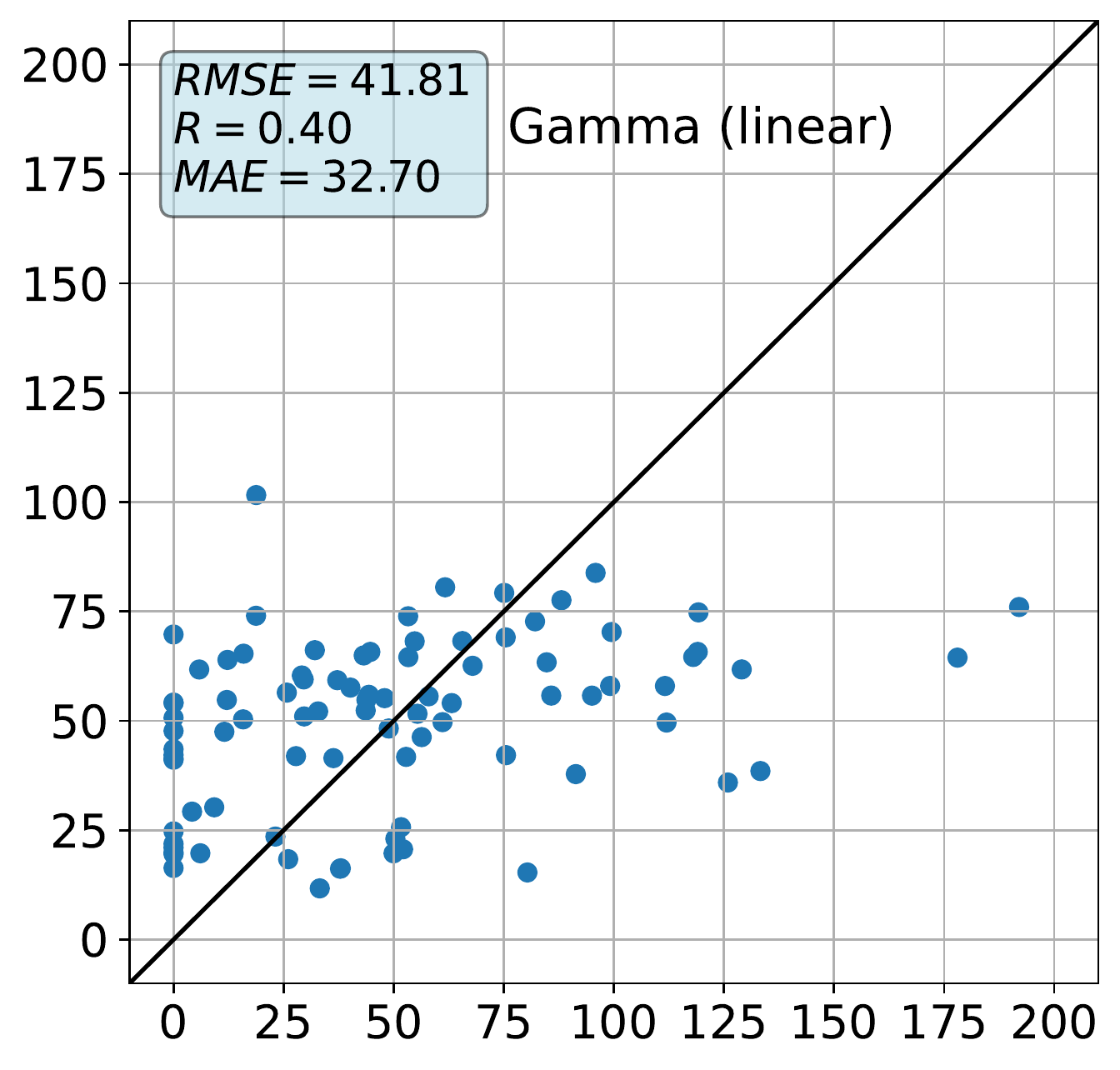}}
\subfloat[\label{lsgan_linear_Linear_scatter}]{\includegraphics[width=0.25\textwidth]{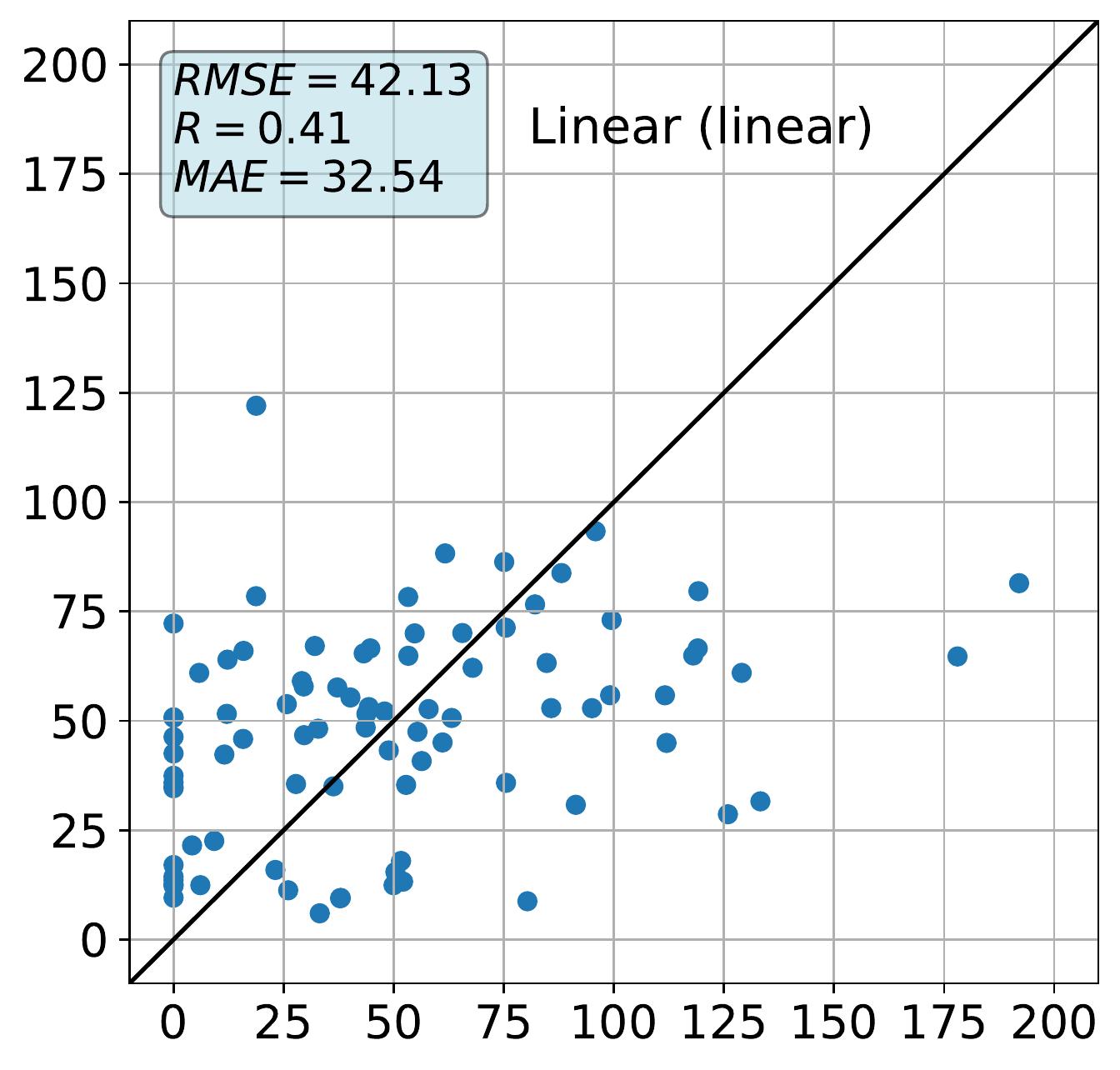}}
\vspace*{.2cm}
\\
\hspace*{.5cm}
 \subfloat[\label{s1a_seq_orig_scatter}]{\includegraphics[width=0.25\textwidth]{images/scatter/scatter_s1a_sequential_rmse_p.pdf}}
 \subfloat[\label{s1a_seq_gamma_scatter}]{\includegraphics[width=0.25\textwidth]{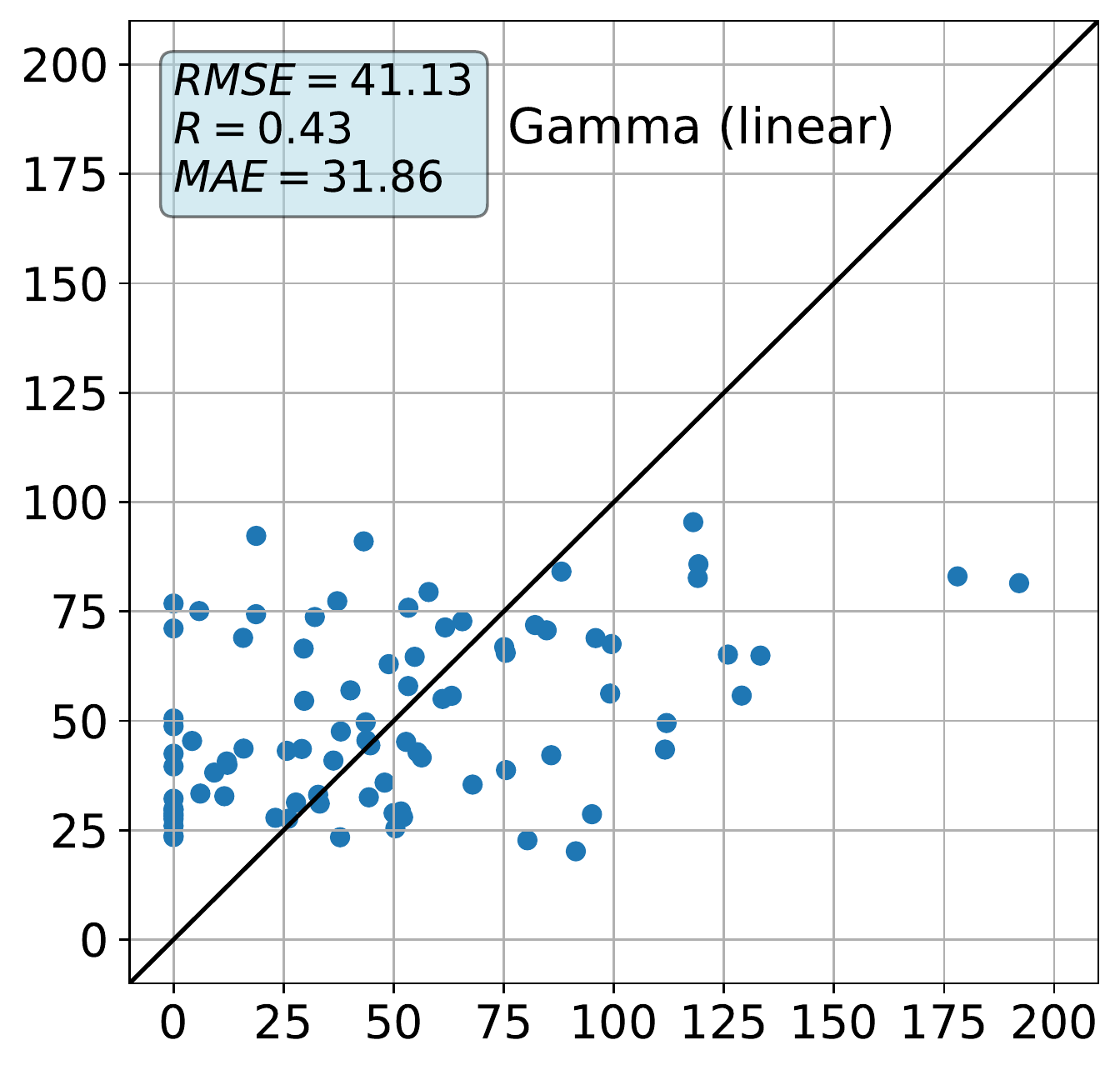}}
\subfloat[\label{s1a_seq_linear_scatter}]{\includegraphics[width=0.25\textwidth]{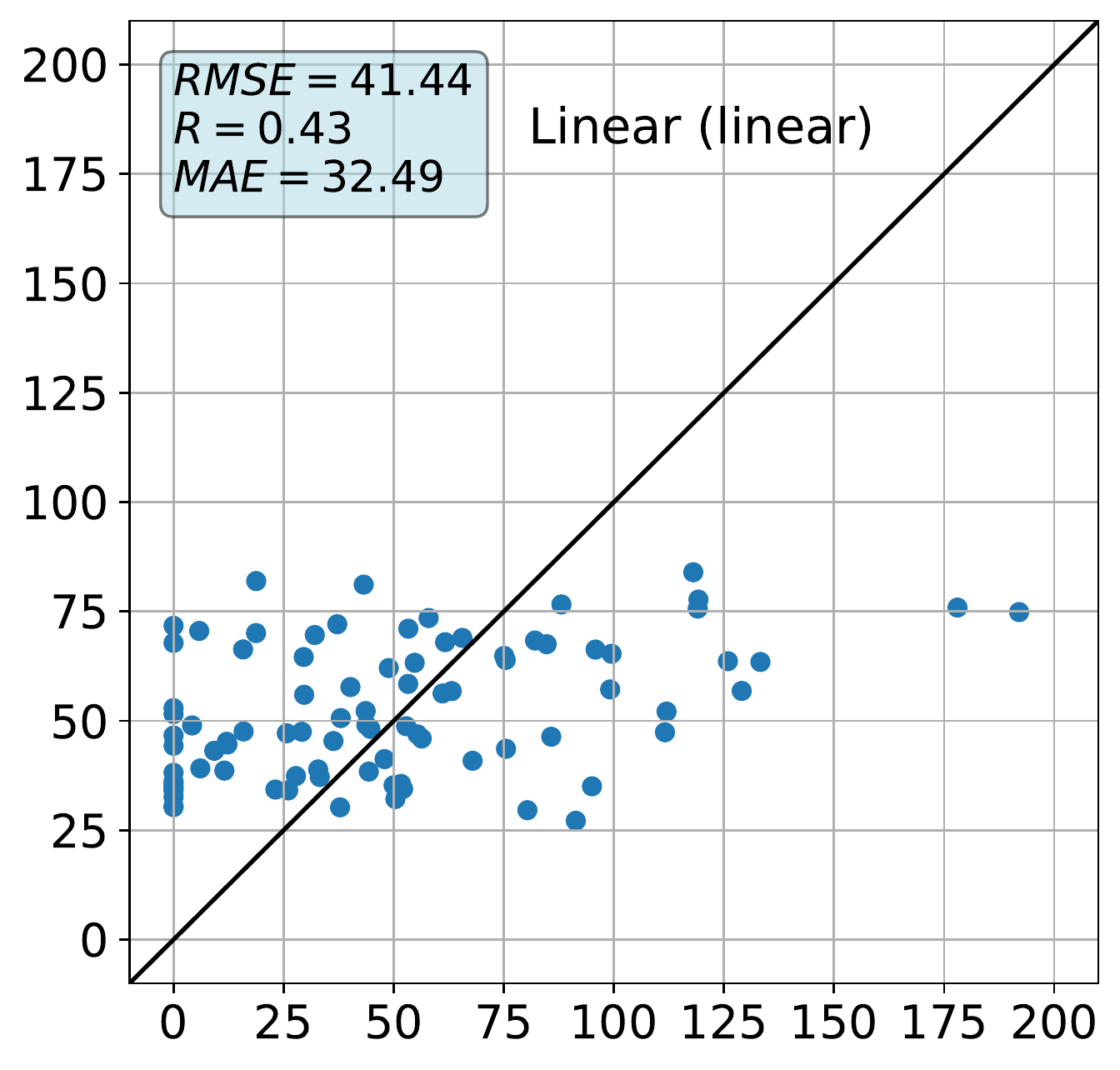}}
\vspace*{.2cm}
\\
\leavevmode\smash{\makebox[0pt]{\hspace{-30em}        
  \rotatebox[origin=l]{90}{\hspace{15em}
    Predicted AGB (Mg ha$^{-1}$)}
}} 
\vspace*{5cm}\hspace*{.50cm}{Ground Reference biomass (Mg ha$^{-1}$)}\\
\vspace{-5cm}
\caption{Scatter plots between predicted AGB and ground reference AGB data, $z$. Fig. \textbf{(a)}-\textbf{(c)}, \textbf{(f)} and \textbf{(i)} are reference images, corresponding to AGB predictions from the ALS-based regression model, the non-sequential \sen\ model, the LSGAN model trained with dataset on dB scale, the LSGAN model trained with dataset on linear scale and the baseline sequential \sen\ model trained with dataset on linear scale. Fig. \textbf{(d)}, \textbf{(g)} and \textbf{(j)} shows AGB predictions from respective model after calibration with gamma transform, while \textbf{(e)}, \textbf{(h)} and \textbf{(k)} shows corresponding results after calibration with a linear transform. The black lines are reference lines indicating 100\% correlation between $z$ and predictions. Units are in Mg ha$^{-1}$.}
    \label{fig:allscatter_cal}
\end{figure*}

\textbf{Results:} \figr{fig:allscatter_cal} shows results from the experiment with post-calibration of $\zhatyx$, i.e.\ scatter plots between $z$ and calibrated model-predicted AGB. To ease the comparison, we have provided some reference images, which are retrieved from the results presented in \Sec{sec:res}, i.e.\ scatter plots for the ALS-based model (a), the non-sequential \sen-based model (b), LSGAN on dB scale (c), LSGAN on linear scale (f) and the baseline sequential  \sen\ model (i). We show results for the calibrated LSGAN model on dB scale using gamma calibration in \figr{fig:allscatter_cal} (d) and linear calibration in (e). Furthermore, we show results for the calibrated LSGAN model on linear scale using gamma calibration in (g) and linear calibration in (h). \figr{fig:allscatter_cal} (j) and (k) show results for the calibrated baseline sequential  \sen\ model on linear scale using gamma calibration (j) and linear calibration (k).

We note from the figure that the gamma and linearly calibrated models yield slightly lower or lower RMSE/MAE for all models included in the evaluation. For the LSGAN models, the gamma calibration reduces R some, while the correlation coefficient is unchanged for the linear calibration. For the baseline sequential model, R is unchanged for both the gamma and the linear calibration. Unfortunately, neither of the models achieve as high R and low RMSE/MAE as the non-sequential \sen-based model, nor the non-sequential ALS-based model. However, the LSGAN models, with or without calibration, can still predict 0 AGB, while neither of the baseline sequential  models, with or without calibration, can produce such low AGB predictions. We conclude from this experiment that post-calibrating sequential AGB predictions against $z$ can yield some modest improvements to higher accuracy. However, as these possible modest improvements come with the cost of applying an extra step to the prediction process, we choose to omit it in the results provided in \Sec{sec:seq_eval}.

\end{document}